\documentclass[sigconf]{acmart}
\copyrightyear{2025}
\acmYear{2025}
\acmConference[KDD '25]{Proceedings of the 31st ACM SIGKDD Conference on Knowledge Discovery and Data Mining V.1}{August 3--7, 2025}{Toronto, ON, Canada}
\acmBooktitle{Proceedings of the 31st ACM SIGKDD Conference on Knowledge Discovery and Data Mining V.1 (KDD '25), August 3--7, 2025, Toronto, ON, Canada}
\acmDOI{10.1145/3690624.3709213}
\acmISBN{979-8-4007-1245-6/25/08}

\makeatletter
\gdef\@copyrightpermission{
 \begin{minipage}{0.3\columnwidth}
 \href{https://creativecommons.org/licenses/by/4.0/}{\includegraphics[width=0.90\textwidth]{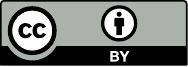}}
 \end{minipage}\hfill
 \begin{minipage}{0.7\columnwidth}
 \href{https://creativecommons.org/licenses/by/4.0/}{This work is licensed under a Creative Commons 
Attribution International 4.0 License.}
 \end{minipage}
 \vspace{5pt}
}
\makeatother

\usepackage{amsmath,flushend,boldline,paralist}
\usepackage{booktabs}
\usepackage{amsfonts}
\usepackage{amsbsy}
\usepackage{bbm}
\usepackage{algorithm}
\usepackage{algorithmic}
\usepackage{bm}
\usepackage{enumitem}
\usepackage{lipsum}
\usepackage{multirow}
\usepackage[normalem]{ulem}

\usepackage{graphicx}
\usepackage{subcaption}
\usepackage{hyperref}

\usepackage{xurl}

\AtBeginDocument{%
  }

\newtheorem*{problem}{Problem}
\renewcommand{\mathbf}{\bm}

\begin{document}
\title{APEX$^2$: Adaptive and Extreme Summarization for\\ Personalized Knowledge Graphs}

\author{Zihao Li}
\affiliation{%
  \institution{University of Illinois Urbana-Champaign}
  \state{Illinois}
  \country{USA}
}
\email{zihaoli5@illinois.edu}

\author{Dongqi Fu}
\affiliation{%
  \institution{Meta AI}
  \state{California}
  \country{USA}
}
\email{dongqifu@meta.com}

\author{Mengting Ai}
\affiliation{%
  \institution{University of Illinois Urbana-Champaign}
  \state{Illinois}
  \country{USA}
}
\email{mai10@illinois.edu}

\author{Jingrui He}
\affiliation{%
  \institution{University of Illinois Urbana-Champaign}
  \state{Illinois}
  \country{USA}
}
\email{jingrui@illinois.edu}

\renewcommand{\shortauthors}{Li et al.}

\begin{abstract}
Knowledge graphs (KGs), which store an extensive number of relational facts, serve various applications.
Recently, \textit{personalized knowledge graphs} (PKGs) have emerged as a solution to optimize storage costs by customizing their content to align with users' specific interests within particular domains.
In the real world, on the one hand, user queries and their underlying interests are inherently evolving, requiring PKGs to adapt continuously; on the other hand, the summarization is constantly expected to be as small as possible in terms of storage cost.
However, the existing PKG summarization methods implicitly assume that the user's interests are constant and do not shift.
Furthermore, when the size constraint of PKG is extremely small, the existing methods cannot distinguish which facts are more of immediate interest and guarantee the utility of the summarized PKG.
To address these limitations, we propose APEX$^2$, a highly scalable PKG summarization framework designed with robust theoretical guarantees to excel in adaptive summarization tasks with extremely small size constraints. To be specific, after constructing an initial PKG, APEX$^2$ continuously tracks the interest shift and adjusts the previous summary.
We evaluate APEX$^2$ under an evolving query setting on benchmark KGs containing up to 12 million triples, summarizing with compression ratios $\leq 0.1\%$. The experiments show that APEX outperforms state-of-the-art baselines in terms of both query-answering accuracy and efficiency.

\end{abstract}

\maketitle

\section{Introduction}

Knowledge graphs (KGs) have been proven an effective tool for constructing solutions in many application domains, such as healthcare, finance, cyber security, education, question answering, and social network analysis \cite{rizun2019knowledge, zou2020survey, ernst2014knowlife, kompare, liu2019combining, binet,prefnet,he2020constructing, DBLP:conf/sigir/LiAH24}.
Due to the ever-growing amount of data, encyclopedic knowledge graphs are becoming increasingly large and complex~\cite{faber2018adaptive, DBLP:journals/semweb/FarberBMR18}, such as DBpedia~\cite{DBLP:conf/semweb/AuerBKLCI07}, Freebase~\cite{DBLP:conf/sigmod/BollackerEPST08}, Wikidata~\cite{DBLP:journals/cacm/VrandecicK14}, and YAGO~\cite{DBLP:conf/www/SuchanekKW07}.
In contrast, KG users (e.g., individual people, systems, software packages) usually do not have very general interests but only care about a small portion of the whole KG for certain topics.
Therefore, personalized knowledge graphs (PKGs) have recently attracted much research attention for balancing storage cost and query-answering accuracy~\cite{DBLP:conf/icdm/SafaviBFMMK19, DBLP:conf/icde/KangLS22, DBLP:conf/esws/VassiliouAPK23}. In brief, a personalized knowledge graph (PKG) is extracted (summarized, compressed, or distilled) from a larger comprehensive knowledge graph. For KG and an individual user, a PKG has a limited size, but contains many entities/triples in the KG that the user is interested in, and can answer their personal queries appropriately. Furthermore, on the application side, each user will have a PKG. Since there might be many individual users, each PKG is expected to store as few facts as possible, to minimize the total storage cost.

\begin{figure}[t]
    \centering
    \includegraphics[width=0.40\textwidth]{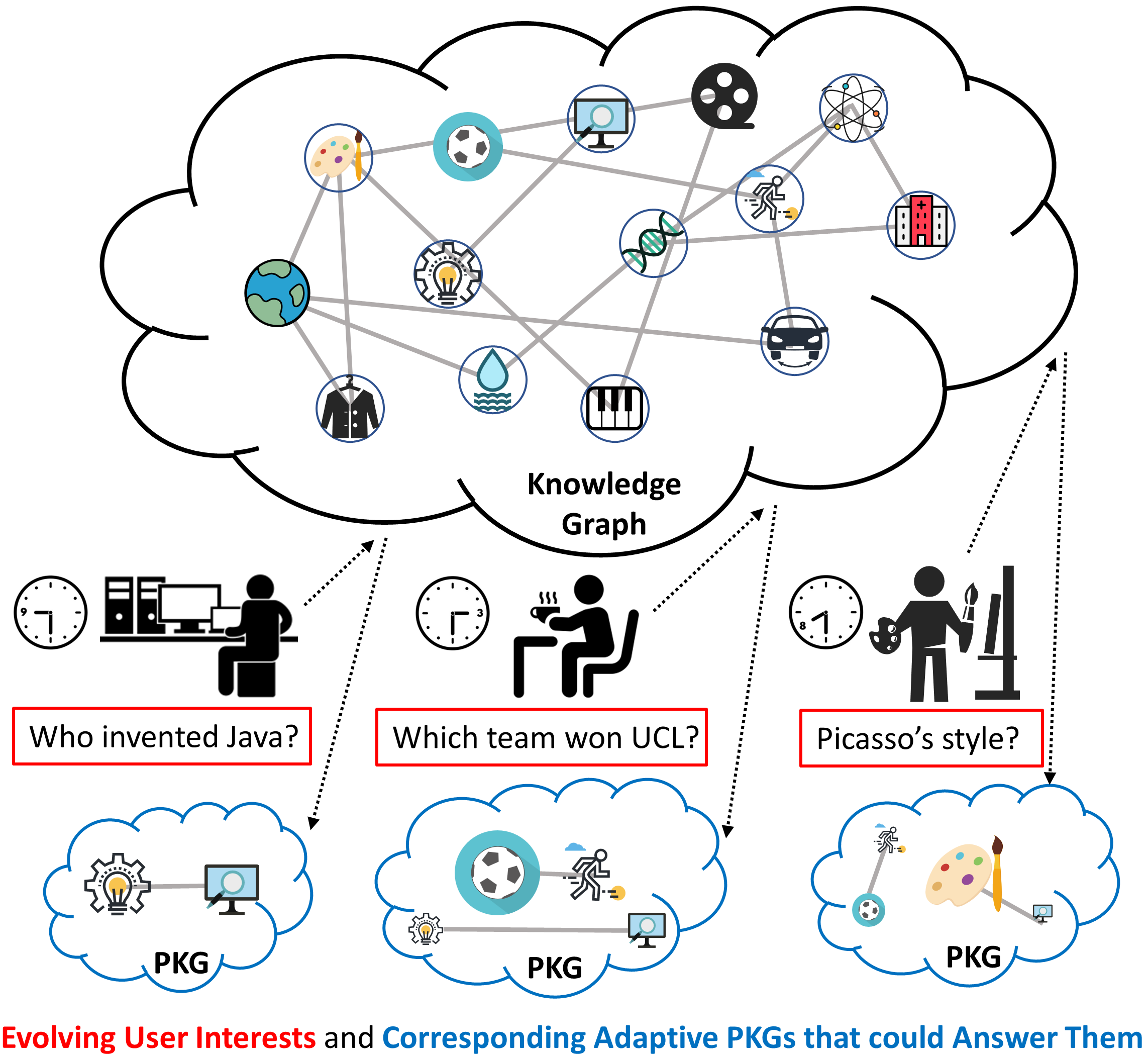}
    \caption{Example of Adaptive PKGs. The entire KG is stored on a cloud server, and the PKG is stored on the user's device. The initial PKG is constructed based on the query "Who invented Java" at 9am. Then, this PKG adapts to the same user's evolving queries at 3pm and 8pm.}
    \label{fig:HIN}
    \vspace{-3mm}
\end{figure}

As the individual KG’s size has been very large, re-summarizing PKG from scratch requires many computational resources, but in the real world, users' personal query interests may shift over time. As shown in Figure~\ref{fig:HIN}, in the morning (9am), the user works as a programmer and wonders about software engineering. During the afternoon coffee break (3pm), the same user cares about the UEFA Champions League (soccer games). In the evening (8pm), the same user likes painting and is curious about art. Under these circumstances, on one hand, an outdated summarized PKG might be sub-optimal, since outdated information agglomerates and adversely affects both storage cost and search performance~\cite{faber2018adaptive}. On the other hand, re-summarizing PKG from scratch at each timestamp causes unaffordable computational complexity. Given the fact that the KG is massively large~\cite{DBLP:conf/semweb/AuerBKLCI07, DBLP:conf/www/SuchanekKW07, DBLP:conf/sigmod/BollackerEPST08, DBLP:journals/cacm/VrandecicK14}, to make the PKG acceptably small, the compression ratio has to be extremely small. For example, the entire YAGO3 is about 300GB. Even with a 1\% compression ratio, the PKG is 3GB, which is still larger than most mobile applications. For the KGs measured in TB, more extreme compression is needed.

Motivated by the above use case, the previous work \cite{faber2018adaptive} informally introduced the problem of \textbf{adaptive PKG summarization}, which seeks to find a compact summary given the knowledge graph and query history. 
Take Figure~\ref{fig:HIN} as an example. We expect the adaptive PKG could, during the afternoon coffee break, quickly adapt from software engineering to more sports-related topics, decaying but not eliminating topics on software engineering. Then, at night, the PKG evolves with the user's interests in art topics. 
Moreover, in this paper, we study how to adaptively summarize the PKG under \textbf{extremely small storage constraints}.
Theoretically, in Appendix \ref{non-adaptability analysis}, we show that the existing PKG summarization methods~\cite{DBLP:conf/icdm/SafaviBFMMK19, DBLP:conf/icde/KangLS22}, even if re-run from scratch for the new interested topics, could not incorporate the new interests into the previously summarized PKG when query topics change, or corrupt under extremely small storage constraints.

To address these limitations, we propose APEX$^2$ (\textbf{A}da\textbf{p}tive and \textbf{Ex}treme Summ\textbf{a}rization for \textbf{Pe}rsonalized \textbf{K}nowledge Graph\textbf{s}), which enables summarization to incrementally evolve with user interests over time while satisfying extremely small storage constraints.
To the best of our knowledge, this work presents the first adaptive PKG summarization framework tailored for evolving query topics.
In brief, given a bunch of queries with different interested topics at different timestamps, 
APEX$^2$ works by modeling user interests through a heat diffusion process~\cite{DBLP:journals/jpdc/Cybenko89} and maintaining dynamic data structures that allow incremental updates. Under the extremely small storage limitation, APEX$^2$ incrementally infers the interest scores of the facts and picks the ones with the highest scores (i.e., immediately more interested by the user). Our contributions are summarized as:
\begin{itemize}[noitemsep,topsep=0pt,parsep=0pt,partopsep=0pt, leftmargin=*]
    \item \textbf{Problem Formulation and Theoretical Analysis}. We formally formulate the problem of adaptive PKG summarization with storage limitation. We provide theoretical analysis of the adaptability of existing PKG summarization methods. We also prove the efficiency and topic adaptability of our methods.
    \item \textbf{Algorithm}. We propose the adaptive and extreme PKG summarization solution APEX$^2$ and its variant APEX$^2$-N to address different circumstances with theoretical guarantees.
    \item \textbf{Experiment Evaluation}. We design extensive experiments under real-world query-answering scenarios on real KGs to show the effectiveness and efficiency of our proposed methods.
\end{itemize}

The rest of the paper is organized as follows.
In Section \ref{problem definition}, we introduce the problem setting and background.
In Section \ref{'APEX for Adaptive PKG Summarization'}, we introduce how APEX$^2$ dynamically and incrementally models user's evolving interests.
In Section \ref{sec: algorithms}, we formally propose APEX$^2$ and its variant APEX$^2$-N. We theoretically analyze both APEX$^2$ and APEX$^2$-N from multiple aspects. 
In Section \ref{experiments}, we report the experimental results showing the effectiveness and efficiency of our methods. We discuss related works in Section \ref{rw} and conclude in Section \ref{conclusion}.

\textbf{Reproducibility.} The code and the download instructions of KG datasets are provided. Refer to Appendix \ref{AP: reproducibility} for more details.

\section{Problem Definition}
\label{problem definition}
We use calligraphic letters (e.g., $\mathcal{A}$) for sets, bold capital letters for matrices (e.g., $\bm{A}$), parenthesized superscript to denote the temporal index (e.g., $\bm{A}^{(t)}$), unparenthesized superscript to denote the power (e.g., $\bm{A}^{k}$). For matrix indices, we use $\bm{A}_{i, j}$ to denote the entry in the $i^{th}$ row and the $j^{th}$ column. The notation used in our proposed APEX is summarized in Table~\ref{TB:Notations}.

\begin{table}[t]
\caption{Table of Notation}
\vspace{-3mm}
\centering
\scalebox{0.88}{
\begin{tabular}{|p{65pt}|p{185pt}|}
\hline Symbol&Definition and Description\\
\hline
\hline$\mathcal{G} = (\mathcal{E}, \mathcal{R}, \mathcal{T})$ & Knowledge graph being investigated, with entity set $\mathcal{E}$, relation set $\mathcal{R}$ and triple set $\mathcal{T}$\\
\hline$n$ & number of entities in $\mathcal{G}$ (i.e., $n = |\mathcal{E}|$)\\
\hline$\mathcal{P} = (\mathcal{E}_p, \mathcal{R}_p, \mathcal{T}_p)$ & Personalized Knowledge Graph, with entity set $\mathcal{E}_p$, relation set $\mathcal{R}_p$ and triple set $\mathcal{T}_p$\\
\hline$\mathcal{Q}$ & query log by the user\\
\hline$x_{ijk}$ & triple with entity $e_i$, $e_j$ and relation $r_k$ (i.e., $x_{ijk} = (e_i, r_k, e_j)$)\\
\hline$\mathbf{A}$ & (Sparse $n \times n$) adjacency matrix of $\mathcal{G}$\\
\hline$\mathbf{H}$ & (Sparse $n \times n$) heat matrix storing heat of entities (diagonal entries) and triples (non-diagonal entries)\\
\hline$\alpha$ & damping factor of neighbor\\
\hline$\epsilon$ & tunable tolerance\\
\hline$\gamma$ & decay factor\\
\hline$d$ & diffusing diameter\\
\hline$|\mathcal{S}|$ & number of elements in the set $\mathcal{S}$; Specifically, $|\mathcal{G}|$ represents number of triples in knowledge graph $\mathcal{G}$\\
\hline
\end{tabular}}
\label{TB:Notations}
\end{table}

\textbf{Knowledge Graph.} A knowledge graph $\mathcal{G} = (\mathcal{E}, \mathcal{R}, \mathcal{T})$ is defined by an entity set $\mathcal{E}$, a relation set $\mathcal{R}$ and a triple set $\mathcal{T}$. A triple $x_{ijk} = (e_i, r_k, e_j) \in \mathcal{T}$ is defined by entities $e_i$, $e_j$ and their relationship $r_k$. The undirected adjacency matrix $\mathbf{A}$ is defined as
\begin{equation}
    \bm{A}_{i, j}  = 1 \iff i \neq j \wedge \exists k \ s.t.\ x_{ijk} \ or \ x_{jik} \in \mathcal{T}
\end{equation}
\textbf{Personalized Knowledge Graph}. A PKG $\mathcal{P}$ of KG $\mathcal{G}$ is defined by an entity set $\mathcal{E}_p \subseteq \mathcal{E}$, a relation set $\mathcal{R}_p \subseteq \mathcal{R}$ and a triple set $\mathcal{T}_p \subseteq \mathcal{T}$. In our problem setting, a PKG is summarized from a KG according to the user's query log $\mathcal{Q}$. 

\textbf{Query Log}. Since most real queries to KGs consist of only one or two triples \cite{DBLP:journals/vldb/BonifatiMT20}, assuming we have the full access to the whole KG, we study simple queries with known answers. A \textbf{query log} $\mathcal{Q}$ consists of a number of queries.
Each \textbf{query} $q$ consists of a \textbf{query entity} $e$, \textbf{query relation} $r$, and a set of \textbf{answer entities} $\mathcal{A}$. A triple $x_{ijk} = (e_i, r_k, e_j)$ in the knowledge graph $\mathcal{G}$ is said to be an \textbf{answer triple} to a query q, if  $e_i = e \wedge r_k = r \wedge e_j \in \mathcal{A}$. A query may have multiple answer triples. For example, for the query “What movie did Christopher Nolan direct”, the query entity is ``Christopher Nolan” ("Nolan" for short), and the query relation is “directed\_movie” ("d\_m" for short). Suppose in the KG all triples with the head entity “Nolan” and the relation “d\_m” are (“Nolan”, “d\_m”, “Interstellar”), (“Nolan”, “d\_m”, “Tenet”) and (“Nolan”, “d\_m”, “Oppenheimer”).  Then these three triples are answer triples to the query, and the set \{“Interstellar”, “Tenet”, “Oppenheimer”\} is the set of answer entities. In this paper, ``for triples in $q$'' iterates the answer triples of $q$; ``$e \in q$'' iterates all entities accessed by $q$ as the query entity or answer entities; "$\mathcal{Q}^{(t+1)}\setminus\mathcal{Q}^{(t)}$" stands for all queries in $\mathcal{Q}^{(t+1)}$ but not in $\mathcal{Q}^{(t)}$.

\begin{problem}{Adaptive Personalized Knowledge Graph Summarization. ``$\{\}$'' means ``a sequence of'' here. $F1$ score is a conventional measure of seaching accuracy, defined in Appendix \ref{AP: F1Score}.}
\begin{description}
\item[Input:] 
(i) a knowledge graph $\mathcal{G}$\\
(ii) a sequence of varying end-user query interests, represented by a temporal query log $\{\mathcal{Q}^{(0)}, \mathcal{Q}^{(1)}, \mathcal{Q}^{(2)}, \ldots, \mathcal{Q}^{(T)}\}$\\ 
(iii) constant size budget $K$
\item[Output:] a sequence of personalized knowledge graph $\{\mathcal{P}^{(t)}\}$ of $\mathcal{G}$ for $t \in \{0,1, \ldots ,T-1\}$; each $\mathcal{P}^{(t)}$, whose number of triples |$\mathcal{P}^{(t)}$| is not greater than $K$, is able to answer as many queries in $\mathcal{Q}^{(t+1)}\setminus\mathcal{Q}^{(t)}$ as possible and as correctly as possible:
    \begin{equation}
        \underset{\{\mathcal{P}^{(t)}\}}{\arg\max} \sum_{0 \leq t \leq T-1}\sum_{q \in \mathcal{Q}^{(t+1)}/\mathcal{Q}^{(t)}}F1(\mathcal{P}^{(t)}, q) \ s.t. \ |\mathcal{P}^{(t)}| \leq K
    \end{equation}
\end{description}
\end{problem}
In this paper, when the size budget $K\leq 1\%$, we say the problem becomes \textbf{adaptive and extreme PKG summarization}. The smallest $K$ value that existing PKG summarization methods \cite{DBLP:conf/icdm/SafaviBFMMK19, DBLP:conf/icde/KangLS22, DBLP:conf/esws/VassiliouAPK23} have explicitly used is $10\%$. In our experiments, $K\leq 0.1\%$.

\textbf{Heat Diffusion}. Heat diffusion process is a natural way to model users' interest ~\cite{DBLP:conf/cikm/MaYLK08, DBLP:journals/jnca/PengZCYNJ18, DBLP:journals/corr/abs-2411-01410}, as warmer nodes are considered more immediately interesting to the user~\cite{faber2018adaptive}. 
Once a query is performed by the user, the queried area will gain more heat, and then globally, the heat of each node will be partially pushed to its neighbors.
In this work, we adopt a heat decay-inject-diffuse framework, with a visual example and more details provided in Appendix \ref{AP: heat_diffuse_visual_aid}.

\section{Adaptive PKG Summarization}
\label{'APEX for Adaptive PKG Summarization'}
In the adaptive PKG summarization scenario, compared to previous static PKG summarization~\cite{DBLP:conf/icdm/SafaviBFMMK19, DBLP:conf/icde/KangLS22, DBLP:conf/esws/VassiliouAPK23}, the major difference is that the user's interest may be shifting, manifested by the user's evolving query history.
The adapting may be simply achieved by re-applying summarization methods from scratch every time the user's query log has evolved.
But such a from-scratch solution lacks real-time efficiency.
A natural follow-up is: can we reuse the results from the previous summarization and further develop a real-time framework that can evolve incrementally with the user's interest? Moreover, when the storage constraint is extremely small, can we incrementally maintain a descending-order rank of the user's interests in each entity/relation/triple? To address these questions, we propose our solution APEX$^2$.

The core idea of APEX$^2$ is to maintain a real-time sparse heat structure that stores the user's interest and incrementally updates its content, then greedily chooses the triples with the highest heat to construct the summarization, even under extremely small storage limitations.
To adapt to the user's interests, we introduce a decay factor $\gamma$ into our framework. This factor is crucial to both adapting effectiveness and efficiency: (i) decaying previous interests gives higher priority to recent queries, which are more likely to represent the user's current interest; (ii) for a set of elements, decaying all of them does not affect their order and will not introduce additional computations to the heat ranking process. $\gamma$ controls the trade-off between adapting to new interests and retaining relevant information from past queries.

Systematically, our APEX$^2$ consists of three components for adaptive PKG summarization, i.e., \textit{Dynamic Model of User Interests}, \textit{Incremental Updating}, and \textit{Incremental Sorting}. In brief, \textit{Dynamic Model of User Interests} is proposed to model the user interest dynamically. Then based on the evolving interests modeled, \textit{Incremental Updating} tracks the new interests. \textit{Incremental Sorting} is necessary to construct a high-quality newly summarized PKG, under an extremely small storage constraint. Details of the three components are introduced through Subsections \ref{Dynamic Model of User Interests} -- \ref{Incremental Sorting}. Our end-to-end APEX$^2$ is summarized in Algorithm \ref{AG: APEX}.

\subsection{Dynamic Model of User Interests}
\label{Dynamic Model of User Interests}

In order to formulate more conveniently, we \textbf{start by freezing at a specific timestamp} $T$ ($T$ \textbf{as a constant}), and define $\bm{q}_{total}$ to store the number of times each entity is accessed as query entity or answer entity in the query log. If accessed as the answer to a query, the marginal value will be weighted by $\frac{1}{\# \textit{ of answers}}$.
We provide a simple example here. Suppose we have 5 entities (indexed 0, 1, 2, 3, 4). The first query is (entity 0, some relation, \{entity 1, entity 3\}), the second query is (entity 2, some relation, \{entity 0, entity 3\}), then $\mathbf{q}_{\rm total}$ will be a column vector $(1+0.5, 0.5+0, 0+1, 0.5+0.5, 0+0)=(1.5, 0.5, 1, 1, 0)$.
Assuming the user's temporal query log is $\mathcal{Q}^{(t)}$, we use $\mathcal{Q}^{(t)}\setminus\mathcal{Q}^{(t-1)}$ to denote new queries arriving at time $t$. Additionally, $\mathcal{Q}^{-1} = \emptyset$. Then,
\begin{equation}
\mathbf{q}_{\rm total} = \sum_{t=0}^{T}\mathbf{q}^{(t)} = \sum_{t=0}^{T}\sum_{i \in \mathcal{Q}^{(t)}\setminus\mathcal{Q}^{(t-1)}}\mathbf{q}_i
\label{EQ: q_total_static}
\end{equation}
where for each query $i \in \mathcal{Q}^{(t)}\setminus\mathcal{Q}^{(t-1)}$ with answer set $\mathcal{A}_i$, $\bm{q}_i$ is a vector with dimension $(|\mathcal{E}| \times 1)$ whose entries are
\begin{equation}
    \mathbf{q}_i[e] = \begin{cases} 1 & e \ \text{is the query entity of} \ \text{query}~ i \\ \frac{1}{|\mathcal{A}_i|} & e \ \text{is an answer entity to} \ \text{query}~ i \\ 0 & \text{otherwise}  \ \text{(} e \ \text{is unrelated to} \ \text{query}~ i \text{)} \end{cases}
\end{equation}
where $e \in \{1, \ldots, |\mathcal{E}|\}$ is the index of entities.

We model user's interest on an entity $e$ in a heat-diffusing style. Define $N_l(e)$ to be the $l$-hop neighbors of entity $e$. Additionally $N_0(e) = e$. A \textbf{topic} is a sub-area in the KG that has implicit inner connections. Such connections carry semantic meanings for humans (e.g., artistic, physical entities) and are modeled topologically by entities\footnote{Freebase documentation explicitly define topics to correspond to nodes in the KG~\cite{DBLP:conf/icdm/SafaviBFMMK19}. \url{https://developers.google.com/freebase/guide/basic_concepts?hl=en}}. With the straightforward inspiration that entities near the searched ones are likely to be in the user's interested topics, we model the user's static preference for entities as
\begin{equation}
\begin{split}
    {\rm Pr}(e| \mathcal{Q}) = \sum_{l=0}^d \alpha^l \sum_{e_o \in N_l(e)} \sum_{q \in \mathcal{Q}^{(T)}}\mathbbm{1}(e_o \in q)
\end{split}
\end{equation}

Equivalently, by Equation \ref{EQ: q_total_static}, the vector $\mathbf{e}(e)={\rm Pr}(e| \mathcal{Q})$ can be written in matrix expression as
\begin{equation}
\begin{split}
    \mathbf{e} = \mathbf{q}_{\rm total} + \alpha\mathbf{A}\mathbf{q}_{\rm total} + \alpha^2\mathbf{A}^2\mathbf{q}_{\rm total} + ... = \sum_{l = 0}^{d}\alpha^l\mathbf{A}^l\mathbf{q}_{\rm total}
\end{split}
\label{EQ: e_static}
\end{equation}
and we can use a closed form \cite{DBLP:conf/www/LiFH23} to calculate the case $d \to +\infty$:
\begin{equation}
    \lim_{d \to +\infty}\mathbf{e} = \sum_{l = 0}^\infty\alpha^l\mathbf{A}^l\mathbf{q}_{\rm total} = (\mathbf{I} - \alpha \mathbf{A})^{-1}\mathbf{q}_{\rm total}
\end{equation}

To model the user's interest in relations, we simply use a frequency-based approach. For a query $i\in\mathcal{Q}$,
\begin{equation}
\begin{split}
    {\rm Pr}(r_k|\mathcal{Q}) \propto {\sum_{i\in\mathcal{Q}} (\mathbbm{1}(r_k \in i))}
\end{split}
\label{EQ: rk_def_static}
\end{equation}
where $r_k$ is the $k$-th relation in KG $\mathcal{G}$, and $\mathbf{r}$ is a vector with dimension $(|\mathcal{R}| \times 1)$ as follows
\begin{equation}
\mathbf{r} = \sum_{t=0}^{T}\mathbf{q}_r^{(t)} = \sum_{t=0}^{T}\sum_{i \in \mathcal{Q}^{(t)}\setminus\mathcal{Q}^{(t-1)}}\mathbf{\tilde{q}}_{i}
\label{r}
\end{equation}
where $\mathbf{\tilde{q}}_{i}$ is the one-hot vector with dimension $(|\mathcal{R}| \times 1)$ corresponding to the searched relation in query $i \in \mathcal{Q}^{(t)}\setminus\mathcal{Q}^{(t-1)}$ at time $t$,
\begin{equation}
    \mathbf{\tilde{q}}_{i}[r] = \begin{cases} 1 & r \ \text{is the query relation of} \ \text{query}~ i \\ 0 & \text{otherwise}  \ \text{(} r \ \text{is unrelated to} \ \text{query}~ i \text{)} \end{cases}
\end{equation}
where $r \in \{1, \ldots, |\mathcal{R}|\}$ is the index of relations.

\textbf{Then, we are ready to introduce the dynamics} (\textbf{from now} $T$ \textbf{starts to evolve}) \textbf{and decaying factor} $\gamma$ into the problem on both entities and relations. The decaying factor $\gamma$ controls the trade-off between adapting to new interests and retaining relevant information from past queries. Involving decay factor $\gamma$ into $\mathbf{q}_{\rm total}, \mathbf{e}, \mathbf{r}$, while a heat-diffusing style still applies, at timestamp $T$, temporal expressions of Eqs. \ref{EQ: q_total_static}, \ref{EQ: e_static}, \ref{EQ: rk_def_static} become
\begin{equation}
\mathbf{q}_{\rm total}^{(T)} = \sum_{t=0}^{T}\gamma^{T-t}\mathbf{q}^{(t)} = \sum_{t=0}^{T}\gamma^{T-t}\sum_{i \in \mathcal{Q}^{(t)}\setminus\mathcal{Q}^{(t-1)}}\mathbf{q}_i
\label{q_total^T}
\end{equation}
\begin{equation}
\begin{split}
    \mathbf{e}^{(T)} = \sum_{l = 0}^{d}\alpha^l\mathbf{A}^l\mathbf{q}_{\rm total}^{(T)}
\end{split}
\label{e^T}
\end{equation}
\begin{equation}
\mathbf{r}^{(T)} = \sum_{t=0}^{T}\gamma^{T-t}\mathbf{q}_r^{(t)} = \sum_{t=0}^{T}\gamma^{T-t}\sum_{i \in \mathcal{Q}^{(t)}\setminus\mathcal{Q}^{(t-1)}}\mathbf{\tilde{q}}_{i}
\label{r^T}
\end{equation}

As for the objective function, we stick to GLIMPSE's choice for triple preference. Further, we define our objective to be based only on triple preference. The mathematical expressions are
\begin{equation}
\begin{split}
    {\rm Pr}(x_{ijk}|\mathcal{Q}) &\propto {\rm Pr}(e_i|\mathcal{Q}){\rm Pr}(r_k|\mathcal{Q}){\rm Pr}(e_j|\mathcal{Q})\\
    &= \mathbf{e}^{(T)}[i] \mathbf{r}^{(T)}[j] \mathbf{e}^{(T)}[k]
\end{split}
\label{EQ: pr_triple}
\end{equation}
\begin{equation}
\begin{split}
    \phi(\mathcal{P}, \mathcal{Q}) &= \sum_{x_{ijk} \in \mathcal{T}_p} \log {\rm Pr}(x_{ijk}|\mathcal{Q})\\
    &= \sum_{x_{ijk} \in \mathcal{T}_p} \log \mathbf{e}^{(T)}[i] \mathbf{r}^{(T)}[j] \mathbf{e}^{(T)}[k]
\end{split}
\end{equation}
in which case the greedy algorithm on triple preference still leads to the optimum in our setting (Please refer to Appendix \ref{non-adaptability analysis} for details).

\subsection{Incremental Updating}
An advantage of our model is that the user's preference on entities and relations at time $T$ can be incrementally updated from the previous timestamp $T-1$. Denoting $\bm{q}^{(T)} = \sum_{i \in \mathcal{Q}^{(T)}\setminus\mathcal{Q}^{(T-1)}}\mathbf{q}_i$, we derive the incremental updating equations for $\mathbf{q}_{\rm total}^{(T)}$, $\mathbf{e}^{(T)}$, and $\mathbf{r}^{(T)}$ as follows.
\begin{equation}
\mathbf{q}_{\rm total}^{(T)} = \sum_{t=0}^{T}\gamma^{T-t}\mathbf{q}^{(t)} =\gamma\sum_{t=0}^{T-1}\gamma^{T-1-t}\mathbf{q}^{(t)} + \mathbf{q}^{(T)}= \gamma\mathbf{q}_{\rm total}^{(T-1)} + \mathbf{q}^{(T)} 
\end{equation}
\begin{equation}
\begin{split}
    \mathbf{e}^{(T)} &= \sum_{l = 0}^{d}\alpha^l\mathbf{A}^l\mathbf{q}_{\rm total}^{(T)} = \sum_{l = 0}^{d}\alpha^l\mathbf{A}^l\gamma\mathbf{q}_{\rm total}^{(T-1)} + \sum_{l = 0}^{d}\alpha^l\mathbf{A}^l\mathbf{q}^{(T)} \\
    &= \gamma\mathbf{e}^{(T-1)} + \sum_{l = 0}^{d}\alpha^l\mathbf{A}^l\mathbf{q}^{(T)}
\end{split}
\end{equation}
\begin{equation}
\mathbf{r}^{(T)} = \sum_{t=0}^{T}\gamma^{T-t}\mathbf{q}_r^{(t)} = \gamma\sum_{t=0}^{T-1}\gamma^{T-1-t}\mathbf{q}_r^{(t)} + \mathbf{q}_r^{(T)}= \gamma\mathbf{r}^{(T-1)} + \mathbf{q}_r^{(T)} 
\end{equation}

To model the user's interest in triples, we define $\mathbf{H}$ as a sparse 3-dimensional temporal array, implemented using a dictionary,
\begin{equation}
\mathbf{H}^{(T)}[i][j][k] = \mathbf{e}^{(T)}[i] \mathbf{r}^{(T)}[j] \mathbf{e}^{(T)}[k]
\label{H^T}
\end{equation}

The updating of $\mathbf{H}^{(T)}$ from $\mathbf{H}^{(T-1)}$ is per-entry conditional. Assume at timestamp $T$, compared to timestamp $T-1$, if entries $i, k$ in $\mathbf{e}^{(T)}$ and entry $j$ in $\mathbf{r}^{(T)}$ are not updated (excluding decay), then $\mathbf{H}^{(T)}[i][j][k] = \gamma^3\mathbf{H}^{(T-1)}[i][j][k]$. Only entries in $\mathbf{H}^{(T)}[i][j][k]$ with any of those being updated need to be recalculated. The number of updated entries is small as we assign $d$ much less than the diameter of $\mathcal{G}$. Therefore, for each timestamp, we multiply $\mathbf{H}$ by $\gamma^3$ and then do a small-scale update.

\subsection{Incremental Sorting}
\label{Incremental Sorting}
After the user's preference is updated in real-time, we can directly sort the triples by their preferences and then pick the ones with the highest heat until the size budget is reached. However, the sorting algorithms usually cost $O(n\log_2 n)$ complexity~\cite{DBLP:journals/cacm/Hoare61b}. 

For every new timestamp, excluding decay (which does not change the order), only part of the triples' heat will be updated and recalculated. This indicates that it is possible to reuse the previous order and accelerate the process to get the up-to-date order. For the following problem, we propose an intuitive solution named incremental binary insertion sort, as shown in Algorithm \ref{AG: IBIS}.

\begin{problem}{Incremental Sorting}
\begin{description}
\item[Input:] (i) a sorted sortable instance $\mathcal{S}$, (ii) a set of changes $\mathcal{C}$, where each change is expressed as a tuple (from, to). For element deleting, the tuple will be (from, None); for element adding, the tuple will be (None, to).
\item[Output:] sorted $\mathcal{S}$ after applying all changes in $\mathcal{C}$.
\end{description}
\end{problem}

We prove the optimality of our incremental binary insertion sort algorithm by applying it to a simpler pure-adding case where all tuples in $\mathcal{C}$ are (None, to).

\begin{theorem}[Optimality of Incremental Binary Insertion Sort]
Given a sorted instance $\mathcal{S}$ and unsorted instance $\mathcal{C}$ with $|\mathcal{S}| = n$ and $|\mathcal{C}| = k$. When $k << n$, any deterministic comparison-based sorting algorithm must perform $\Omega(k\log_2n)$ new comparisons to sort $\mathcal{S} \cup \mathcal{C}$ in the worst case. Incremental Binary Insertion Sort achieves this optimal number of comparisons. (Proof in Appendix \ref{AP: A.5})
\label{THM: IBIS}
\end{theorem}

\vspace{-1mm}
By applying our incremental sorting algorithm when updating heat, the cost to choose triples with the highest heat becomes $\Omega(klog_2n)$ instead of $\Omega(nlog_2n)$ with $k << n$ at most timestamps.

\section{Algorithms}
\label{sec: algorithms}
\subsection{APEX$^2$ Framework}

Our APEX$^2$ combines user preference modeling, incremental heat updating and incremental sorting as both solutions and optimizations.
As shown in Algorithm \ref{AG: APEX}, in Steps 1--7, APEX$^2$ first initializes the data structures for both heat updating and incremental sorting, then performs pre-computing.
Then, in Steps 8--14, for each later timestamp a user inputs new queries, APEX$^2$ performs macroscopic decaying, incrementally updating and necessary recalculating. After that, the heat of triples gets incrementally sorted and a new PKG is constructed by picking the ones with the highest heat.

The adaptability of APEX$^2$ is ensured by the decaying operations, and we prove the effectiveness of APEX$^2$ in Theorem \ref{THM: APEX-effect}.
As for efficiency, intuitively, though the whole KG is available, APEX$^2$ only accesses the entities, relations and triples of the user's interest. Moreover, when the heat of a triple is decayed to a small enough value, APEX$^2$ would switch it to zero and such an out-of-interest triple does not require any further computational resource. These facts show that APEX$^2$ is highly scalable for large databases. We prove that \textbf{the incremental time complexity of APEX$^2$ is unrelated to the size of KG} in Theorem~\ref{THM: APEX-time}. Here, \textbf{incremental time complexity} means ``time complexity per adapting phase'', which is the total time in the adapting phase amortized by the number of for-loop iterations in Steps 8-14.
In the following theorems, \textbf{connectivity} of an area $\mathcal{V} = (\mathcal{E}_v, \mathcal{R}_v, \mathcal{T}_v)$ is defined as $\frac{\sum_{v \in \mathcal{E}_v}degree(v)}{|\mathcal{E}_v|}$.

\begin{theorem}[Effectiveness of APEX$^2$]
Assume two areas (topics) $\mathcal{U}$ and $\mathcal{V}$ with connectivity $c_u$ and $c_v$ are sub-KGs of $\mathcal{G}$, and the user initially queries $\mathcal{U}$ for $a$ times and starts to query $\mathcal{V}$, then APEX$^2$ takes $\log_\gamma \frac{1}{\frac{A}{B}(1-\gamma^a)+1}$ queries to adapt from $\mathcal{U}$ to $\mathcal{V}$, where $A = (\frac{1-(\alpha c_u)^{d+1}}{1-\alpha c_u})^{\frac{|\mathcal{E}_u| + 2|\mathcal{T}_u|}{|\mathcal{E}_u| + 3|\mathcal{T}_u|}}$ and $B = (\frac{1-(\alpha c_v)^{d+1}}{1-\alpha c_v})^{\frac{|\mathcal{E}_v| + 2|\mathcal{T}_v|}{|\mathcal{E}_v| + 2|\mathcal{T}_v|}}$. (Proof in Appendix \ref{AP: A.3})
\label{THM: APEX-effect}
\end{theorem}

For each query $q = (e, r, \mathcal{A})$, we can decompose it into a set of sub-queries $q_{sub} = \{(e, r, \{a\}) \forall a \in \mathcal{A}\}$. Similarly, we can decompose a query log $Q$ into a sub-query log. Theorem \ref{THM: APEX-time} shows the time complexity of APEX$^2$ is only related to the connectivity of KG and the number of sub-queries the user performed.
\begin{theorem}[Time Complexity of APEX$^2$]
 The incremental time complexity for APEX$^2$'s adapting phase is $O(c\cdot|\mathcal{Q}|^2\cdot \log_2(c\cdot|\mathcal{Q}|))$, where $Q$ is the query log decomposed into sub-queries (each query in $Q$ has only one query entity, one query relation and one answer). $c = \frac{nnz(\sum_{l=0}^{d}(\alpha\mathbf{A})^l)} {|\mathcal{E}|}$, where $nnz$ is the operator outputting the number of non-zero elements in a matrix, $\mathbf{A} \in \mathbb{R}^{n \times n}$ is the adjacency matrix of $\mathcal{G}$, $\mathcal{E}$ is the set of entities of $\mathcal{G}$. (Proof in Appendix \ref{PF: Time Complexity of APEX})
\label{THM: APEX-time}
\end{theorem}

If set a threshold $\epsilon_{ths}$ to eliminate small-enough entries to $0$, then after $log_\gamma \epsilon_{ths}$ timestamps, entries introduced by previous queries will be decayed to 0. In this case, the effective number of queries $|\mathcal{Q}|$ above can be bounded by a constant $log_\gamma \epsilon_{ths}$. By this operation, \textbf{the incremental time complexity is further optimized to} $O(c\cdot \log_2(c))$, where $c = \frac{nnz(\sum_{l=0}^{d}(\alpha\mathbf{A})^l)}{|\mathcal{E}|}$ is the average number of neighbors within $d$-hops. In other words, the time complexity of APEX$^2$ updating is only related to the connectivity of KG.

\subsection{APEX$^2$ Variant: APEX$^2$-N}
In APEX$^2$, we model the user's interest in triples in Eq.~\ref{EQ: pr_triple}, where we assign equal weights to entities and relations. However, the user may put more attention on entities than relations when performing queries. For example, a user who searched ``what pieces of music did Taylor Swift create'' might be more likely to search ``what's Taylor Swift's music style'' than ``what pieces of music did Bill Evens create'' later on. In this case, we propose APEX$^2$-N, a variant of APEX$^2$ that gives higher weights to entities than relations. 

APEX$^2$-N only incrementally tracks and sorts the heat of entities but not for relations. In other words, APEX$^2$-N gives weight $1$ to entities and $0$ to relations. APEX$^2$-N is designed mainly for adaptive solutions of PKG summarization, and we leave the trade-off between weights on entities and relations to future work. Since APEX$^2$-N is a variation of APEX$^2$, we summarize the detailed operations of APEX$^2$-N in Algorithm \ref{AG: APEX-N} in the Appendix. We also give proof of the effectiveness and efficiency of APEX$^2$-N as follows.

\begin{theorem}[Effectiveness$^2$ of APEX-N]
Assume two areas (topics) $\mathcal{U}$ and $\mathcal{V}$ with connectivity $c_u$ and $c_v$ are sub-KGs of $\mathcal{G}$. If a user initially queries $\mathcal{U}$ for $a$ times and starts to query $\mathcal{V}$, then APEX$^2$-N takes $\log_\gamma \frac{1}{\frac{A}{B}(1-\gamma^a)+1}$ queries to adapt from $\mathcal{U}$ to $\mathcal{V}$, where $A = \frac{1-(\alpha c_u)^{d+1}}{1-\alpha c_u}$ and $B = \frac{1-(\alpha c_v)^{d+1}}{1-\alpha c_v}$. (Proof in Appendix \ref{AP: A.4})
\label{THM: APEX-N-effect}
\end{theorem}

\begin{theorem}[Time Complexity of APEX$^2$-N]
 The incremental time complexity for APEX$^2$-N's adapting phase is $O(c\log_2(c|\mathcal{Q}|))$, where $\mathcal{Q}$ is the query log decomposed into sub-queries (each query in $\mathcal{Q}$ has only one query entity, one query relation and one answer). And $c = \frac{nnz(\sum_{l=0}^{d}(\alpha\mathbf{A})^l)} {|\mathcal{E}|}$, where $\mathbf{A} \in \mathbb{R}^{n \times n}$ is the adjacency matrix of $\mathcal{G}$, $\mathcal{E}$ is the set of entity of $\mathcal{G}$, and $nnz$ is the operator outputting the number of non-zero elements in a matrix. (Proof in Appendix \ref{PF: Time Complexity of APEX-N})
\label{THM: APEX-N-time}
\end{theorem}

Like APEX$^2$, \textbf{the time complexity of APEX$^2$-N can also be optimized to $O(c\log_2c)$ by setting a threshold value}. In the future, Both APEX$^2$ and APEX$^2$-N may be extended to the fully dynamic setting, where the KG itself can evolve. New entities can be reserved as dummy nodes. When a new entity, relation, or triple is added, the initial heat is zero, therefore we only need to update the adjacency matrix and start tracking their heat from the next timestamp. When a triple is deleted, we clear its heat to zero. When an entity or relation is removed, it means there is no triple with that entity, and we can safely clear its heat to zero.

\section{Experiments}
\label{experiments}
\subsection{Experimental Setup}
\subsubsection{Datasets}
\label{datasets}
We use YAGO 3 \cite{DBLP:conf/www/SuchanekKW07}, DBPedia 3.5.1 \cite{DBLP:conf/semweb/AuerBKLCI07}, MetaQA \cite{DBLP:conf/aaai/ZhangDKSS18} and Freebase~\cite{DBLP:conf/sigmod/BollackerEPST08} as knowledge graph datasets in our experiments. The basic information of knowledge graphs is summarized in Table \ref{statistics of KGs}.
For YAGO, DBPedia and Freebase, we use synthetic queries that follow the logic and structure of Linked SPARQL Queries’ DBpedia data dump\footnote{\url{https://files.dice-research.org/archive/lsqv2/dumps/dbpedia/}}. The same format has been used in previous work \cite{DBLP:conf/icdm/SafaviBFMMK19}.
For MetaQA, we use the queries provided in the dataset.
More details about the datasets can be found in Appendix \ref{AP: Datasets}, and we validate the high quality of our synthetic queries in Appendix \ref{AP: Queries}.

\vspace{-2mm}

\begin{table}[h]
\centering
\caption{Statistics of Knowledge Graphs}
\vspace{-4mm}
\label{statistics of KGs}
\scalebox{0.85}{
\begin{tabular}{|l|l|l|l|l|}
\hline
KG      & $|\mathcal{E}|$  & $|\mathcal{R}|$     & $|\mathcal{T}|$ \\ \hline\hline
YAGO    & 4,267,316 & 38 & 12,403,275 \\ \hline
DBPedia & 4,616,347 & 1,043   & 10,974,936 \\ \hline
MetaQA  & 43,234 & 13   & 231,103 \\ \hline
Freebase& 14,541 & 237   & 310,116 \\ \hline
\end{tabular}
}
\end{table}

\vspace{-3mm}

\subsubsection{Baselines}
We choose several state-of-the-art methods as the baselines. We compare sampling-based knowledge graph summarization algorithm (GLIMPSE \cite{DBLP:conf/icdm/SafaviBFMMK19}), merging-based graph summarization algorithm (PEGASUS \cite{DBLP:conf/icde/KangLS22}), workload-based knowledge graph summarization algorithm (iSummary \cite{DBLP:conf/esws/VassiliouAPK23}), random walk with restart on knowledge graph (Personalized PageRank \cite{DBLP:conf/www/LiFH23}), together with our APEX$^2$ and APEX$^2$-N.
Details of GLIMPSE, PEGASUS and iSummary are provided in section \ref{non-adaptability analysis}.
The PPR baseline calculates the PageRank vector personalized to $\mathbf{q}_{\rm total}$ and constructs the summarization by continuously adding the most relevant entity.

\subsubsection{Re-summarization Interval.}\label{sec: re-sum interval} By design, the baselines cannot take temporal query logs as inputs. To enable the baselines to handle adaptive PKG summarization problem, we let them output new PKGs after a certain amount $R$ of timestamps. The choice of $R$ affects the performances of baseline methods. If $R$ is small (i.e., $R=1$ means re-summarize every timestamp), then the re-summarization happens frequently, and the baselines will become very slow. If $R$ is large, then the baseline summaries are outdated for most timestamps. To pick a good $R$ for fair comparisons, we conduct pre-experiments in Appendix \ref{AP: pre-exp} and find that $R=9$ is a good effectiveness-efficiency trade-off for baselines. 

In our design, APEX$^2$ and APEX$^2$-N can also take multiple queries at one time by masking the summary updating phase (line 13-14 in Algorithm 
\ref{AG: APEX} and line 13-17 in Algorithm \ref{AG: APEX-N}) for non-summary timestamps. We use $R_{APEX}$ to denote that they update the PKG every $R_{APEX}$ timestamp. By default, $R_{APEX} = 1$, and we provide a comprehensive study on $R_{APEX}$ in Section \ref{sec: study of R_APEX}.

\vspace{-1mm}

\subsubsection{Metrics}
Same as previous research works \cite{DBLP:conf/icdm/SafaviBFMMK19, DBLP:journals/access/TadesseLXY19, DBLP:journals/nar/0001AL21}, we use F1 score \cite{DBLP:conf/ecir/GoutteG05} on the very next query as the metric for searching effectiveness. More details of this can be found in Appendix \ref{AP: F1Score}.

\subsection{Experimental Settings}
\label{'Experimental Setting'}
We show the outperformance of APEX$^2$ and APEX$^2$-N through auto-regressive\footnote{Term borrowed from statistics. We feed PKG adaption methods the very next query for testing, then the same query is used for training. Iterate until all queries are used.} style experiments. We set the default hyperparameters $\gamma = 0.5$, $\alpha = 0.3$, $d = 1$ and PageRank restart probability to be $0.85$. We set the compression ratio to be $0.000001 = 0.0001\%$ (one in a million) for YAGO and DBPedia, $0.0001=0.01\%$ (one in ten thousand) for MetaQA, $0.0005 = 0.05\%$ for Freebase.

\textbf{Generate user queries}. To calculate the average and standard deviation, we simulate 10 users to query the KGs. Following the norm set by Freebase discussed in section \ref{Dynamic Model of User Interests}, we model the abstract concept of topic by "queries with the same query entity"\footnote{In fact, if a topic is defined in terms of entities, then the ability to adapt entity shift is a sufficient condition for adapting topic shift.}. To simulate a real-world querying scenario with interest shift, for each KG, we generate 200 queries on 20 topics for each user. Each group of 10 consecutive queries are in the same topic. We associate each query $t$ with timestamp $t-1$.
For MetaQA, we first categorize the provided queries into different topics by query entity, then randomly sample 20 distinct query entities. After that, we randomly choose 10 queries on each of the 20 topics.
For DBPedia, YAGO and Freebase, we synthetically generate queries by randomly choosing 20 query entities in the KG. Then for each query, we randomly choose 10 relations (with possible multiplicity) that the query entity has. Finally, we include all entities $e$ satisfying (query entity, chosen relation, $e$) $\in \mathcal{T}$ into the answer set. 
We study 1-hop simple queries with known answers because in real life there is only a small portion of complex queries~\cite{DBLP:journals/vldb/BonifatiMT20}, which can be decomposed into simple queries.

\textbf{Query Answering Evaluation}. After loading KG and the queries of a user, we adaptively summarize the KG. (i) For GLIMPSE, PEGASUS, iSummary, PageRank, we construct an initial summary using the first query, then re-summarize after each $R=9$ timestamps
(i.e., queries). (ii) For APEX$^2$ and APEX$^2$-N, they both evolve every timestamp whenever the user performs a new query.
We calculate the F1 score of the very next query after re-summarization for all the methods. For example, if at timestamp $t$ the summarization method performs re-summarization or evolving using query $t+1$, then at timestamp $t+1$ we search the query $t+2$ in the new PKG and calculate the F1 score. 

\begin{figure}[t]
    \centering
    \includegraphics[width=0.48\textwidth]{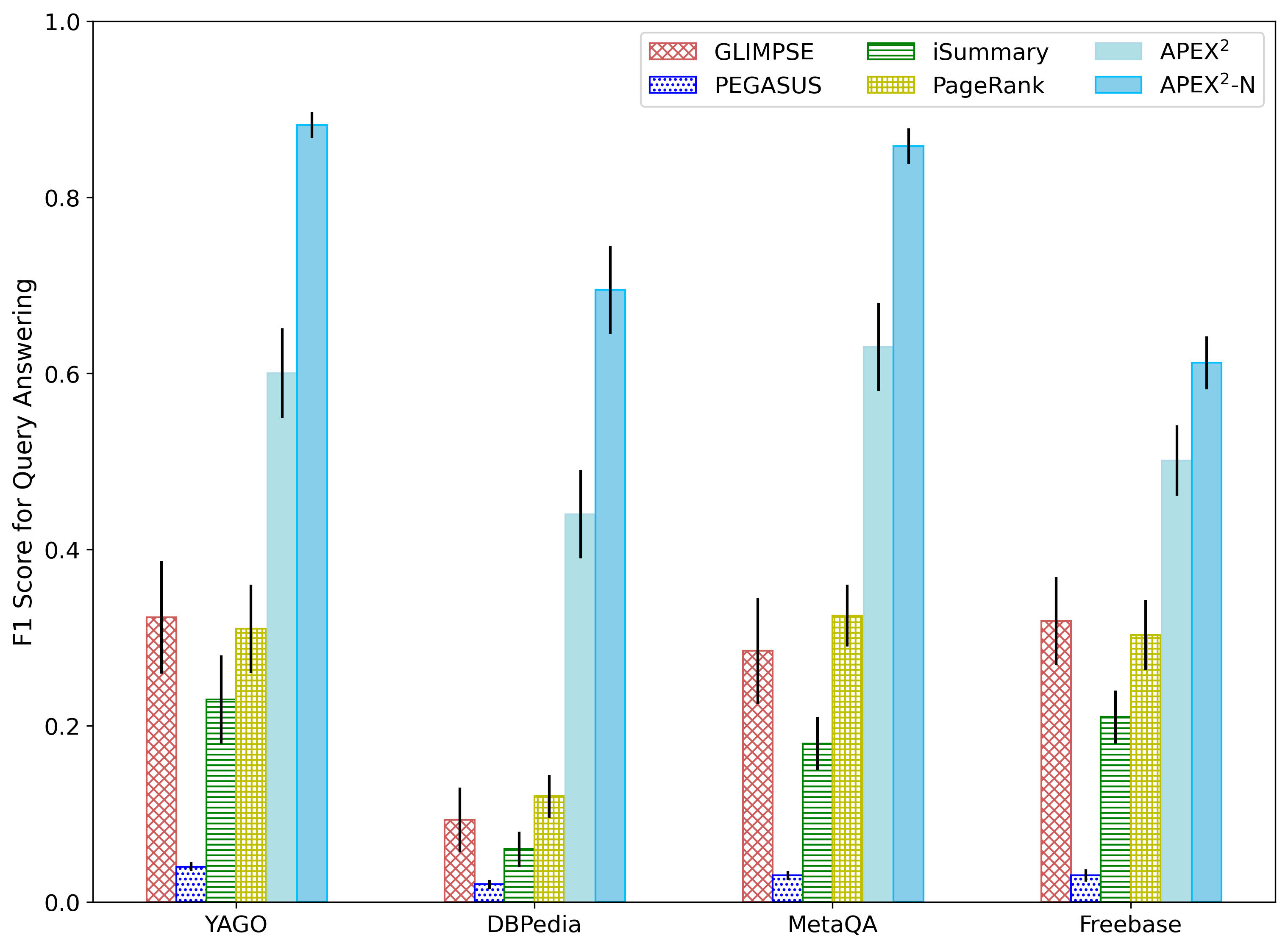}
    \caption{Effectiveness Comparison Under Querying Scenario}
    \label{fig: effectiveness}
\end{figure}

\subsection{Comparisons}
\textbf{Effectiveness Comparison}.
We measure the F1 score of each PKG summarization method in the auto-regressive querying scenario for 10 users.
We report the mean and standard deviation of sampled timestamps in Figure \ref{fig: effectiveness}.
First, both APEX$^2$ and APEX$^2$-N outperform the existing baseline methods on the F1 score in all cases.
Second, APEX$^2$ and APEX$^2$-N remain highly effective even if the knowledge graph is large and the compression rate is extremely small, showing the scalability of our methods.
Third, APEX$^2$-N outperforms all methods, showing the necessity of considering users' attention more on entities than relations. The baseline PEGASUS performs worst, possibly because this method is not designed specifically for knowledge graphs. This provides evidence that traditional graph algorithms should be reconsidered before being applied to the KG domain. In later comparisons, we only compare with GLIMPSE and PageRank baselines because they have acceptable sub-optimal performances.

\textbf{Efficiency Comparison}.
We measure the time consumption of PKG summarization methods to produce one summary under the $R=R_{APEX}=1$ setting.
We report the mean and standard deviation in Table \ref{efficiency comparison}.
PEGASUS and iSummary are not coded in Python. We add an extremely strong baseline in terms of efficiency: the parallel implementation of PageRank ``ParallelPR" with walk-length 1 and record its execution time, which is almost the optimal time cost that any diffusion-based algorithm can achieve.

\begin{table}[h]
\centering
\caption{Efficiency Comparison $(\downarrow)$ (unit: seconds)}
\vspace{-3mm}
\label{efficiency comparison}
\scalebox{0.9}{
\begin{tabular}{|l|l|l|l|l|}
\hline
Methods    & YAGO & DBPedia & MetaQA & Freebase\\ \hline \hline
GLIMPSE    & 192.1$\pm$27.92      & 148.4$\pm$114.8      & 1.366$\pm$0.089  & 1.581$\pm$0.093 \\ \hline
PageRank   & 22.81$\pm$259.7      & 2.615$\pm$0.136      & 0.032$\pm$0.003  & 0.144$\pm$0.011 \\ \hline
ParallelPR & 1.947$\pm$2.061      & 1.442$\pm$0.031      & 0.016$\pm$0.002  & 0.019$\pm$0.002 \\ \hline
APEX$^2$   & 6.354$\pm$5.388      & 4.655$\pm$1.108      & 0.055$\pm$0.035  & 0.112$\pm$0.048 \\ \hline
APEX-N$^2$ & 2.528$\pm$0.502      & 3.305$\pm$0.041      & 0.018$\pm$0.002  & 0.024$\pm$0.003 \\ \hline
\end{tabular}
}
\vspace{-2mm}
\end{table}


Our experiments show that both APEX$^2$ and APEX$^2$-N are much faster than re-running GLIMPSE for adaptive PKG summarization. Respectively, APEX$^2$ and APEX$^2$-N outperform GLIMPSE by $\mathbf{20\times}$ to $\mathbf{30\times}$ and $\mathbf{40\times}$ to $\mathbf{75\times}$. If we regard ParallelPR as the optimal time complexity, then compared to GLIMPSE, APEX$^2$ and APEX$^2$-N get $\mathbf{30\times}$ to $\mathbf{45\times}$ and $\mathbf{80\times}$ to $\mathbf{400\times}$ closer to the optimal. Overall, APEX$^2$ and APEX$^2$-N have similar efficiency performance to PageRank (walk length 1), which is a very strong baseline in time complexity. Furthermore, our APEX$^2$ and APEX$^2$-N have efficiency close to ParallelPR. Moreover, our experiments show that APEX$^2$ and APEX$^2$-N are more scalable because they outperform all the baseline methods in the largest YAGO dataset. Considering APEX$^2$-N and APEX$^2$ together, APEX$^2$-N has better experimental efficiency because it does not consider the user's interest in relationships. 
This means APEX$^2$-N could be a good choice for entity interest tracking and suits tasks where the interest in triples is not very necessary.

\begin{figure}[t]
    \centering
    \includegraphics[width=0.35\textwidth]{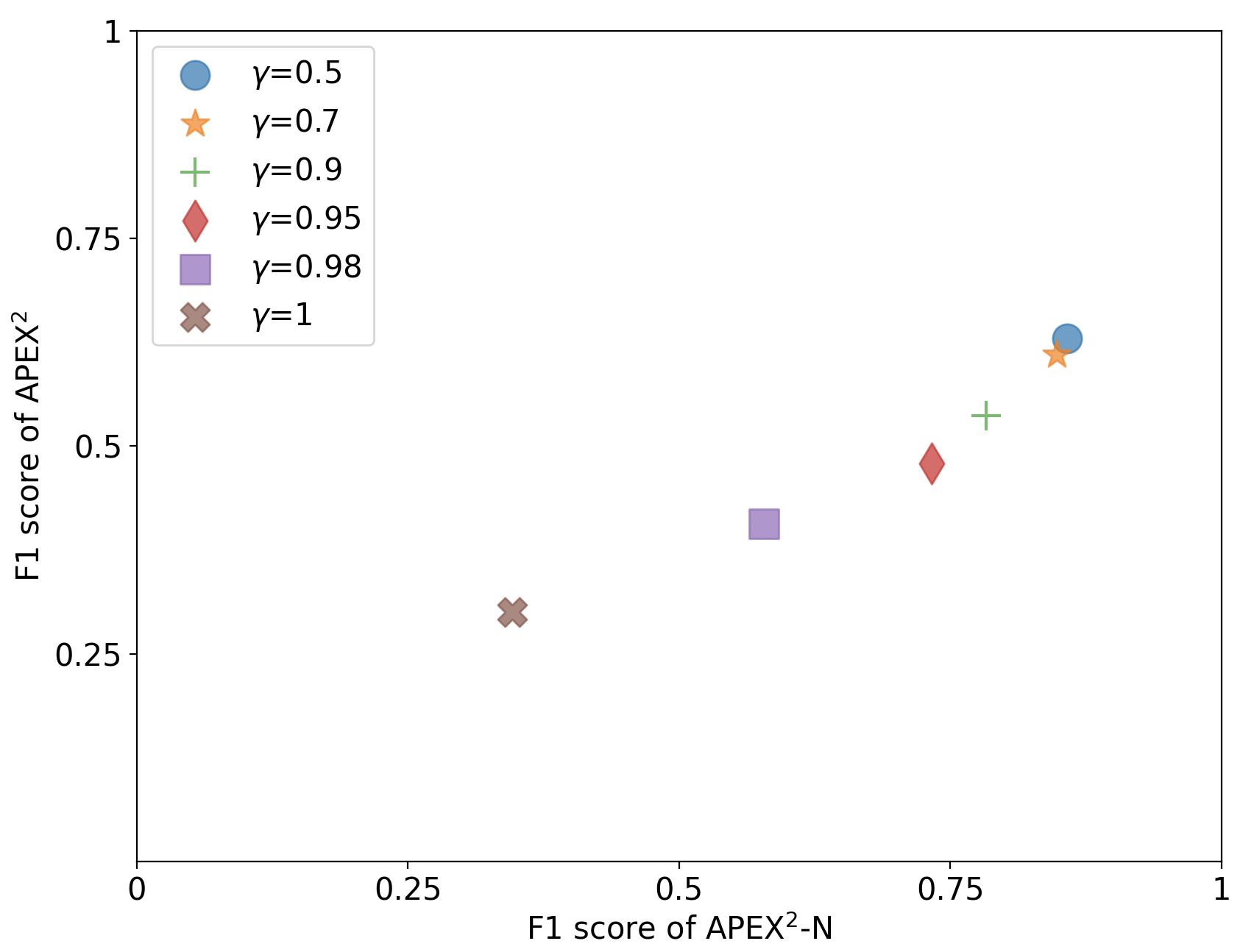}
    \caption{Querying Effectiveness in Multiple Decay Levels}
    \label{fig: ablation_gamma}
\end{figure}

\subsection{Ablation Study}
\subsubsection{Decay Ablation}
In our design, decaying (i.e., a forgetting mechanism), is the key to the adaptive and extreme summarization. Here, we compare the effectiveness under different levels of decaying to further show the importance of decaying.
We run the user querying scenario in MetaQA with varying $\gamma$ values. Other hyperparameters are set to be the same as Section \ref{'Experimental Setting'}. We use the same query generated for MetaQA and report both APEX$^2$ and APEX$^2$-N's average F1 score of 10 users in Figure \ref{fig: ablation_gamma}. 
Larger $\gamma$ means lower decay level (less extent of decay).
When $\gamma = 1$, the decaying is completely eliminated, and the PKG is full of outdated interests, resulting in low F1 scores of both APEX$^2$ and APEX$^2$-N.
In general, starting from $\gamma = 0.5$, the effectiveness does not change much until $\gamma = 0.9$, and then the performance gets worse massively from $\gamma = 0.9$ to $\gamma = 1$.
It turns out that when the storage space is very limited, decaying the previous interests can pave the way for personalized summarization.

\subsubsection{Component Ablation}
Aiming to accelerate the computation, after the necessary dynamic modeling of user interests, we designed incremental updating and incremental sorting. These two components serve to accelerate the computation and do not affect the computational results; therefore, to show that all three components of the APEX2$^2$ framework contribute to the overall efficiency performance, we can compare the mean execution time in seconds shown as follows.

\begin{table}[h]
\centering
\caption{APEX$^2$ Ablations' Efficiency $(\downarrow)$ in seconds}
\vspace{-3mm}
\label{TB: component_ablation}
\scalebox{0.85}{
\begin{tabular}{|l|l|l|l|l|}
\hline
Ablations    & YAGO & DBPedia & MetaQA & Freebase\\ \hline \hline
APEX$^2$-complete    & 6.354      & 4.655      & 0.055  & 0.110 \\ \hline
APEX$^2$-without-inc-updating & 124.155      & 91.679      & 0.845  & 0.912 \\ \hline
APEX$^2$-without-inc-sorting  & 18.643      & 15.212      & 0.124  & 0.162 \\ \hline
APEX$^2$-dynamic-model-only	  & 167.654      & 125.687     & 1.287  & 1.421 \\ \hline
\end{tabular}
}
\end{table}

Here, APEX$^2$-complete is the complete version of APEX$^2$; APEX$^2$-without-inc-updating removes incremental updating; APEX$^2$-without-inc-sorting replaces incremental sorting with normal re-sorting; APEX$^2$-dynamic-model-only removes both. From the results, all the components contributes to the acceleration, and incremental updating plays a crucial role in design.

\begin{figure*}[t]
    \centering
    \begin{subfigure}[]{0.24\textwidth}
        \centering
        \includegraphics[width=\textwidth]{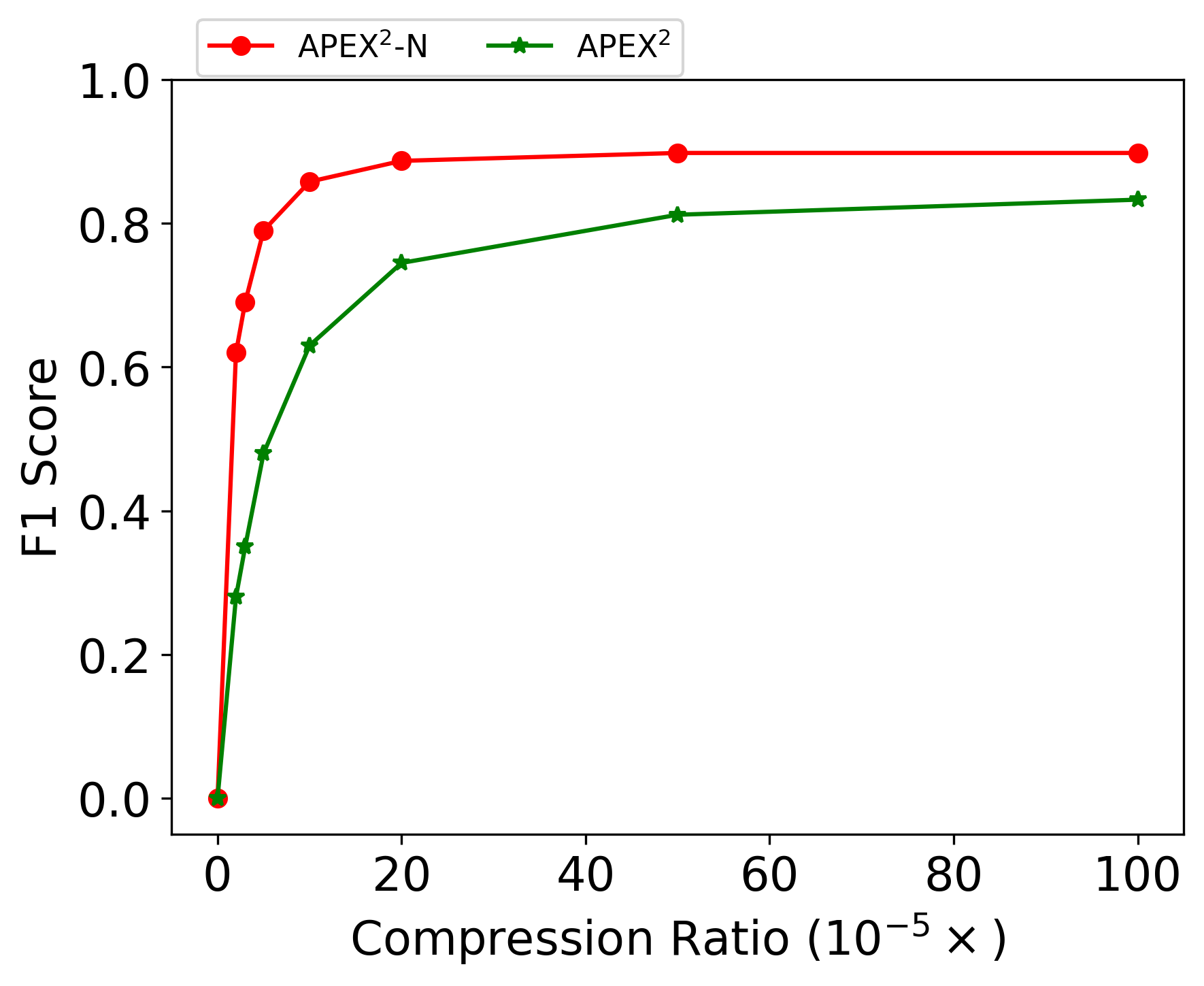}
    \end{subfigure}
    \begin{subfigure}[]{0.24\textwidth}
        \centering
        \includegraphics[width=\textwidth]{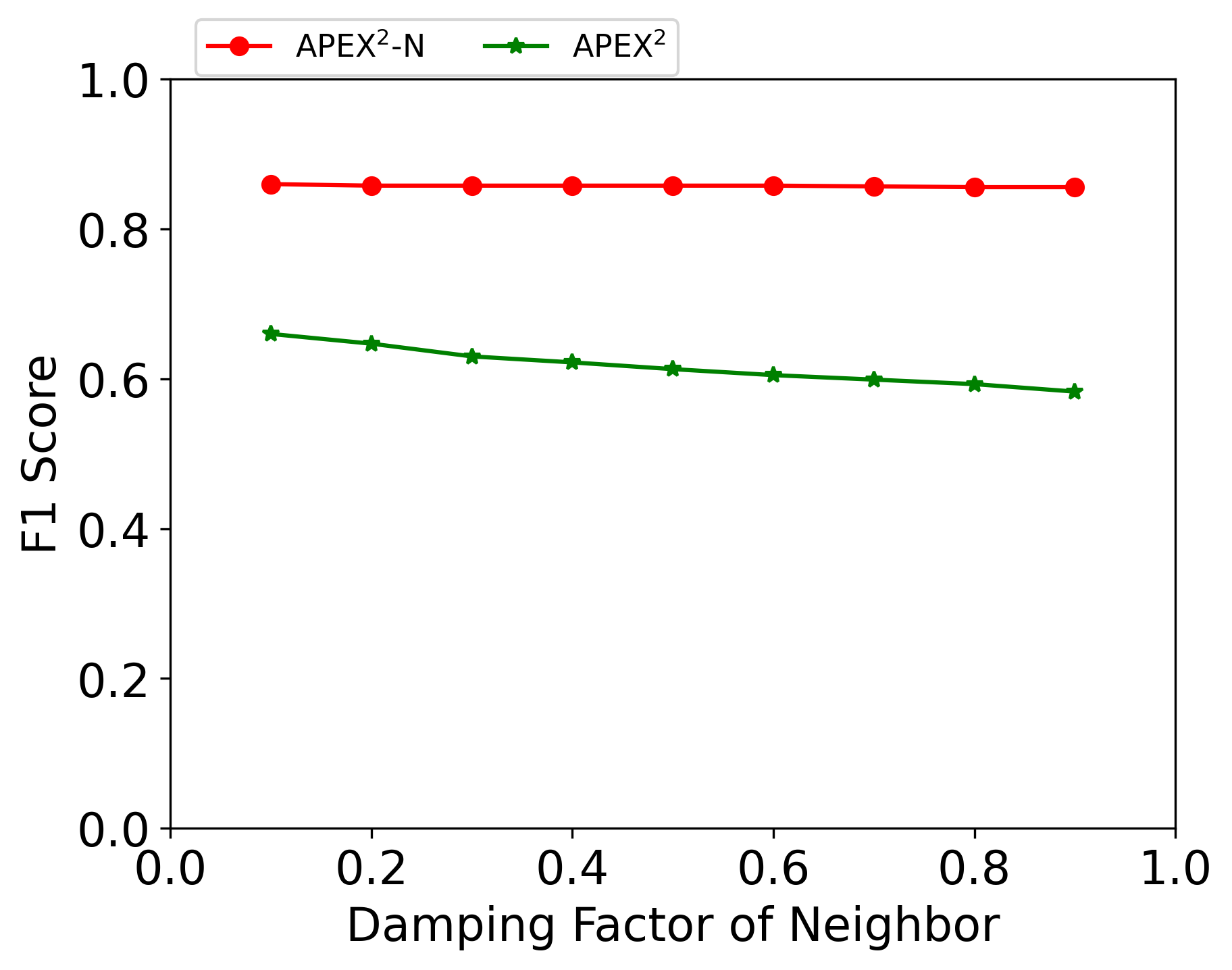}
    \end{subfigure}
    \begin{subfigure}[]{0.24\textwidth}
        \centering
        \includegraphics[width=\textwidth]{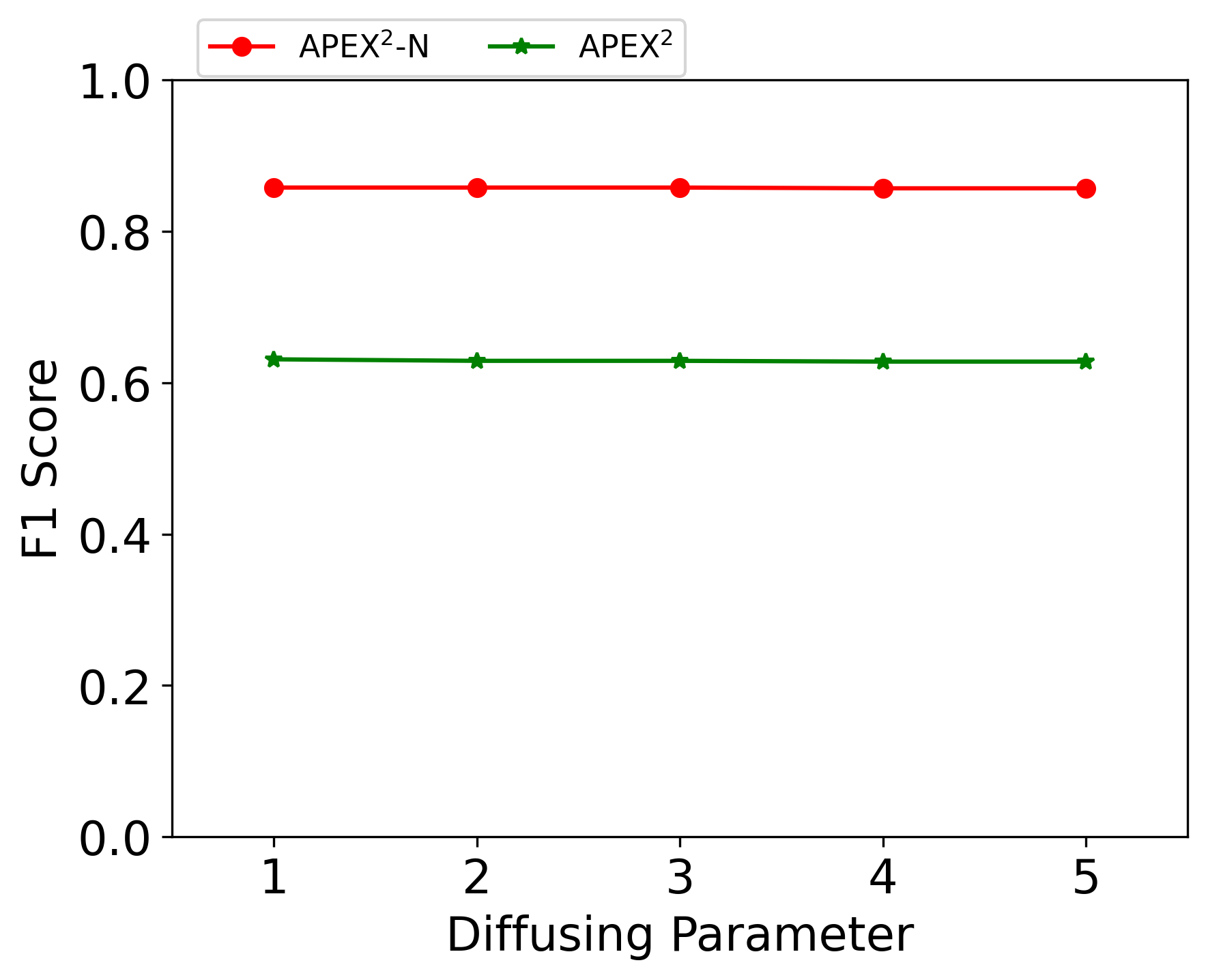}
    \end{subfigure}
    \begin{subfigure}[]{0.24\textwidth}
        \centering
        \includegraphics[width=\textwidth]{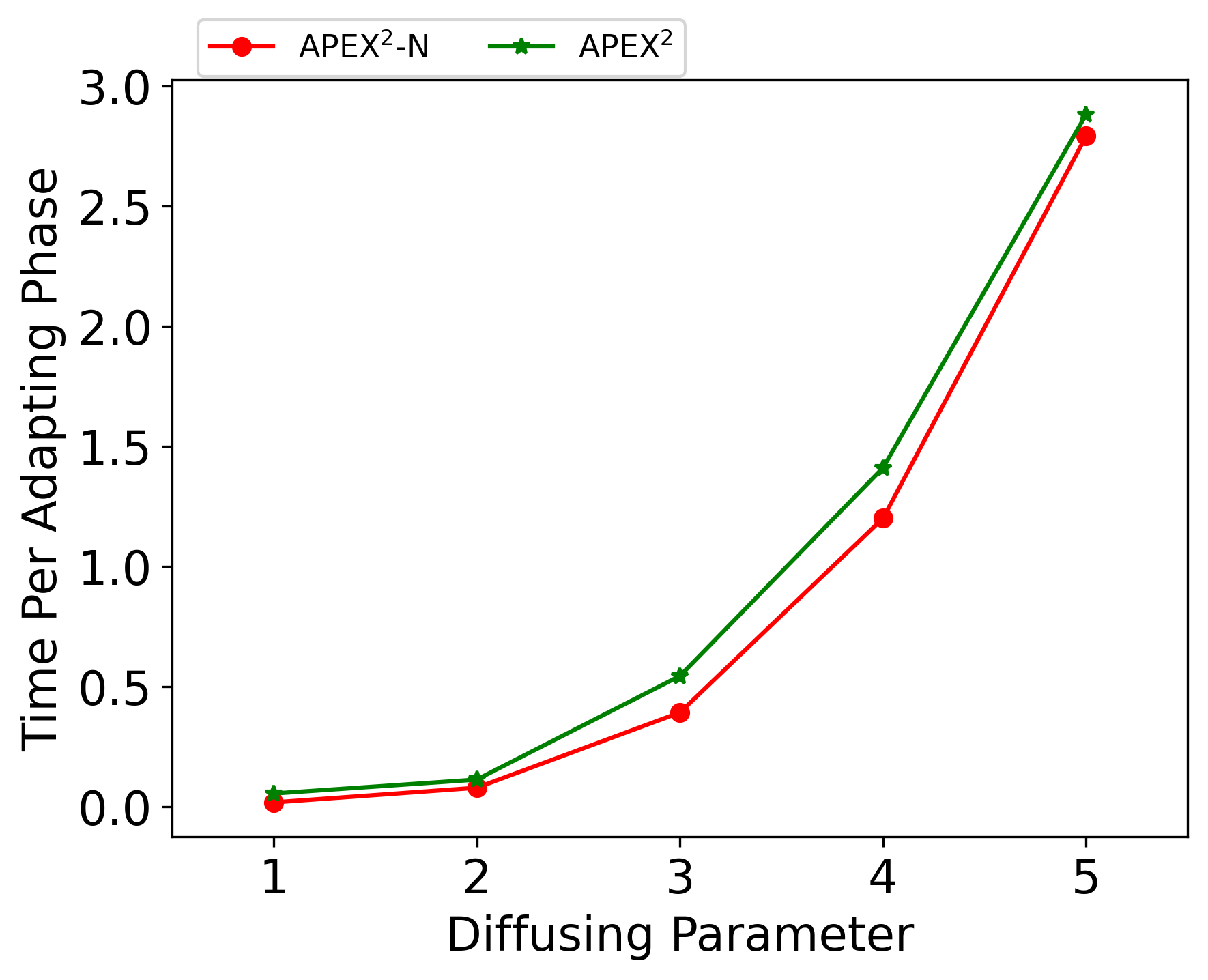}
    \end{subfigure}
    \vspace{-3mm}
    \caption{Parameter Study. From left to right: compression ratio $\kappa$, damping factor of neighbor $\alpha$, diffusing diameter $d$}
    \label{Fig: parameter_study}
    \vspace{-3mm}
\end{figure*}

\subsection{Hyperparameter Study}
We study the parameter sensitivity to show our models' robustness using MetaQA dataset. From the result in Figure \ref{Fig: parameter_study}, larger compression ratio leads to better F1 score, but after a certain value F1 score does not change much. This is intuitive as the searching accuracy increases with more triples stored in the PKG. In general, our methods are robust with damping factor and diffusing parameter. A larger diffusing diameter takes more time because more items get non-zero heat. Time per adapting phase increases quadratically with diffusing parameter, because the interested area grows with $d$.

\subsection{Handling Multiple Queries at One Time}
\label{sec: study of R_APEX}
As mentioned in Section \ref{sec: re-sum interval}, APEX$^2$ and APEX$^2$-N can re-summarize the KG every $R_{APEX}$ timestamp to increase overall efficiency. We conduct comprehensive experimental analysis on varying $R_{APEX}$ and report the results in Appendix \ref{sec: R_APEX exp} due to page limitation. In short, on the effectiveness side, our methods, with varying $R_{APEX}$ of 2, 3, 6, achieved competitive query accuracy (i.e., F1 score) and still outperformed baseline methods. On the efficiency side, the time consumed by the varying $R_{APEX}$ methods is similar to that of $R=1$. These results suggest that our heat tracking method has a robust performance over timestamps.

\subsection{Case Study}
We provide a comparative case study to illustrate how our methods work and that they can summarize high-quality PKGs. 

\subsubsection{Dataset and Query}
We do a case study on MetaQA dataset using the queries of user 0 to show in which way our methods are able to adaptively summarize highly interested items into PKG. We use the first 33 queries from the user. Most of the interested part of the KG is shown in Figure \ref{fig: case_study}. In the first 3 groups of 10 queries, the query entities are respectively the actor "Stevan Riley", the movie "The Disappearance of Haruhi Suzumiya" and the movie "LOL".

\vspace{-1mm}
\subsubsection{Settings}
We aim to study how APEX$^2$, APEX$^2$-N, GLIMPSE and PageRank deal with topic shift. Starting from the $31^{st}$ query, the user asks \textit(Chad Michael Murray, movie\_to\_actor, ?), i.e., "which movies did Chad Michael Murray act in" three times. The correct answer entities are "A Cinderella Story", "House of Wax", and "Left Behind". For APEX$^2$ and APEX$^2$-N, we use the \textbf{same} setting as in the main experiments: both of them evolve every timestamp from the beginning. For GLIMPSE and PageRank, we give them \textbf{higher} privileges that they can re-summarize \textbf{each new timestamp}, compared to per 9 timestamps in the main experiments. We use the same hyperparameters as the main experiments.

\vspace{-1mm}
\subsubsection{Results}
The PKG results are shown in Figure \ref{fig: case_study_APEX}, \ref{fig: case_study_APEX_N}, \ref{fig: case_study_glimpse}, \ref{fig: case_study_pagerank}. 
The summarized PKG at timestamp $t-1$ have been fed query $t$.
\textbf{(i)} The sub-figure (a) (i.e., the PKGs after the first three groups of queries) of GLIMPSE and PageRank still contains all information about the out-of-interest topic "Blue Blood" and "Stevan Riley", which were accessed in the first 10 queries. 
\textbf{(ii)} For both APEX$^2$ and APEX$^2$-N, after the first query on the new topic, the three answer triples (Chad Michael Murray, movie\_to\_actor, A Cinderella Story), (Chad Michael Murray, movie\_to\_actor, House of Wax), and (Chad Michael Murray, movie\_to\_actor, Left Behind) get into the PKG. However, after three times querying on the new topic, the PKG summarized by GLIMPSE only contains (Chad Michael Murray, movie\_to\_actor, House of Wax), and the PKG summarized by PageRank does not contain any of them. 
\textbf{(iii)} From APEX$^2$ and APEX$^2$-N (b) to (d), as topic on "Chad Michael Murray" is queried again and again,  the connected group centered at "Chad Michael Murray" grows larger. At timestamp 32 (Figure d), some of its 2-hop neighbors are included, such as (House of Wax, actor\_to\_movie, Nicolas Cage) in APEX$^2$ and (A Cinderella Story, director\_to\_movie, Mark Rosman) in APEX$^2$-N. The reason why the actor relationship is included first is that, this relation is recently queried many times and has a high relational interest. 
\textbf{(iv)} There are some triples that are not very related to the user queries but are summarized in the GLIMPSE PKG, for example (Onibi, movie\_to\_language, Japanese), (john lithgow, tag\_to\_movie, 2010) and (Hercules, movie\_to\_genre, Animation). The PageRank PKG almost remains the same given three queries on the new topic. Compared to these, APEX$^2$ and APEX$^2$-N produce PKGs that have intuitively higher quality in terms of user's interest.

\section{Related Work}
\label{rw}

Graph compression or graph summarization has been a popular research topic in recent years.
In 2018, Liu et al. wrote a survey \cite{DBLP:journals/csur/LiuSDK18} and provided a taxonomy for graph summarization algorithms: static plain graph summarization \cite{DBLP:conf/sdm/KoutraKVF14, DBLP:conf/sigmod/YongH0T21, DBLP:conf/kdd/LeeJKLS20}, static labeled graph summarization \cite{DBLP:conf/icdm/SafaviBFMMK19, DBLP:conf/www/BelthZVK20} and dynamic plain graph summarization \cite{DBLP:conf/kdd/ShahKZGF15, DBLP:conf/sigmod/0001CM16, DBLP:conf/kdd/KoKS20}.
Different techniques and metrics have been adopted for graph summarization. Kleindessner et al. \cite{DBLP:conf/icml/KleindessnerAM19} proposes using k-center clustering for fair data summarization; SSumM \cite{DBLP:conf/kdd/LeeJKLS20} greedily merges supernodes to minimize the reconstruction error; MoSSo \cite{DBLP:conf/kdd/KoKS20} approximates the optimal utility by random search; GraphZIP \cite{DBLP:journals/jbd/RossiZ18} focus on effective summarization for clique structures and compress the graph by decomposing it into a set of cliques; NETCONDENSE \cite{DBLP:journals/tkde/AdhikariZABP18} shrink the temporal networks by propagating and merging the unimportant node and time-pairs; 
It is worth noting that macroscopically deep-learning-based methods may not always be a good choice for large-scale graph summarization purposes. This is because a graph-scale embedding requires much larger space than the graph itself, which is already massive and needs to be summarized. However, partial embeddings might still be eligible.
Adaptive summarization could be useful in many cases where the computational resource is limited, but the original graph is large. One example topic is Neural Graph Databases, which are considered the next step in the evolution of graph databases~\cite{DBLP:journals/corr/abs-2303-14617, DBLP:journals/pvldb/ThorneYSS0L21}. Neural Graph Databases are powered by neural query embedding methods, which take large space complexity to be applied on the whole input graph~\cite{DBLP:conf/iclr/MediniCS21}. Adaptive summarization has the potential to fill this gap by adaptively summarizing the whole input graph into a domain-specific partial graph personalized to the user.

\section{Conclusion}
\label{conclusion}
In this paper, we propose APEX$^2$, the first framework for adaptive personalized knowledge graph summarization. We prove the adapting effectiveness, time complexity of APEX$^2$ and its variant APEX$^2$-N. Then we find that by adopting APEX$^2$ in a real-world real-time knowledge graph summarization scenario, much of the storage space can be saved while maintaining high searching effectiveness. We design extensive experiments to show the superiority of APEX$^2$ over baseline methods.

\section{Limitations}
Our PKG summarization technique is particularly beneficial in scenarios where (1) users anticipate challenges in communicating with the central server and loading the entire KG, (2) are concerned about their future query privacy and prefer not to send queries directly to the server, and (3) expect a large volume of future queries and want to reduce query response times.
In cases where none of these conditions apply, there may be no significant advantage to summarizing a PKG rather than querying the KG directly.
For example, if a user queries the KG infrequently, direct querying might be a better choice since querying KG directly takes less time than summarizing it. For instance, querying entire YAGO directly takes approximately 1 second, although querying summarized YAGO costs < 0.0001 second with competitive effectiveness, generating a summary takes around 6 seconds.
Making the summarization for infrequent user is interesting and challenging, we would like to explore it in the future work.

\section{Broader Impact}
In the past decades, graphs and knowledge graphs have been serving various real-world applications, from ranking \cite{DBLP:conf/nips/HeTMS12}, social network analysis~\cite{DBLP:conf/www/Fu0MCBH23, yan2021dynamic, DBLP:conf/kdd/ZhouZ0H20, DBLP:conf/aaai/ZengDZX0T24}, transportation \cite{DBLP:journals/corr/abs-2410-17576}, anomaly detection \cite{yikun2019no, DBLP:conf/icdm/ZhouWCH15, DBLP:journals/corr/abs-2409-09770, DBLP:journals/corr/abs-2409-10951, DBLP:conf/kdd/0001QZZHT24, DBLP:conf/icml/0002QZYZ0ZWHT24}, community detection \cite{li2024provably, li2024hypergraphs} and recommendation \cite{DBLP:journals/corr/abs-2411-01410, qi2023graph}, to molecular biology \cite{DBLP:conf/kdd/FuFMTH22, li2024can, DBLP:conf/cikm/ZhouZF0H22} and climate sciences \cite{DBLP:journals/corr/abs-2212-12794, DBLP:journals/corr/abs-2408-04254}. Stepping into the era of large and foundation models \cite{DBLP:conf/kdd/ZhengJLTH24, DBLP:journals/corr/abs-2410-12126}, graphs have been leveraged in Retrieval Augmented Generation (RAG) \cite{DBLP:journals/corr/abs-2312-10997, DBLP:journals/corr/abs-2404-16130, DBLP:journals/corr/abs-2410-05983}, which enhances the retrieval and integration of relevant context for improving the quality and relevance of generated responses, and knowledge graph personalization still holds immense potential for enabling more accurate, context-aware, and adaptive decision-making by integrating domain-specific knowledge and leveraging the scalability and representation power of these advanced models. Notably, recent works on personalization \cite{DBLP:conf/sigir/Braga24, he2024llm, DBLP:journals/corr/abs-2402-09269} and compression \cite{DBLP:conf/iclr/0005AC0H24, DBLP:conf/emnlp/ZouZLH024} of Large Language Models (LLMs) underscore the significance of enhancing adaptability, efficiency, and user-specific customization to better meet diverse application needs.

\clearpage
\bibliographystyle{ACM-Reference-Format}
\bibliography{reference.bib}

\appendix
\clearpage

\section{Pseudo-code of Algorithms}
\subsection{GLIMPSE}
\vspace{-4mm}
\begin{algorithm}
    \caption{The GLIMPSE framework}
    \begin{algorithmic}[1]
        \REQUIRE knowledge graph $\mathcal{G}$; query log $\mathcal{Q}$; size budget $K$
        \ENSURE personal summarization $\mathcal{P} \subseteq \mathcal{G}$ with $|\mathcal{T}_p| \leq K$
        \STATE Compute $\mathcal{T}^{\Delta \neq 0}$ with Pr$(e|\mathcal{Q})$, Pr$(x_{ijk}|\mathcal{Q})$
        \STATE $\mathcal{P} \leftarrow \emptyset$
        \WHILE{$|\mathcal{T}_p| \leq K$}{
            \STATE Sample a set $\mathcal{S}$ of size $\frac{|\mathcal{T}^{\Delta \neq 0}}{K} \log\frac{1}{\epsilon}$ from $T^{\Delta \neq 0}$
            \STATE Select $\widetilde{x}_{ijk} \leftarrow \arg \max_{x_{ijk} \in \mathcal{S}}\Delta_\delta(x_{ijk}|\mathcal{P}, \mathcal{Q})$ 
            \STATE Add triple $\widetilde{x}_{ijk} = (e_i, r_k, e_j)$ to $\mathcal{P}$
        }        
        \ENDWHILE
        \RETURN $\mathcal{P}$
    \end{algorithmic}
\label{AG: GLIMPSE}
\end{algorithm}
\vspace{-5mm}

\subsection{PEGASUS}
\vspace{-4mm}
\begin{algorithm}
    \caption{The PEGASUS framework}
    \begin{algorithmic}[1]
        \REQUIRE input graph $\mathcal{G} = (\mathcal{V},  \mathcal{E})$; size budget $K$; target node set $\mathcal{T}$; degree of personalization $\alpha$; parameter for adaptive thresholding $\beta$; max number of iterations $t_{max}$ 
        \ENSURE personal summary $\mathcal{P} = (\mathcal{V}_{P}, \mathcal{E}_{P})$ within size budget $K$
        \STATE $\mathcal{V}_{P} \leftarrow \{\{u\}: u \in \mathcal{V}\}$; $\mathcal{E}_{P} \leftarrow \{\{\{u\}, \{v\}\}:\{u, v\} \in \mathcal{E} \} $
        \STATE $t \leftarrow 1; \theta \leftarrow 0.5; \mathcal{L} \leftarrow []$
        \WHILE{$t \leq t_{max} $ and $Size(\mathcal{P} ) > K$}{
            \STATE $C \leftarrow$ generate candidate groups
            \FOR{\textbf{each} group $C_i \in C$}{
                \STATE Greedily merge nodes in $C_i$ with the threshold $\theta$; update $\mathcal{V}_{P}, \mathcal{E}_{P}, \mathcal{L}$
            }
            \ENDFOR
            \STATE $\theta \leftarrow [\beta \times |\mathcal{L}|]$-th largest entry in $\mathcal{L}$
            \STATE $\mathcal{L} \leftarrow []; t \leftarrow t+1$
        }        
        \ENDWHILE
        \IF{$Size(\mathcal{P}) > K$}
            \STATE Sparsify $\mathcal{P}$ further
        \ENDIF
        \RETURN $\mathcal{P}$
    \end{algorithmic}
\label{AG: PEGASUS}
\end{algorithm}
\vspace{-5mm}

\subsection{Incremental Binary Insertion Sort}
\vspace{-4mm}
\begin{algorithm}
    \caption{Incremental Binary Insertion Sort}
    \begin{algorithmic}[1]
        \REQUIRE previously sorted sortable instance $\mathcal{S}$, set of changes $\mathcal{C}$
        \ENSURE sorted instance $\mathcal{S}$ that is updated as described in $\mathcal{C}$
        \\/* extract unchanged entries in $\mathcal{S}$ by deleting */  
        \FOR{\textit{(from, to)} in $\mathcal{C}$}
            \IF {\textit{from} is not None}
                \STATE pos = binary\_search(\textit{from}, $\mathcal{S}$)
                \STATE delete($\mathcal{S}$, pos)
            \ENDIF
        \ENDFOR
        \FOR{\textit{(from, to)} in $C$}
            \IF {\textit{to} is not None}
                \STATE binary\_insert(\textit{to}, $\mathcal{S}$)
            \ENDIF
        \ENDFOR
        \RETURN $\mathcal{S}$ 
    \end{algorithmic}
\label{AG: IBIS}
\end{algorithm}
\vspace{-5mm}

\vspace{5mm}
\subsection{APEX$^2$}
\begin{algorithm}[h]
    \caption{APEX$^2$ framework}
    \begin{algorithmic}[1]
        \REQUIRE Knowledge Graph $\mathcal{G} = (\mathcal{E}, \mathcal{R}, \mathcal{T})$; (temporal) user query log $\mathcal{Q}^{(t)}$ on triples; \# triples (size budget) $K$; decay factor $\gamma$; diffuse diameter $d$
        \ENSURE (temporal) Personal summary $\mathcal{P} = (\mathcal{E}_p^{(t)}, \mathcal{R}_p^{(t)}, \mathcal{T}_p^{(t)})\subseteq \mathcal{G}$ with $|\mathcal{T}_p| \leq K$
        \\/* Initializing Phase */
        \STATE $\mathbf{H} \leftarrow |\mathcal{E}|\times|\mathcal{R}|\times|\mathcal{E}|$ Sparse Matrix with 0s, $\mathcal{E}_p^{(0)} \leftarrow \emptyset$, $\mathcal{R}_p^{(0)} \leftarrow \emptyset$, $\mathcal{T}_p^{(0)} \leftarrow \emptyset$
        \FOR{$i = 0$; $i < d; i\texttt{++}$}{    
            \STATE Calculate ($\alpha\mathbf{A})^{l}$ and $\sum_{l = 0}^{d}\alpha^l\mathbf{A}^l$
        }
        \ENDFOR
        \STATE Calculate $\mathbf{q}_{\rm total}^{(0)}$, $\mathbf{e}^{(0)}$, $\mathbf{r}^{(0)}$, $\mathbf{H}^{(0)}$ by equation \ref{q_total^T}, \ref{e^T}, \ref{r^T} and \ref{H^T}
        \STATE Choose triples with top-$K$ highest heat to construct $\mathcal{T}_p^{(0)}$
        \STATE $\mathcal{E}_p^{(0)} \leftarrow \{e \in \mathcal{E} \quad {\rm s.t.} \quad \exists x \in \mathcal{T}_p^{(0)}, e \in x\}$ and $\mathcal{R}_p^{(0)} \leftarrow \{r \in \mathcal{R} \quad {\rm s.t.} \quad \exists x \in \mathcal{T}_p^{(0)}, r \in x\}$
        \\/* Adapting Phase */  
            \FOR{timestamp $t = 1$; $t \leq T$; $t\texttt{++}$}{
            \STATE Decay nonzero elements in $\mathbf{H}$ with $\gamma^3$   
            \STATE Incrementally update $\mathbf{q}_{\rm total}$, $\mathbf{e}$, $\mathbf{r}$
            \STATE Recalculate entry $\mathbf{H}[i][j][k]$ if $\mathbf{e}[i], \mathbf{r}[j]$ or $\mathbf{e}[k]$ is changed. Meanwhile construct $\mathcal{C}$
            \STATE Incrementally sort $\mathbf{H}$ using $\mathcal{C}$
            \STATE Choose triples with top-$K$ highest heat to construct $\mathcal{T}_p^{(t)}$
            \STATE $\mathcal{E}_p^{(t)} \leftarrow \{e \in \mathcal{E} \quad {\rm s.t.} \quad \exists x \in \mathcal{T}_p^{(t)}, e \in x\}$ and $\mathcal{R}_p^{(t)} \leftarrow \{r \in \mathcal{R} \quad {\rm s.t.} \quad \exists x \in \mathcal{T}_p^{(t)}, r \in x\}$
        }
        \ENDFOR
        \RETURN \{$\mathcal{P}^{(t)}$\}
    \end{algorithmic}
\label{AG: APEX}
\end{algorithm}

\newpage
\vspace{5mm}
\subsection{APEX$^2$-N}

\begin{algorithm}
    \caption{APEX$^2$-N Framework}
    \label{AG: APEX-N}
    \begin{algorithmic}[1]
        \REQUIRE Knowledge Graph $\mathcal{G} = (\mathcal{E}, \mathcal{R}, \mathcal{T})$; (temporal) user query log $\mathcal{Q}^{(t)}$; \# size budget $K$; decay factor $\gamma$; diffuse diameter $d$
        \ENSURE (temporal) Personal summary $\mathcal{P} = (\mathcal{E}_p^{(t)}, \mathcal{R}_p^{(t)}, \mathcal{T}_p^{(t)})\subseteq \mathcal{G}$ with $|\mathcal{T}_p| \leq K$
        \\/* Initializing Phase */
        \STATE $\mathbf{H} \leftarrow |\mathcal{E}|\times|\mathcal{E}|$ Sparse Matrix with 0s, $\mathcal{E}_p^{(0)} \leftarrow \emptyset$, $\mathcal{R}_p^{(0)} \leftarrow \emptyset$, $\mathcal{T}_p^{(0)} \leftarrow \emptyset$
        \FOR{$q \in \mathcal{Q}^{(0)}$}{    
            \STATE HeatDiffuse$(\mathbf{H}, \mathcal{G}, q, d)$
        }
        \ENDFOR
        \STATE sort non-zero elements in $\mathbf{H}$
        \STATE choose entities $e$ with top-$K$ highest heat to construct $\mathcal{E}_p^{(0)}$
        \STATE $\mathcal{T}_p^{(0)} \leftarrow \{x_{ijk} \in \mathcal{T} \quad {\rm s.t.} \quad i,j \in \mathcal{E}_p^{(0)}, v \in e\}$
        \STATE $\mathcal{R}_p^{(0)} \leftarrow \{r \in \mathcal{R} \quad {\rm s.t.} \quad \exists x_{ijk} \in \mathcal{T}_p^{(0)}, r \in x_{ijk}\}$
        \\/* Updating Phase */   
        \FOR{timestamp $t = 1$; $t \leq T$; $t\texttt{++}$}{
            \STATE decay nonzero elements in $\mathbf{H}$ with $\gamma$   
            \STATE $\mathcal{C}$ = HeatDiffuse$(\mathbf{H}, \mathcal{G}, \mathcal{Q}^{(t)}\setminus\mathcal{Q}^{(t-1)}, d)$
            \STATE incrementally sort $\mathbf{H}$ using $\mathcal{C}$
            \WHILE {$|\mathcal{T}_p^{(t)}| \leq K$}
            \STATE add $v$ with highest heat to $\mathcal{E}_p^{(t)}$
            \STATE $\mathcal{T}_p^{(t)} \leftarrow \{x_{ijk} \in \mathcal{T} \quad {\rm s.t.} \quad i,j \in \mathcal{E}_p^{(t)}\}$
            \STATE $\mathcal{R}_p^{(t)} \leftarrow \{r \in \mathcal{R} \quad {\rm s.t.} \quad \exists x_{ijk} \in \mathcal{T}_p^{(t)}, r \in x_{ijk}\}$
            \ENDWHILE
        }
        \ENDFOR
        \RETURN \{$\mathcal{P}^{(t)}$\}
    \end{algorithmic}
\end{algorithm}

\section{Analysis of Existing Methods}
\label{non-adaptability analysis}

In this section, we introduce two pioneering solutions on personalized knowledge graph summarization and analyze their adaptability for evolving user query interests, which motivate our adaptive PKG summarization framework, APEX$^2$, proposed in Section \ref{'APEX for Adaptive PKG Summarization'}.

\subsection{Preliminary}

\subsubsection{GLIMPSE}
GLIMPSE~\cite{DBLP:conf/icdm/SafaviBFMMK19} is a sampling-based method to summarize personalized KG. It first infers user preferences over the KG, then constructs a personal summary by maximizing an inferred utility by sampling. Denoting an entity $e$'s neighbor set as $N(e)$. Then for a query log $\mathcal{Q}$, GLIMPSE captures the user's preference as
\begin{equation}
\begin{split}
    {\rm Pr}(e| \mathcal{Q}) \propto \sum_{q\in\mathcal{Q}} (\mathbbm{1}(e \in q) + \alpha \sum_{e_o \in N(e)} \mathbbm{1}(e_o \in q))
\end{split}
\end{equation}
where ${\rm Pr}(x|\mathcal{Q})$ stands for speculative preference on $x$ given the query log, $\mathbbm{1}$ is an indicator function, i.e., $\mathbbm{1}(\mathcal{X}) = 1 \iff \mathcal{X} = \textit{True}$, and $\alpha$ denotes the damping factor of neighbors $N(e)$ in the given knowledge graph $\mathcal{G}$. This equation assumes user's preference is more on the searched entity and can be generally pushed to its neighboring entities. The preference for relationship is defined by the query frequency, normalized by the total number of queries,
\begin{equation}
\begin{split}
    {\rm Pr}(r|\mathcal{Q}) \propto \frac{\sum_{q\in\mathcal{Q}} (\mathbbm{1}(r \in q))}{|\mathcal{Q}|}
\end{split}
\label{rk_def}
\end{equation}

Then, following the standard conditional independence assumption~\cite{DBLP:conf/kdd/GroverL16, DBLP:journals/pieee/Nickel0TG16}, GLIMPSE assumes user's preference for a triple to be proportional to a multiplication form as
\begin{equation}
\begin{split}
    {\rm Pr}(x_{ijk}|\mathcal{Q}) \propto {\rm Pr}(e_i|\mathcal{Q}){\rm Pr}(r_k|\mathcal{Q}){\rm Pr}(e_j|\mathcal{Q})
\end{split}
\label{xxjk_calc}
\end{equation}

After inferring user's preference, GLIMPSE constructs the PKG $\mathcal{P} = (\mathcal{E}_p, \mathcal{R}_p, \mathcal{T}_p)$ by maximizing the following utility,
\begin{equation}
\begin{split}
    \phi(\mathcal{P}, \mathcal{Q}) \propto \sum_{e\in\mathcal{E}_p} \log {\rm Pr}(e|\mathcal{Q}) + \sum_{x_{ijk} \in \mathcal{T}_p} \log {\rm Pr}(x_{ijk}|\mathcal{Q})
\end{split}
\label{phiPQ}
\end{equation}

In the optimizing phase, GLIMPSE continuously calculates the marginal utility of triples $x_{ijk}$ and greedily samples triples with the highest marginal utilities\footnote{Generally, $\Delta_F(x|\mathcal{S}) = F(\mathcal{S} \cup \{x\}) - F(\mathcal{S})$ is the marginal utility gained in the set function $F$ by adding $x$ to $\mathcal{S}$. Here $\Delta_\phi(x_{ijk}|\mathcal{P},\mathcal{Q}) = \phi({x_{ijk}} \cup \mathcal{P}, \mathcal{Q}) - \phi(\mathcal{P}, \mathcal{Q})$} $\Delta_\phi(x_{ijk}|\mathcal{P},\mathcal{Q})$,
\begin{equation}
\begin{split}
    \mathcal{T}^{\Delta \neq 0} \triangleq \{x_{ijk} \in \mathcal{G} \quad {\rm s.t.} \quad \Delta_\phi(x_{ijk}|\mathcal{P},\mathcal{Q}) \neq 0\}
\end{split}
\end{equation}
where $\mathcal{T}^{\Delta \neq 0}$ is the set of triples with nonzero marginal utility for any given KG $\mathcal{G}$ and the corresponding PKG $\mathcal{P}$. ``$\triangleq$'' means ``defined as''. The sampling process stops when the number of triples in the summarization reaches the restriction or bound of size. The overall GLIMPSE framework is shown in Algorithm~\ref{AG: GLIMPSE}.



\subsubsection{PEGASUS}
PEGASUS~\cite{DBLP:conf/icde/KangLS22} is a merging-based method for summarizing personalized graphs (not specific to knowledge graphs). It determines how to merge supernodes and superedges by minimizing the reconstruction error, during which more consideration will be put on a given set of targeted nodes.
Formally, given a graph $\mathcal{G} = (\mathcal{V}, \mathcal{E})$, a set of target nodes $\mathcal{T}$, and a space budget $k$, PEGASUS aims to find a summarized graph $\mathcal{P} = (\mathcal{V}_{P}, \mathcal{E}_{P})$ of $\mathcal{G}$ that is personalized to $\mathcal{T}$ while satisfying the budget $k$.
The general logic is that attributes (edge connectivities) near the target set $\mathcal{T}$ are more likely to be retained in $\mathcal{P}$ than those far from the target set. The optimization of PEGASUS is based on the weighted reconstruction error $RE^{(\mathcal{T})}$,
\begin{equation}
\begin{split}
    RE^{(\mathcal{T})}(\mathcal{P}) = \sum_{i=1}^{|\mathcal{V}|}\sum_{j=1}^{|\mathcal{V}|}\bm{W}_{ij}^{(\mathcal{T})}|\bm{A}_{ij}^{(\mathcal{G})} - \bm{A}_{ij}^{(\hat{\mathcal{G}})}|
\end{split}
\end{equation}
where $\hat{\mathcal{G}} = (\mathcal{V}, \hat{\mathcal{E}})$ is the reconstructed graph from $\mathcal{P}$. $\bm{A}_{ij}$ stands for the adjacency matrix. $\mathcal{T}$ is the given target node set as interests. $\bm{W}_{ij}^{\mathcal{(T)}}$ is the personalized weight of each pair of nodes $(i, j)$. $\bm{W}_{ij}^{\mathcal{(T)}}$ depends on the pairwise distance to the target nodes,
\begin{equation}
\begin{split}
    W_{ij}^{(\mathcal{T})} = \frac{\alpha^{-(D(i, \mathcal{T})+D(j, \mathcal{T}))}}{Z}
\end{split}
\label{Wuv}
\end{equation}
where $\alpha > 1$ controls the degree of personalization, $Z$ is the constant that makes the average weight 1, and
\begin{equation}
\begin{split}
    D(u, \mathcal{T}) = \min_{t \in \mathcal{T}} \# hops(u, t)
\end{split}
\label{Dut}
\end{equation}
is the minimum number of hops between node $u$ and any node in $\mathcal{T}$. The overall PEGASUS framework is shown in Algorithm \ref{AG: PEGASUS}.

\subsubsection{iSummary} iSummary \cite{DBLP:conf/esws/VassiliouAPK23} summarizes knowledge graphs according to weights and seed nodes provided by users. iSummary tries to find a PKG by solving an NP-complete Steiner tree problem. The authors formulate the $\lambda/\kappa$-Personalized Summary problem as the following. Given (1) a knowledge graph $G=(V, E)$; (2) a non-negative weight assignment to all nodes representing user preferences; (3) $\lambda$ seed nodes; (4) a number $\kappa$ ($\lambda \leq \kappa$). The $\lambda/\kappa$-Personalized Summary problem aims to find the smallest maximum-weight tree $G' = (V', E') \in G$ that includes the $\kappa$ most preferred nodes. In the original iSummary paper \cite{DBLP:conf/esws/VassiliouAPK23}, the form of KG and the summarization problem is different from ours and the seminal work \cite{DBLP:conf/icdm/SafaviBFMMK19}.

Given a SPARQL query log, iSummary works by first determining which entity the users are interested in the most (the $\lambda$ seed nodes), then finding the shortest paths connecting them and including more facts in the summary. The constructed PKG is a maximum-weight tree that includes the $\kappa$ most preferred nodes, which means the storage cost is much more than the cost of $\kappa$ nodes themselves. We convert our scenario to the $\lambda/\kappa$ Personalized Summary problem as the following. (1) we use our knowledge graphs $\mathcal{G} = (\mathcal{E}, \mathcal{R}, \mathcal{T})$; (2) we use the heat on nodes as the user preferences; (3) we use the explicitly queried nodes as seed nodes, the number of which is $\lambda$; (4) we sweep $\kappa$ from $\lambda$ so that the summarized PKG from iSummary satisfies the storage limitation. By the time this paper is submitted, the source code of iSummary hasn't been open-sourced yet\footnote{\url{https://anonymous.4open.science/r/iSummary-47F2/}}. Therefore, we implemented iSummary on our own based on the pseudo-code \cite{DBLP:conf/esws/VassiliouAPK23} provided in the original paper.

\subsection{Proof of non-adaptability or corruption} Here, we show that neither GLIMPSE~\cite{DBLP:conf/icdm/SafaviBFMMK19} or PEGASUS~\cite{DBLP:conf/icde/KangLS22} can deal with interest-evolving user queries very well.
Our discussions here are under the circumstance that the summarized PKG is already full.
Due to the page limit, we briefly introduce the proof idea and flow. The detailed and formal proof is placed in the appendix.

For GLIMPSE~\cite{DBLP:conf/icdm/SafaviBFMMK19}, the triples with higher utility will be more likely to be included in the summarization. We show that once the user's interest has shifted from a previous topic to a new topic, GLIMPSE can adapt to new interest only if the volume of new queries is considerably large, with respect to the volume of queries on previous interests. From this perspective, GLIMPSE is not very agile for quick and ad-hoc new interest queries.

\begin{theorem}[Non-adaptability of GLIMPSE in PKG Summarization]
In adaptive PKG summarization setting, GLIMPSE could not swiftly adapt to user's new interest after the previous summarized PKG reaches the size budget. (Proof in Appendix \ref{AP: A.1})
\label{THM: GLIMPSE}
\end{theorem}

For PEGASUS~\cite{DBLP:conf/icde/KangLS22}, according to Eq.~\ref{Wuv} and Eq.~\ref{Dut}, a historical target node $j$ permanently gives a lower bound to all $W_{ij}$. In other words, if a node $i$ is close to a target node $j$, node $i$ (as well as the nodes that are close to node $i$) will be given high weights. Then, we show that once these nodes (e.g., $i$ and its $k$-hop neighbors) gets high weights and are considered important, they are hard to be replaced by the nodes of later new interests. This means that though the user's interest may already shift to other topics, previous topics still have high weights and may occupy the summary storage.

\begin{theorem}[Non-adaptability of PEGASUS in PKG Summarization]
In adaptive PKG summarization setting, PEGASUS could not effectively evolve with user's new interest after the previous summarized PKG reaches the size budget. (Proof in Appendix \ref{AP: A.2})
\label{THM: PEGASUS}
\end{theorem}



For iSummary \cite{DBLP:conf/esws/VassiliouAPK23}, under the extremely small storage limitation in our experiment scenario, even if we set $\kappa = \lambda$, the summarized PKG is still too large in terms of size. In this case, even though iSummary finds the shortest paths, those paths will not be used, and the PKG simply stores the explicitly searched nodes that have the highest heat. In the extreme summarization case, iSummary becomes a simple caching algorithm and does not have the ability to infer the user's interests. In other words, iSummary corrupts under extremely small storage constraints since $\lambda > \kappa$, i.e., the number of seed nodes exceeds the maximum number of nodes to summarize.

\section{Experiment Details}
\subsection{Metric: F1 Score}
\label{AP: F1Score}
For a query $q$ with query entity $e_i$, query relation $r_k$ and a set of answers $\mathcal{A}$, we define the answer sets in the KG $\mathcal{G}$ and PKG $\mathcal{P}$ are, respectively, $\mathcal{A}_{kg} = \{e_j \ s.t.\ \exists x_{ijk} = (e_i, r_k, e_j) \in \mathcal{T}\}$ and $\mathcal{A}_{pkg} = \{e_j \ s.t.\ \exists x_{ijk} = (e_i, r_k, e_j) \in \mathcal{T}_p\}$. Then we have 
\begin{equation}
    \textit{TP(True Positive)} = |\mathcal{A}_{pkg} \cap \mathcal{A}_{kg}|
\end{equation}

\begin{equation}
    \textit{FP(False Positive)} = |\mathcal{A}_{pkg} \setminus \mathcal{A}_{kg}|
\end{equation}

\begin{equation}
    \textit{FN(False Negative)} = |\mathcal{A}_{kg} \setminus \mathcal{A}_{pkg}|
\end{equation}

\begin{equation}
    \textit{precision} = \frac{TP}{TP + FP},\ \textit{recall} = \frac{TP}{TP + FN}
\end{equation}

\begin{equation}
    \textit{F1} = \frac{2 \times \textit{precision} \times \textit{recall}}{\textit{precision} + \textit{recall}}
\end{equation}

"Precision" calculates the fraction of relevant instances among the summarized instances, while "Recall" is the fraction of relevant instances that were summarized\footnote{Resource: \url{https://en.wikipedia.org/wiki/Precision_and_recall}}. Taking both of them into account, F1 score provides a single number that reflects the model's overall search performance.

In our adaptive setting, we adapt historical PKG to the new query, wishing to make the adaption accurate. In this way, the very next query is the most valuable to test the utility of the current PKG, as future queries may shift again. Moreover, we report the mean and std over multiple single next queries, showing the robustness.

\begin{figure*}[t]
    \centering
    \includegraphics[width=0.9\textwidth]{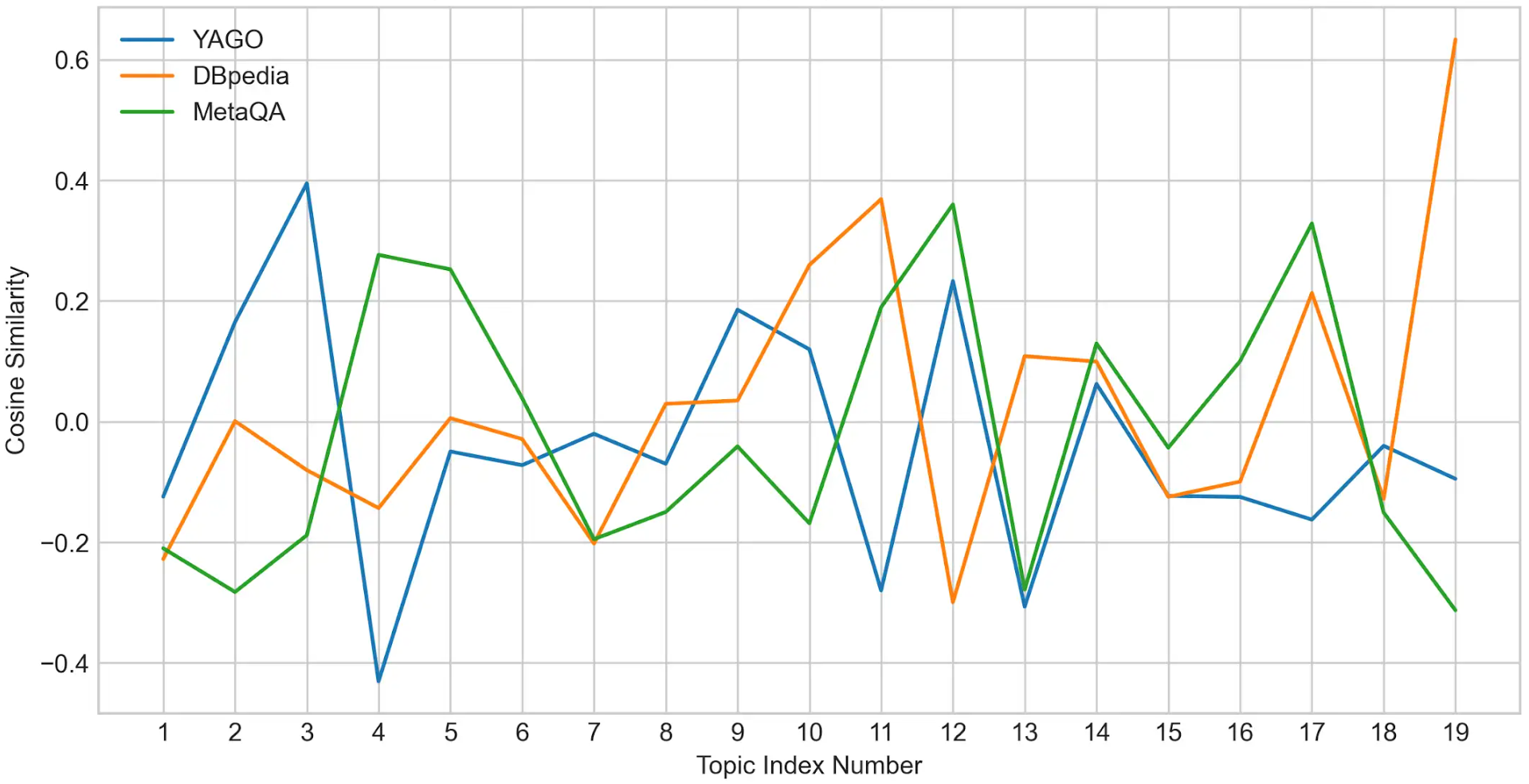}
    \caption{Cosine similarity between consecutive query topics in different datasets for one user. The cosine similarity is computed from Roberta \cite{DBLP:journals/corr/abs-1907-11692} text embeddings.}
    \label{fig: topic_sim_one_user}
\end{figure*}

\subsection{Datasets}
\label{AP: Datasets}

\textbf{YAGO3} is a huge semantic knowledge base, derived from Wikipedia WordNet and GeoNames. Currently, the whole YAGO3 has knowledge of more than 10 million entities and contains more than 120 million facts about these entities\footnote{Resource: \url{https://datahub.io/collections/yago}}. The whole YAGO3 is over 200GB. Due to the computational resource limitation, we use the \textbf{core YAGO3 facts} that contains 4.2 million entities and 12.4 million triples, which is a 40\% sub-YAGO3.

\textbf{DBpedia} was created by extracting semantically-structured information from Wikipedia and other data sources. Specifically, we use the \textbf{DBpedia ontology}, which is the "heart of DBpedia"\footnote{Resource: \url{https://www.dbpedia.org/resources/ontology/}}. The DBpedia Ontology is a shallow, cross-domain ontology, which has been manually created based on the most commonly used infoboxes within Wikipedia\footnote{Resource: \url{http://wikidata.dbpedia.org/services-resources/ontology}}.

\textbf{MetaQA} consists of a movie ontology derived from the WikiMovies Dataset\footnote{Resource: \url{https://paperswithcode.com/dataset/metaqa}}. WikiMovies dataset, introduced by Miller et al.\cite{DBLP:conf/emnlp/MillerFDKBW16}, is constructed along with a graph-based KB consisting of entities and relations, with the guarantee that each query can be answered by the KB. We use the \textbf{whole MetaQA} dataset and the "Vanilla" branch of its query data.

\textbf{Freebase} is a large collaborative knowledge base created by the Freebase community. We use FB15K237, a well-developed and commonly used sub-knowledge-graph of the originally retired Freebase data dump\footnote{Resource: \url{https://paperswithcode.com/dataset/fb15k-237}}.

\subsection{Quality Analysis of Synthetic Queries}
\label{AP: Queries}

\begin{figure*}[t]
    \centering
    \begin{subfigure}[]{0.32\textwidth}
        \centering
        \includegraphics[width=\textwidth]{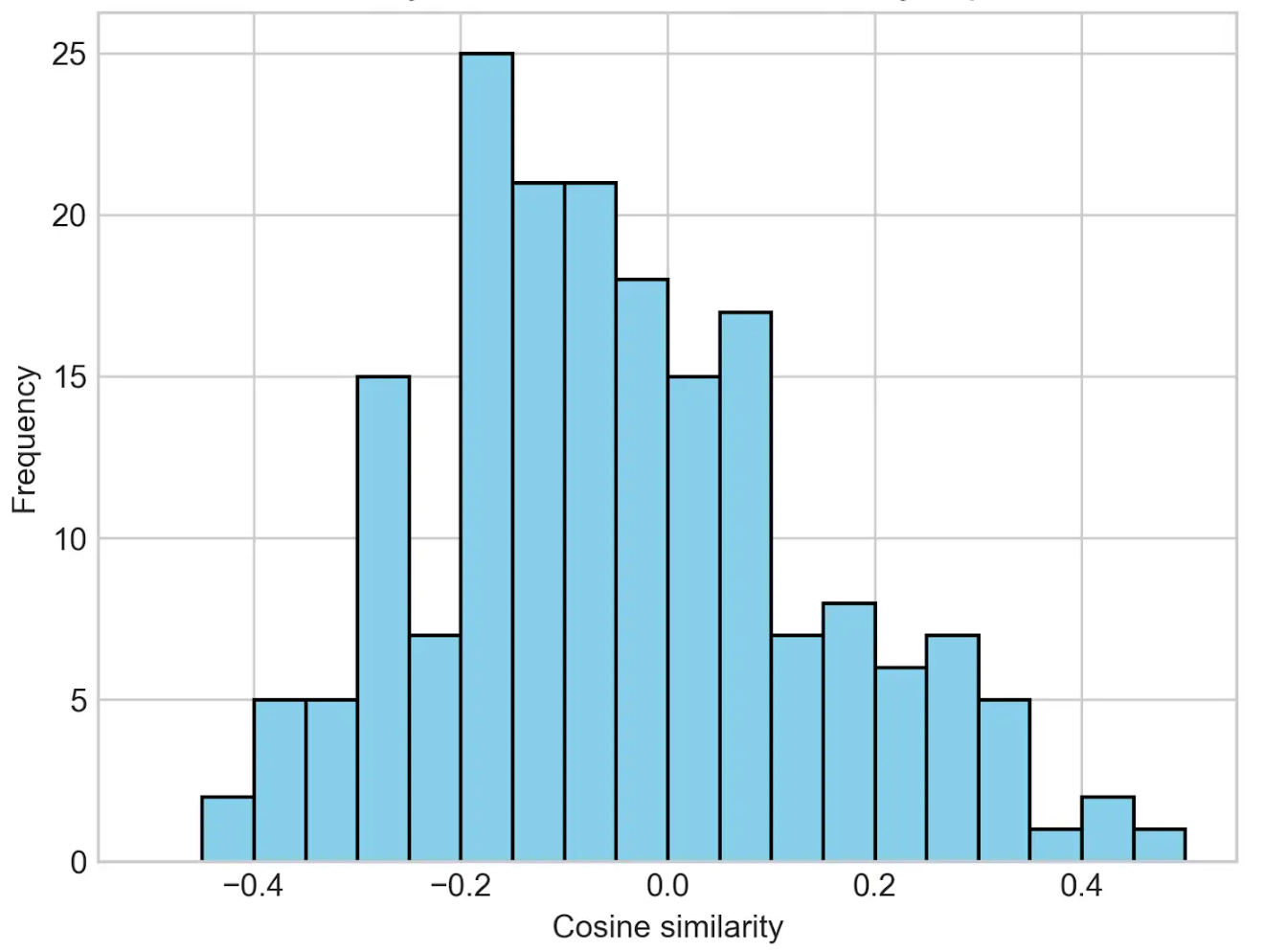}
        \caption{YAGO}
    \end{subfigure}
    \begin{subfigure}[]{0.32\textwidth}
        \centering
        \includegraphics[width=\textwidth]{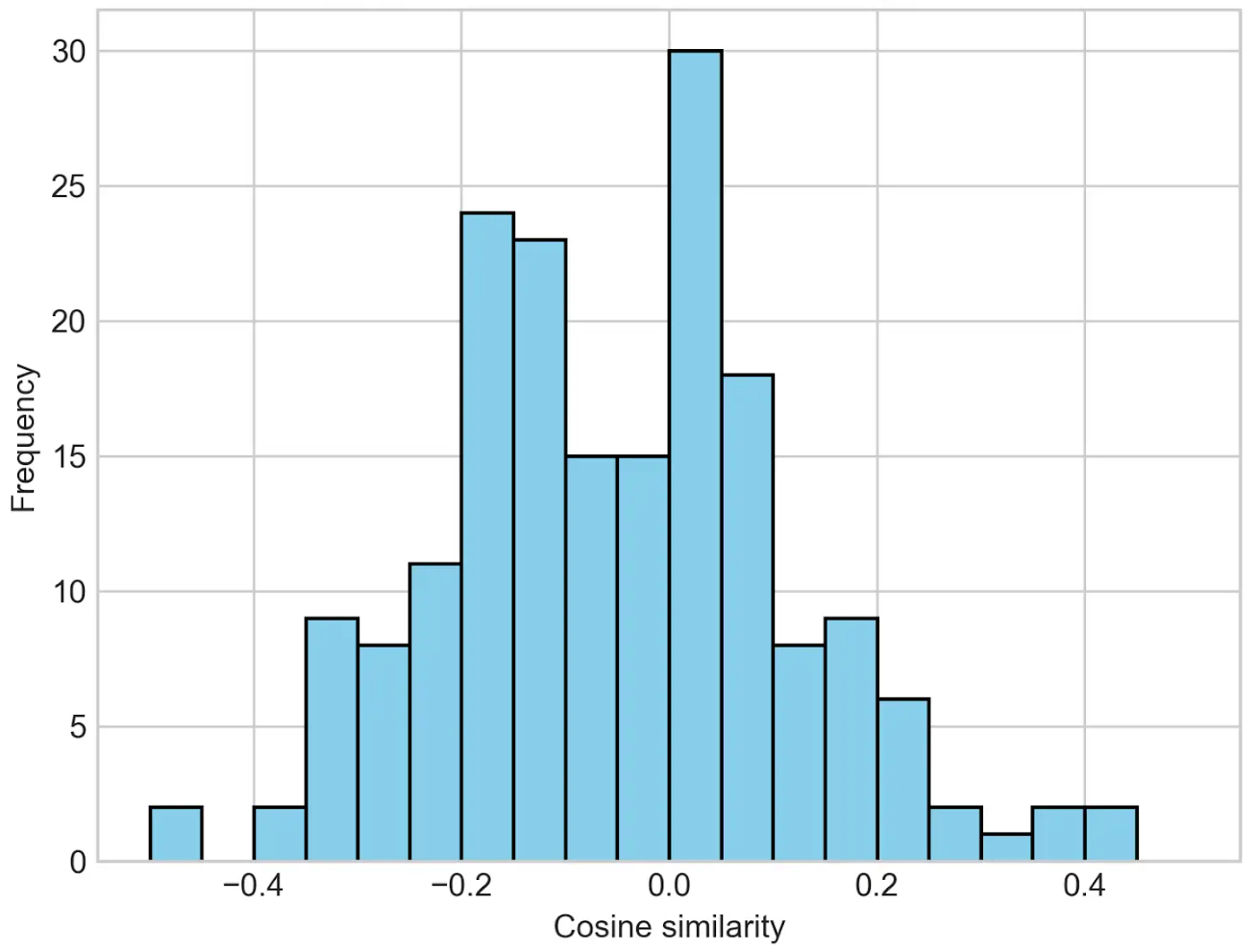}
        \caption{DBpedia}
    \end{subfigure}
    \begin{subfigure}[]{0.32\textwidth}
        \centering
        \includegraphics[width=\textwidth]{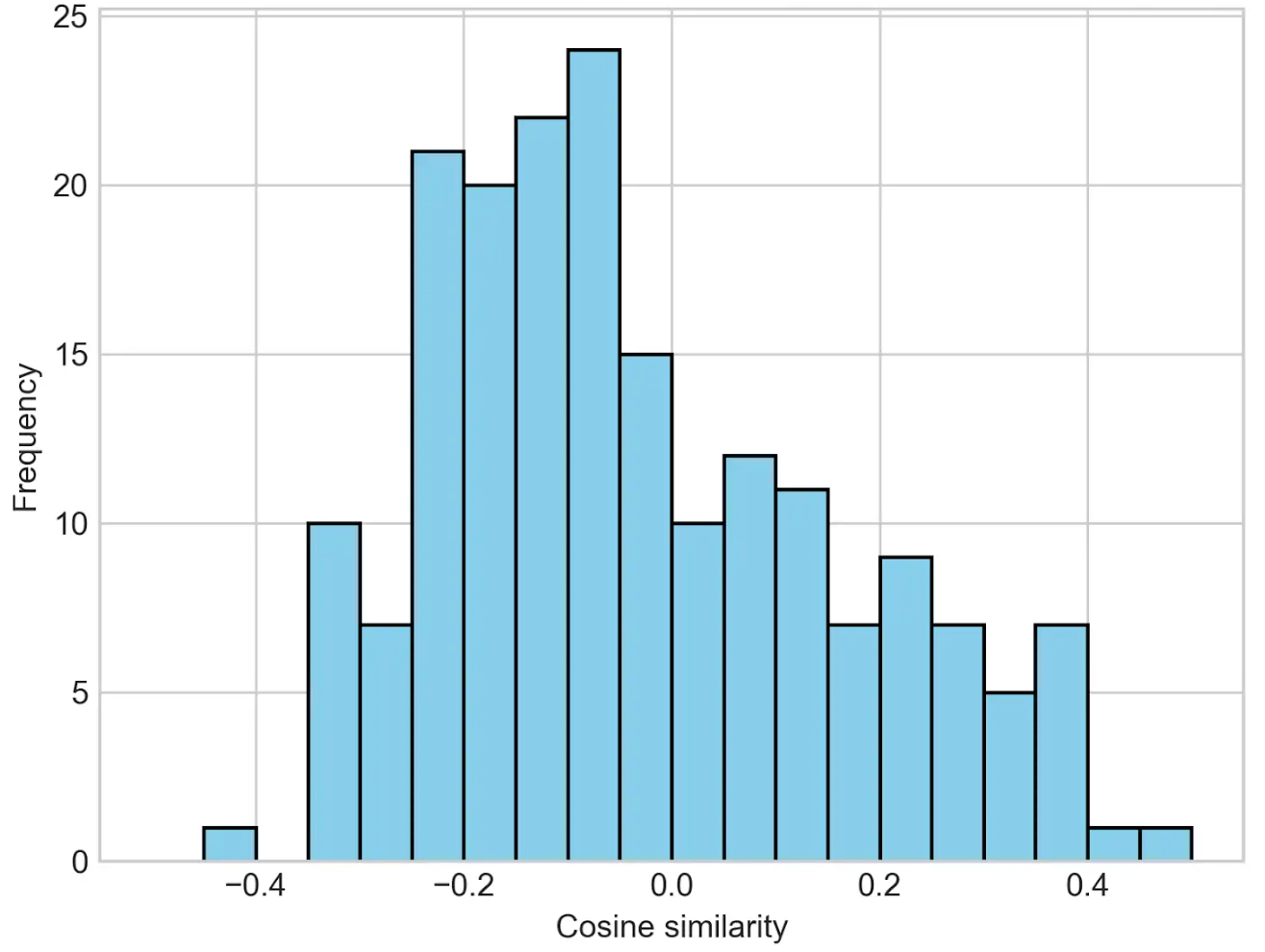}
        \caption{MetaQA}
    \end{subfigure}
    \caption{The distribution of cosine similarities between consecutive query topics in our synthetic queries. From left to right: YAGO, DBpedia, MetaQA. Our synthetic queries cover diverse topic-evolving scenarios. }
    \label{Fig: similarity_plots_datasets}
\end{figure*}

To the best of our knowledge, there is no paired user query log and KG dataset, because the public user query datasets are anonymized by the publisher and do not have user identification. Therefore, it is impossible to extract one specific user’s consecutive queries. Nonetheless, we conduct our experiments over our synthetic but realistic query logs over benchmark KGs. In this section, we further validate that our synthetic queries are of high quality.

One piece of evidence is that our sampled queries follow the logic and structure of Linked SPARQL Queries’ DBpedia anonymous data dump\footnote{\url{https://files.dice-research.org/archive/lsqv2/dumps/dbpedia/}}. In both LSQ and our queries, multiple query relations on the same query head appear in a batch and get answered sequentially. Then query relations start from another entity head and repeat such a pipeline.

We further conduct a study on the semantic of sampled topics. As discussed in Section \ref{experiments}, for each dataset, we sample 10 users and 200 queries for each user. To be specific, for each user, we uniformly randomly sampled 20 entities (i.e., topics) from all the entities in the KG. The user querying topics are evolving, which is exactly the reason why we need adaptive personalized KG summarization. Therefore, we measure the semantic distances of each pair of consecutive topics. For example, sample a random user in the real-world MetaQA dataset, and we can get the query topic history as \textit{['Stevan Riley', 'The Disappearance of Haruhi Suzumiya', 'LOL', 'Chad Michael Murray', 'Orson Scott Card', 'Gérard Lanvin', 'menahem golan', '...All the Marbles', 'Operator 13', 'Heaven Help Us', 'peter pan', 'Mark Saltzman', 'Denise Dillaway', 'Hell Drivers', 'drummer', 'Beware of Pity', 'The Cruel Sea', 'The Great Gabbo', 'Takuya Kimura', 'The Secrets']}. From this log of topics, we can see the topic changes from different genre movies to different directors and even novels. Using the widely-used language model RoBERTa \cite{DBLP:journals/corr/abs-1907-11692}, we can compute the normalized embedding vector for each pair of consecutive topics. Their cosine similarities are [-0.2100357562303543, -0.2824622392654419, -0.18836577236652374, 0.2763717472553253, 0.2524639368057251, 0.03907020390033722, -0.19497732818126678, -0.14980322122573853, -0.041029710322618484, -0.16855204105377197, 0.18958419561386108, 0.3601977825164795, -0.278546005487442, 0.12963563203811646, -0.04322625696659088, 0.10026416927576065, 0.3284454941749573, -0.15075044333934784, -0.3128302991390228]. Such semantic distances range wide and are diverse, validating that our sample user query logs cover diverse topic-evolving scenarios and are of high quality.

We also conduct the visualization for all users in every dataset to show their query topic shifts. The figures of the cosine similarity frequency of all topic shifts for all datasets are shown in Figure \ref{Fig: similarity_plots_datasets}. Here, “frequency” means, among a total of $190 = 10 \times (20-1)$ topic changes, i.e., number of users times topic evolves per user, how many times the cosine similarity between consecutive topics falls into a specific range. In general, we can observe that the various kinds of topics evolve, which demonstrates that our query synthesis and our adaptive summarization setting are challenging. Under such a realistic setting, our proposed methods achieve effectiveness and efficiency outperformance than baseline methods.

\subsection{Reproducibility}
\label{AP: reproducibility} 
\subsubsection{Code} The code for the main experiment is provided at 
\url{https://github.com/iDEA-iSAIL-Lab-UIUC/APEX}.
The data of DBpedia and MetaQA is provided within our submission code, but YAGO3 is not due to its large size. You need to manually download YAGO3 follow our instruction if you want to play with it. Our code only supports ".gz" Compressed Archive Folders, so you need to convert using gzip if the downloaded files are in other format.

\subsubsection{YAGO3} \url{https://www.mpi-inf.mpg.de/departments/databases-and-information-systems/research/yago-naga/yago/downloads/}. Download YAGO Themes -> CORE- > yagoFacts (all facts of YAGO that hold between instances). You need to right click "Download TSV" icon, then click "Save link as..." to download.

\subsubsection{DBpedia} Provided within our code. Original download link: \url{http://downloads.dbpedia.org/3.5.1/en/}. Specifically, download the "mappingbased\_properties\_en.nt" file.

\subsubsection{MetaQA} Provided within our code. Original download link: \url{https://github.com/yuyuz/MetaQA}. The entire MetaQA dataset can be downloaded from the link provided in their README.md.

\subsubsection{Freebase} \url{https://paperswithcode.com/dataset/fb15k-237}. The entire FB15K237 dataset can be downloaded by clicking "Homepage".

\subsubsection{Environments} We run all our experiment on a Windows 10 machine with Intel(R) Core(TM) i7-10750H CPU @ 2.60GHz and 32GB RAM. For other different platforms such as Linux, you may need to reset the path and encoding in code/src/path.py file. The Python version is 3.8.13 in our experiment.

\subsection{Pre-experiments to Pick Baseline Re-summary Interval}
\label{AP: pre-exp}

\begin{figure}[t]
    \centering
    \includegraphics[width=0.47\textwidth]{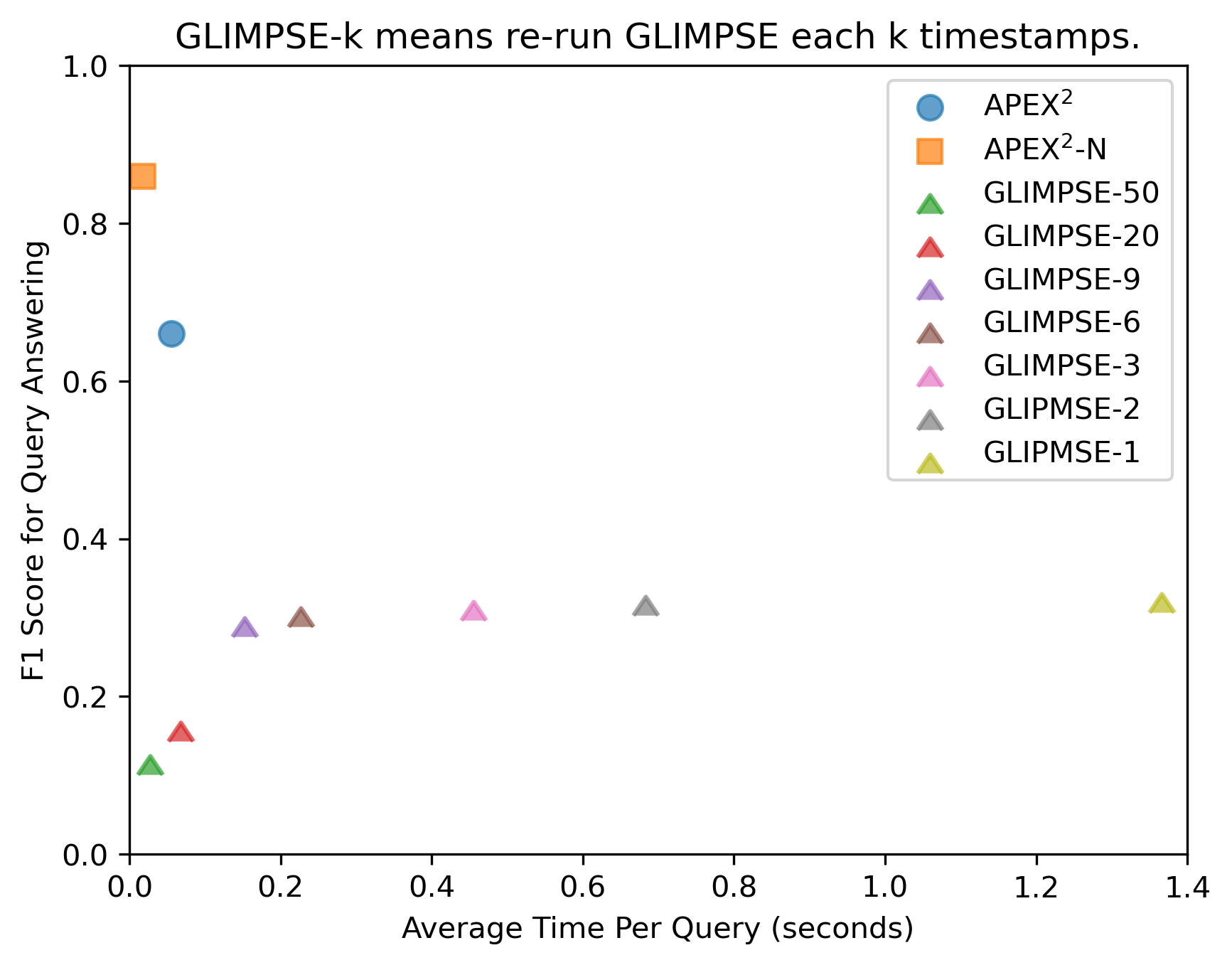}
    \caption{Our pre-experiment to determine that $R=9$ is a good trade-off for accuracy and efficiency.}
    \label{fig: pre-exp}
\end{figure}

For the baseline methods, we vary their $R$ (re-summarization interval) hyperparameter. We observe that when $R=9$, their query answering accuracy does not drop much compared to $R=1$. But if $R$ exceeds $9$, their query answering accuracy drops significantly. An example curve for GLIMPSE on the MetaQA dataset is shown in Figure \ref{fig: pre-exp}. Therefore, we pick the Pareto-optimal $R=9$ for the baselines.

\begin{figure*}[ht]
    \centering
    
    \begin{subfigure}[b]{0.245\textwidth}
        \centering
        \includegraphics[width=\textwidth]{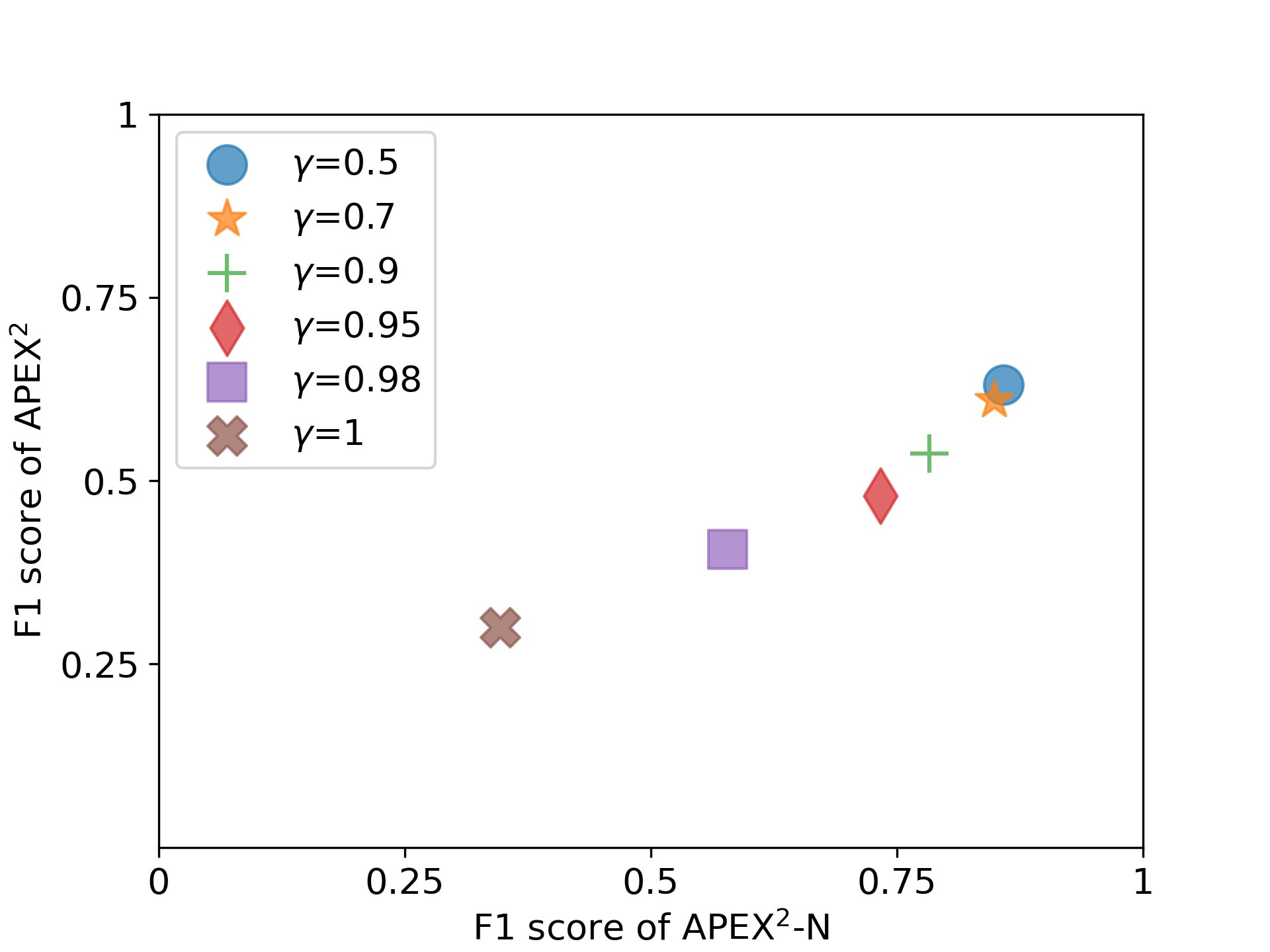}
        \caption{$R_{APEX} = 1$}
    \end{subfigure}
    \hfill
    \begin{subfigure}[b]{0.245\textwidth}
        \centering
        \includegraphics[width=\textwidth]{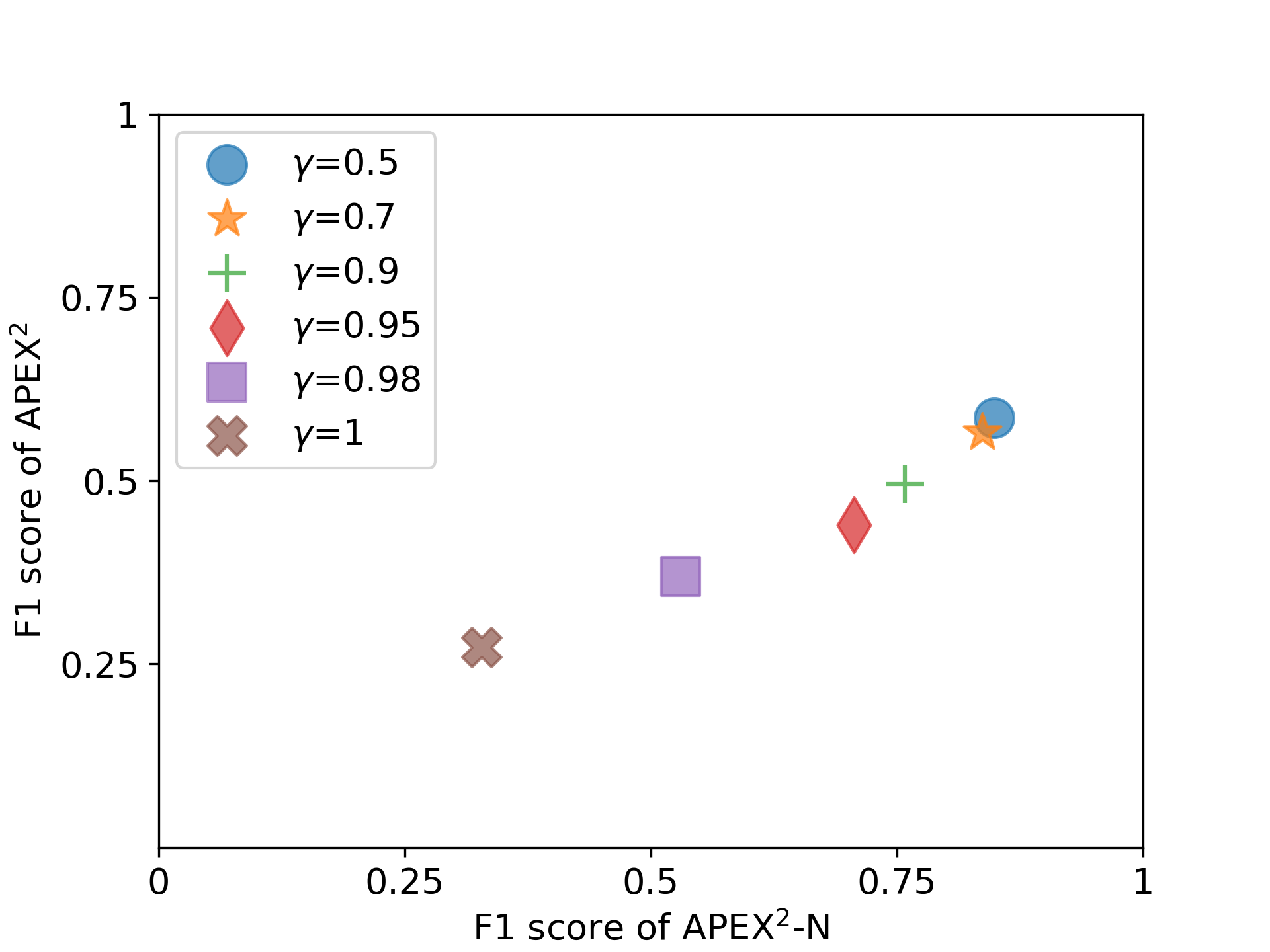}
        \caption{$R_{APEX} = 2$}
    \end{subfigure}
    \hfill
    \begin{subfigure}[b]{0.245\textwidth}
        \centering
        \includegraphics[width=\textwidth]{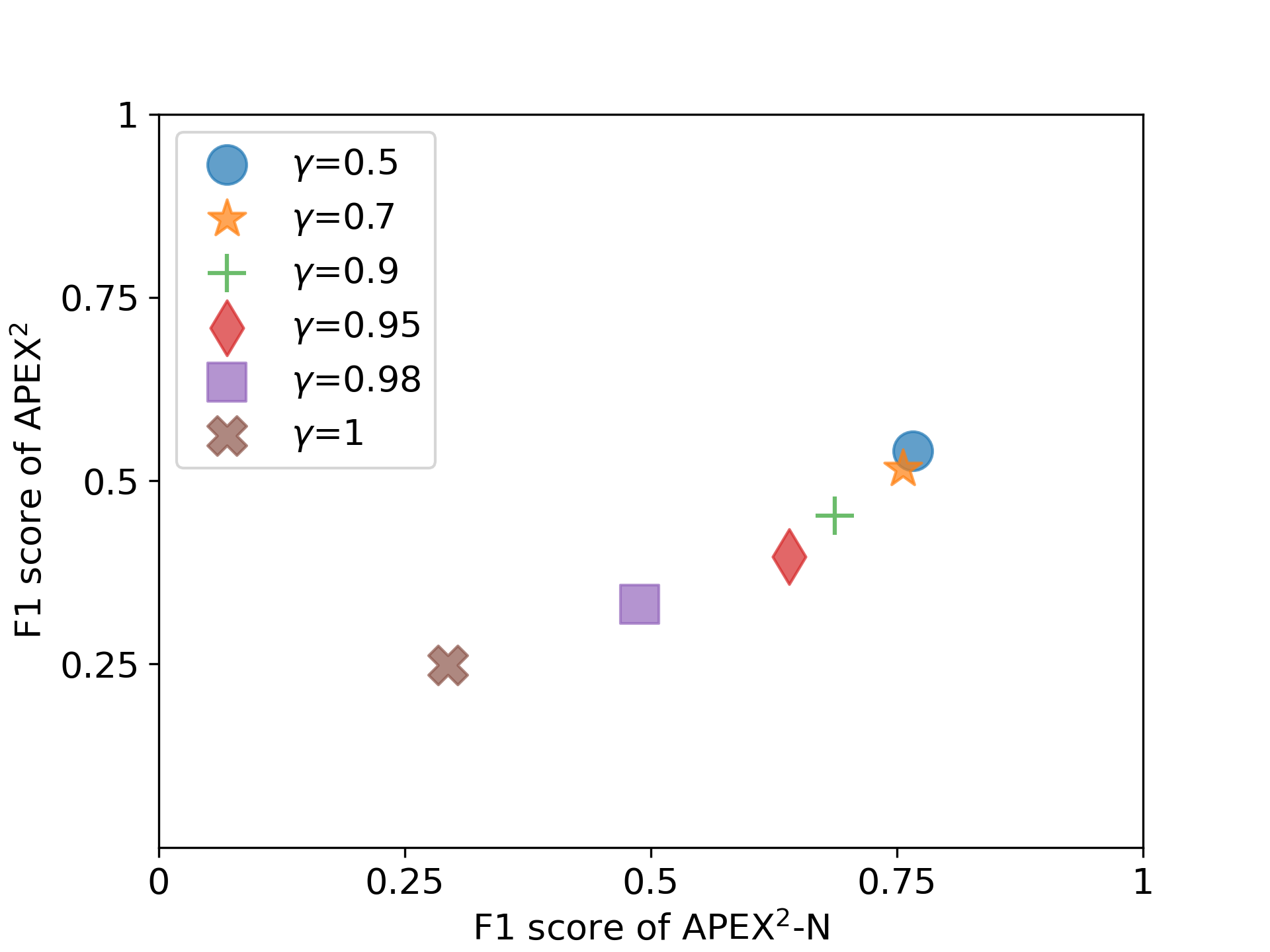}
        \caption{$R_{APEX} = 3$}
    \end{subfigure}
    \hfill
    \begin{subfigure}[b]{0.245\textwidth}
        \centering
        \includegraphics[width=\textwidth]{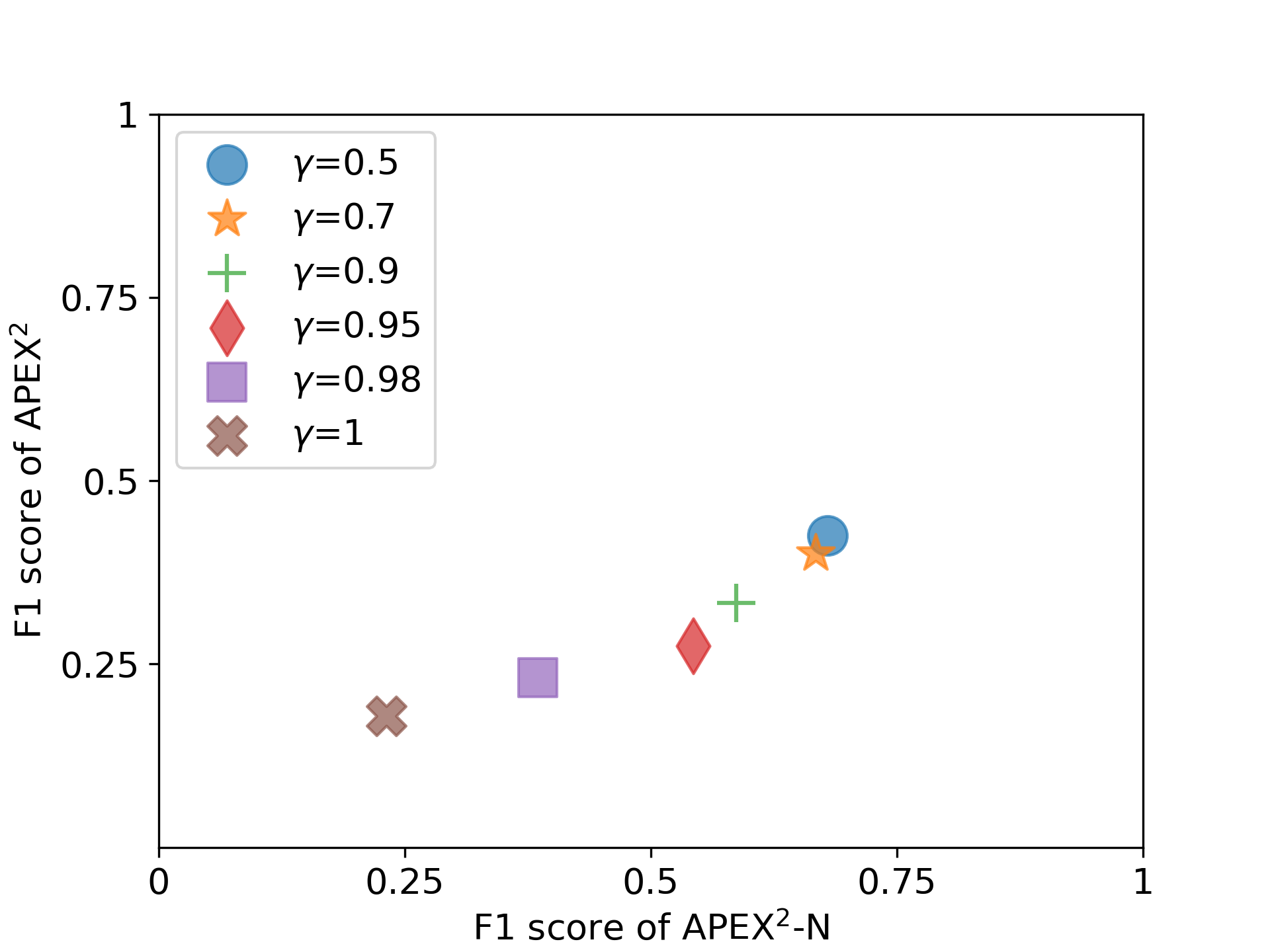}
        \caption{$R_{APEX} = 6$}
    \end{subfigure}
    
    \caption{Querying Effectiveness in Multiple Decay Levels for different $R_{APEX}$.}
    \label{fig: ablation_decay_R}
\end{figure*}

\subsection{Experiments Adjusting $R_{APEX}$}
\label{sec: R_APEX exp}

APEX$^2$ and APEX$^2$-N can re-summarize the KG every $R_{APEX}$ timestamp to increase overall efficiency, denoted by APEX$^2$-$R_{APEX}$ and APEX$^2$-N-$R_{APEX}$. The other experiment settings are the same as the main experiments. The mean performances of the F1 score and per-query time consumption on MetaQA and YAGO are shown in Table \ref{tb: R_APEX effectiveness comparison MetaQA} and \ref{tb: R_APEX effectiveness comparison YAGO}.

\begin{table}[h]
\centering
\caption{The performance of APEX$^2$-$R_{APEX}$ and APEX$^2$-N-$R_{APEX}$ with varying $R_{APEX}$ on MetaQA.}
\vspace{-3mm}
\label{tb: R_APEX effectiveness comparison MetaQA}
\scalebox{0.9}{
\begin{tabular}{|l|c|c|}
\hline
Methods    & F1 Score ($\uparrow$) & Time Consumption Per Query ($\downarrow$) \\ \hline \hline
APEX$^2$-$1$ & 0.631      & 0.055      \\ \hline
APEX$^2$-$2$ & 0.586      & 0.049      \\ \hline
APEX$^2$-$3$ & 0.540      & 0.044      \\ \hline
APEX$^2$-$6$ & 0.426      & 0.036      \\ \hline
APEX$^2$-N-$1$ & 0.858      & 0.018      \\ \hline
APEX$^2$-N-$2$ & 0.849      & 0.011      \\ \hline
APEX$^2$-N-$3$ & 0.766      & 0.007      \\ \hline
APEX$^2$-N-$6$ & 0.680      & 0.004      \\ \hline
\end{tabular}
}
\vspace{-2mm}
\end{table}

\begin{table}[h]
\centering
\caption{The performance of APEX$^2$-$R_{APEX}$ and APEX$^2$-N-$R_{APEX}$ with varying $R_{APEX}$ on YAGO.}
\vspace{-3mm}
\label{tb: R_APEX effectiveness comparison YAGO}
\scalebox{0.9}{
\begin{tabular}{|l|c|c|}
\hline
Methods    & F1 Score ($\uparrow$) & Time Consumption Per Query ($\downarrow$) \\ \hline \hline
APEX$^2$-$1$ & 0.601      & 6.354      \\ \hline
APEX$^2$-$2$ & 0.562      & 4.947      \\ \hline
APEX$^2$-$3$ & 0.524      & 3.433      \\ \hline
APEX$^2$-$6$ & 0.475      & 2.695      \\ \hline
APEX$^2$-N-$1$ & 0.882      & 2.528      \\ \hline
APEX$^2$-N-$2$ & 0.861      & 1.691      \\ \hline
APEX$^2$-N-$3$ & 0.772      & 1.052      \\ \hline
APEX$^2$-N-$6$ & 0.667      & 0.689      \\ \hline
\end{tabular}
}
\vspace{-2mm}
\end{table}

From the results, by adjusting $R_{APEX}$, we can balance between the searching accuracy and the overall time consumption. By choosing a larger $R_{APEX}$ and processing multiple queries together with less granularity, APEX$^2$ and APEX$^2$-N can execute faster, but sacrifice interest tracking accuracy.

\begin{figure*}[h]
    \centering
    
    \begin{subfigure}[b]{1\textwidth}
        \centering
        \includegraphics[width=\textwidth]{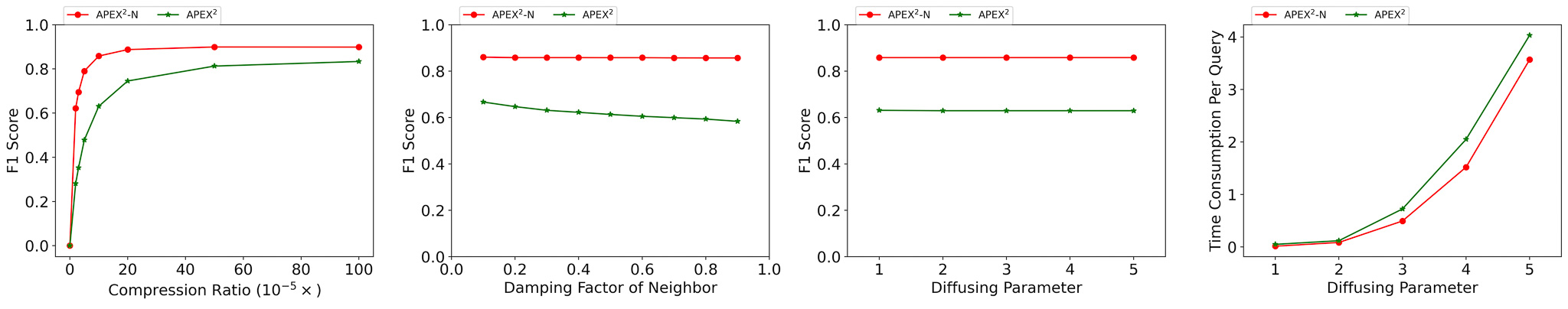}
        \caption{$R_{APEX} = 1$}
    \end{subfigure}

    \begin{subfigure}[b]{1\textwidth}
        \centering
        \includegraphics[width=\textwidth]{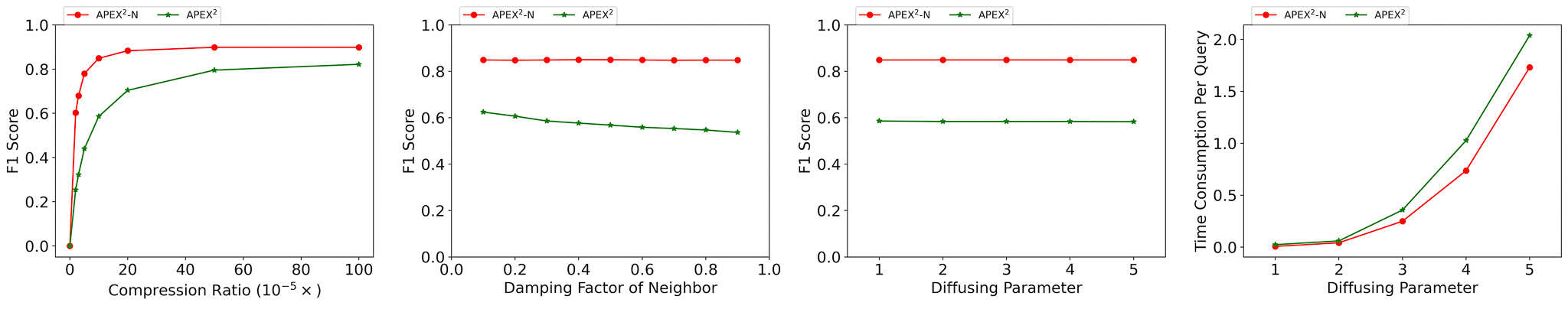}
        \caption{$R_{APEX} = 2$}
    \end{subfigure}

    \begin{subfigure}[b]{1\textwidth}
        \centering
        \includegraphics[width=\textwidth]{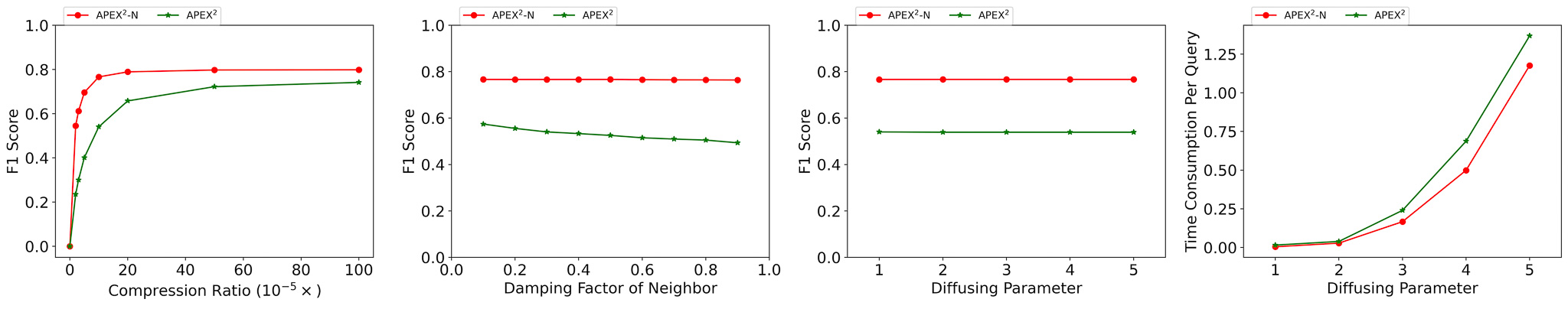}
        \caption{$R_{APEX} = 3$}
    \end{subfigure}

    \begin{subfigure}[b]{1\textwidth}
        \centering
        \includegraphics[width=\textwidth]{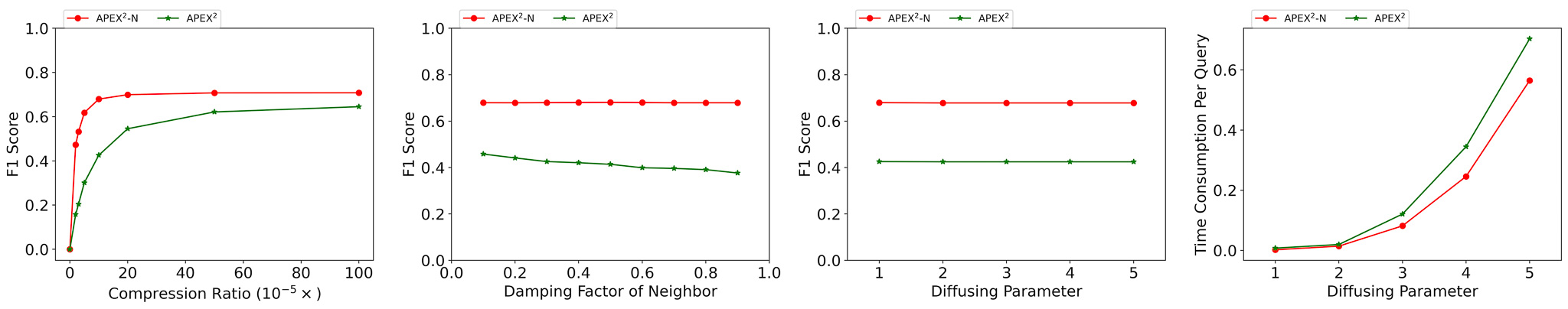}
        \caption{$R_{APEX} = 6$}
    \end{subfigure} 
    
    \caption{Parameter Study adjusting different $R_{APEX}$. From left to right: compression ratio $\kappa$, damping factor of neighbor $\alpha$, diffusing diameter $d$.}
    \label{fig: params_study_R}
\end{figure*}

APEX$^2$-$R_{APEX}$ and APEX$^2$-N-$R_{APEX}$ are similar to APEX$^2$ and APEX$^2$-N, except that the re-summarization interval is different. Because we only masked the summary updating phase in some timestamps, at the timestamps where there is no mask, the output PKGs are identical to those from APEX$^2$ and APEX$^2$-N, respectively. Therefore, the results and analysis from our case study still applies for APEX$^2$-$R_{APEX}$ and APEX$^2$-N-$R_{APEX}$. Additionally, we conduct the decay ablation studies and hyperparameter studies on $R_{APEX} \in \{1, 2, 3, 6\}$ to further study these faster variants. 
The results are shown in Figure \ref{fig: ablation_decay_R} and Figure \ref{fig: params_study_R}. From the ablation results, the decaying mechanism controlled by $\gamma$ is still crucial for APEX$^2$-$R_{APEX}$ and APEX$^2$-N-$R_{APEX}$, as setting $\gamma$ close to 1 will decrease their performances.  From the hyperparameter study results, APEX$^2$-$R_{APEX}$ and APEX$^2$-N-$R_{APEX}$ demonstrate similar trends when adjusting compression ratio $\kappa$, damping factor of neighbor $\alpha$, diffusing diameter $d$. Overall, the effectiveness of APEX$^2$-$R_{APEX}$ and APEX$^2$-N-$R_{APEX}$ are robust to the damping factor of neighbor $\alpha$, diffusing diameter $d$.

\subsection{Handling Very Large Knowledge Graphs}
Theoretically, we give scalability proofs in Theorems 4.2 and 4.4. The complexity is further optimized to be independent of graph size by setting an eliminating threshold. In the main experiments, we verify the scalability of our algorithms by two large KGs containing ~10M triples. To further demonstrate the scalability of our algorithms, we conduct an experiment using a freebase subset of ~30M triples with 0.001\% compression ratio. The mean results are reported in Table \ref{scalability}.

\begin{table}[h]
\centering
\caption{Results Summarizing Freebase subset of 30,000,000 triples.}
\label{scalability}
\scalebox{0.9}{
\begin{tabular}{|l|c|c|}
\hline
Methods    & F1 & time (seconds)\\ \hline \hline
APEX$^2$    & 0.4941      & 8.28  \\ \hline
APEX-N$^2$   & 0.6889      & 2.56  \\ \hline
GLIMPSE  & 0.3273     & 284.86    \\ \hline
PageRank & 0.1448      & 17.2    \\ \hline
\end{tabular}
}
\end{table}

\section{Visual Aid for Heat Diffusion Mechanism}
\label{AP: heat_diffuse_visual_aid}

\begin{figure*}[hbt!]
    \centering
    \includegraphics[width=0.9\textwidth]{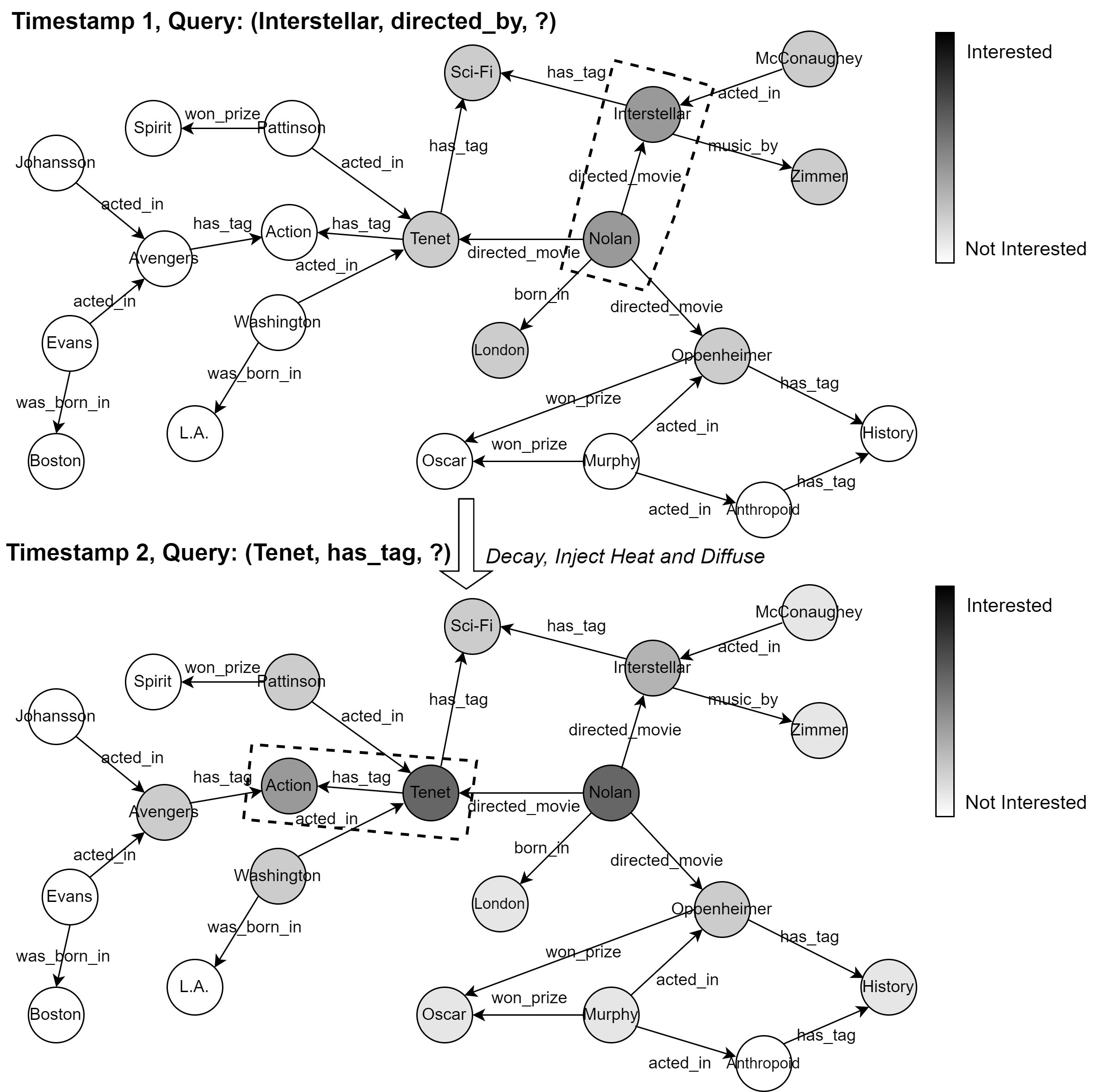}
    \caption{We adopt a heat decay-inject-diffuse framework for heat diffusion. For each timestamp, first, all the heat (i.e., interest) will be decayed; then, an amount of new heat will be injected to the searched entities (marked by the dashed-line boxes); third, a global push is executed, where the heat of each entity will be partially pushed to its neighbors.}
    \label{fig: visual_aid_heat}
\end{figure*}

We use a heat model to simulate and track user's interest from realistic perspectives of human interest phenomenons. \textbf{(i)} When a human searches for one thing over KG or the Internet, it means, in most cases, he or she is interested in that thing. \textbf{(ii)} Human interest transits from one thing to its related things. For example, when a user shows interest in a specific movie, such interest may extend to related aspects such as the actors, director, or genre of the film. \textbf{(iii)} Over time, as humans tend to forget, their interest in a particular thing should naturally decline. From these facts and observations, we adopt a heat decay-inject-diffuse framework for heat diffusion, as shown in Figure \ref{fig: visual_aid_heat}, which shows an illustration of user's interest after each timestamp. 

Initially (Timestamp 0), the user has not interacted with the KG, and the KG is all-white, i.e., we cannot infer if the user is interested in any entity. Then, at timestamp 1, first, there is a \textit{decay} over entities, resulting from all-white to still all-white. Second, the user searches \textit{(Interstellar, directed\_by, ?)}, and gets the answer triple \textit{(Interstellar, directed\_by, Nolan)}. Thus, we \textit{inject} heat into entities \textit{Interstellar} and \textit{Nolan}, as marked by the dashed-line box in Timestamp 1, Figure \ref{fig: visual_aid_heat}. Third, there is a global \textit{push} from all entities to its neighbors. Therefore, the entities connected to \textit{Interstellar} and \textit{Nolan} get partial heat. 

At Timestamp 2, first, there is heat decaying happening for all entities, and therefore entities such as \textit{Interstellar, McConaughey, Zimmer, London} change to lighter gray colors. Second, the user searches \textit{(Tenet, has\_tag, ?)} and gets the answer triple \textit{(Tenet, has\_tag, Action)}. Thus, we \textit{inject} heat into entities \textit{Tenet} and \textit{Action}, as marked by the dashed-line box in Timestamp 2. Third, global push is conducted and entities such as \textit{Pattinson, Averngers, Washington, Oscar, Murphy, History} get pushed partial heat.

For future timestamps, the decay-inject-diffuse mechanism will work similarly as a simulation of user interest tracking.

\section{Proof of Theorems}


\subsection{Proof of Theorem 4.1}
\label{AP: A.5}
\begin{proof}
All the sorting algorithms must output a permutation of $\mathcal{S} \cup \mathcal{C}$. With the fact that there is only one correct permutation that satisfies the sorted condition (the "deterministic" one), reconsider the sorting process to be eliminating the possible permutations by new comparisons. Note that using the prior knowledge in the sorted $\mathcal{S}$ does not count toward new comparisons.

With the input $\mathcal{S}$ and $\mathcal{C}$, let $\mathcal{V}$ be the set of these inputs that are consistent with the answers to all comparisons made so far. Initially, $\mathcal{V}$ contains all $v$ that is a permutation of $ \mathcal{S} \cup \mathcal{C} $ and does not violate any order in $\mathcal{S}$. Each new comparison (\textit{is $a > b?$}) will split $\mathcal{V}$ into two groups: the ones answering "YES" and the ones answering "NO". With the fact that only one group will be the next $\mathcal{V}$, setting the correct answer to the larger group will guarantee $|\mathcal{V}_{next}| \geq \frac{|\mathcal{V}_{current}|}{2}$. 

The algorithm must reduce $|\mathcal{V}|$ to 1 in order to get the output, and the total number of the initial $\mathcal{V}$ can be attained by counting the number of ways to insert $\mathcal{C}$ ($k$ distinct elements) into $\mathcal{S}$ ($n$ already-sorted elements). Use the formula from \textit{Stars and Bars}~\cite{feller1991introduction} ($\mathcal{S}$ contains stars, and $\mathcal{C}$ contains bars, permitting neighboring bars), the total number of ways is
\begin{equation}
\begin{split}
    {{k+n-1}\choose{n-1}} \cdot k! = \frac{(k+n-1)!}{(n-1)!\cdot k!} \cdot k! = \frac{(n+k-1)!}{(n-1)!}
\end{split}
\end{equation}

Thus, when $k << n$, $\log_2 \frac{(n+k-1)!}{(n-1)!} = \Omega(klog_2n)$ is the worst case complexity. The Incremental Binary Insertion Sort has complexity $\log_2n$ for each element in $\mathcal{C}$ to do a binary insertion sort into $\mathcal{S}$, therefore its complexity is $\Omega(klog_2n)$.  \hfill 
\end{proof}

\subsection{Proof of Theorem 4.3}
\label{PF: Time Complexity of APEX}
\begin{proof}
In the initializing phase, the computation of sparse matrix multiplication, $\alpha\mathbf{A}$, summation $\sum_{l=0}^{d}(\alpha\mathbf{A})^l$ and sparse matrix-vector multiplication $\mathbf{q}_{\rm total}^{(0)}$, $\mathbf{e}^{(0)}$, $\mathbf{r}^{(0)}$ needs to be done only one time. Choosing top-$K$ heat triples requires a one-time sort.

In the updating phase, the decaying requires $O(nnz(\mathbf{H}^{(0)})$. The update of $\mathbf{q}_{\rm total}^{(t)}$, $\mathbf{r}^{(t)}$ respectively takes at most $O(nnz(\mathbf{q}_{\rm total}^{(t)})) < O(2|\mathcal{Q}|)$, $O(nnz(\mathbf{r}_{\rm total}^{(t)})) < O(|\mathcal{Q}|)$. The update of $\mathbf{e}^{(t-1)}$ requires $O(nnz(\mathbf{e}_{\rm total}^{(t)}) + 2 |\mathcal{E}|) < O(2|\mathcal{E}||\mathcal{Q}|)$ at most and $O(nnz(\mathbf{e}_{\rm total}^{(t-1)}) + \frac{2nnz(\sum_{l=0}^{d}(\alpha\mathbf{A})^l)}{|\mathcal{E}|}) < O(2\cdot\frac{nnz(\sum_{l=0}^{d}(\alpha\mathbf{A})^l)}{|\mathcal{E}|}\cdot|\mathcal{Q}|)$ on average.

Respectively, for each timestamp $\mathbf{e}[i], \mathbf{r}[j]$ and $\mathbf{e}[k]$ has on average $2$, $1$, and $\frac{2nnz(\sum_{l=0}^{d}(\alpha\mathbf{A})^l))}{|\mathcal{E}|}$ entries updated, therefore $\mathbf{H}$ has on average $O(3\cdot 2 \cdot 2\cdot\frac{nnz(\sum_{l=0}^{d}(\alpha\mathbf{A})^l)}{|\mathcal{E}|}\cdot|\mathcal{Q}|^2) = O(12\cdot\frac{nnz(\sum_{l=0}^{d}(\alpha\mathbf{A})^l)}{|\mathcal{E}|}\cdot|\mathcal{Q}|^2)$ entries changed. The incremental sorting algorithm on $\mathbf{H}$ will take $O(12\cdot\frac{nnz(\sum_{l=0}^{d}(\alpha\mathbf{A})^l)}{|\mathcal{E}|}\cdot|\mathcal{Q}|^2\cdot \log_2nnz(\mathbf{H})) < O(12\cdot\frac{nnz(\sum_{l=0}^{d}(\alpha\mathbf{A})^l)}{|\mathcal{E}|}\cdot|\mathcal{Q}|^2\cdot \log_2(12\cdot\frac{nnz(\sum_{l=0}^{d}(\alpha\mathbf{A})^l)}{|\mathcal{E}|}\cdot|\mathcal{Q}|^3)) = O(36\cdot\frac{nnz(\sum_{l=0}^{d}(\alpha\mathbf{A})^l)}{|\mathcal{E}|}\cdot|\mathcal{Q}|^2\cdot \log_2(12\cdot\frac{nnz(\sum_{l=0}^{d}(\alpha\mathbf{A})^l)}{|\mathcal{E}|}\cdot|\mathcal{Q}|))$.
\end{proof}

\subsection{Proof of Theorem 4.5}
\label{PF: Time Complexity of APEX-N}
\begin{proof}
The calculations in the initializing phase needs to be done only one time. With heat diffuse to be "equally push to neighbors", each HeatDiffuse operation requires $O(\frac{2nnz(\sum_{l=0}^{d}(\alpha\mathbf{A})^l)}{|\mathcal{E}|})$ calculations. 

In the updating phase, the decaying requires $O(nnz(\mathbf{H}^{(0)})$. $\mathbf{H}$ has at most $\frac{2nnz(\sum_{l=0}^{d}(\alpha\mathbf{A})^l)}{|\mathcal{E}|}$ entries changed and an incremental sort requires $O(\frac{2nnz(\sum_{l=0}^{d}(\alpha\mathbf{A})^l)}{|\mathcal{E}|} \log_2 nnz(\mathbf{H})) < O(\frac{2nnz(\sum_{l=0}^{d}(\alpha\mathbf{A})^l)}{|\mathcal{E}|} \cdot \log_2 (\frac{2nnz(\sum_{l=0}^{d}(\alpha\mathbf{A})^l)}{|\mathcal{E}|}|\mathcal{Q}|))$. The construction of $\mathcal{T}_p$ and $\mathcal{R}_p$ requires approximately $O(K^2)$ which is a constant in our problem setting. Therefore the most cost-demanding operation takes $O(c \cdot \log_2(c|\mathcal{Q}|))$ where $c = \frac{2nnz(\sum_{l=0}^{d}(\alpha\mathbf{A})^l)}{|\mathcal{E}|}$.
\end{proof}

\begin{figure*}[h]
    \centering
    \includegraphics[width=0.9\textwidth]{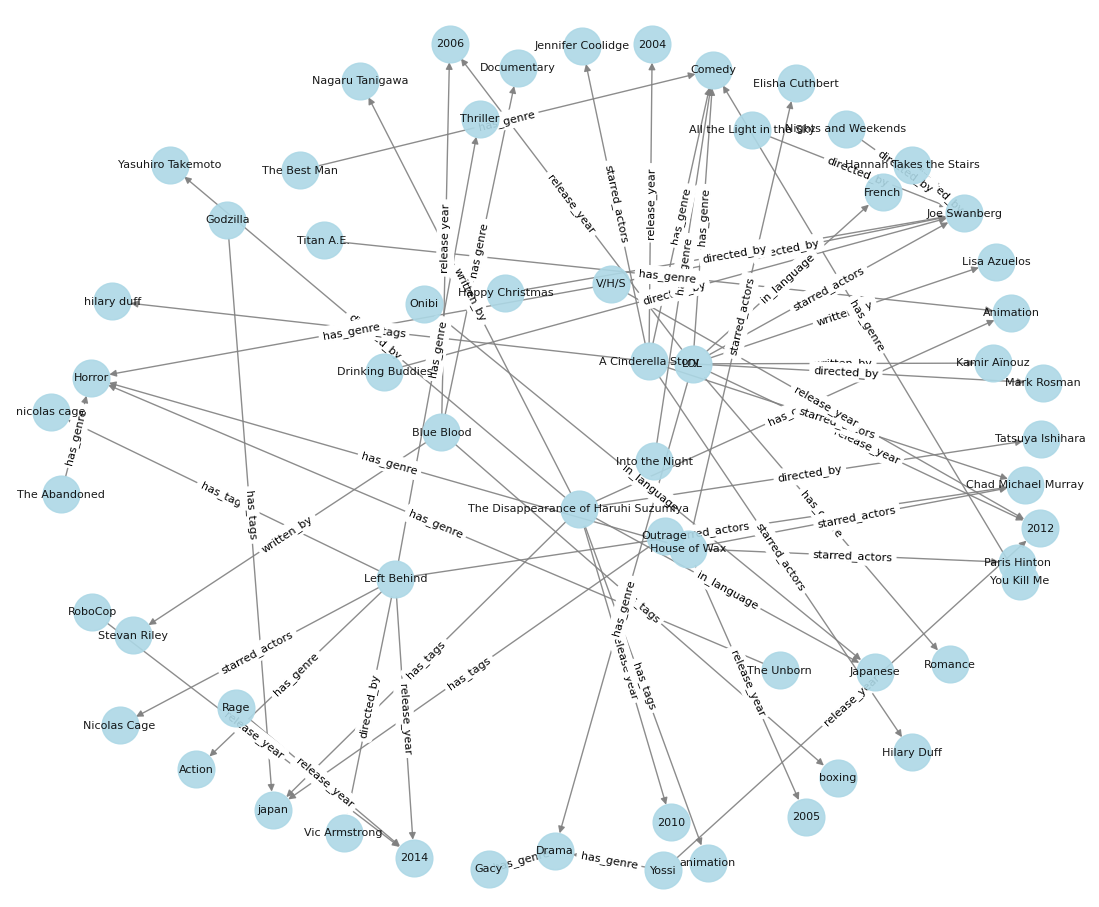}
    \caption{The specific part of MetaQA Knowledge Graph within the interest of our case study. The summarization operates on the whole KG, not only this small portion of KG.}
    \label{fig: case_study}
\end{figure*}

\begin{figure*}[t]
    \centering
    \begin{subfigure}[]{0.47\textwidth}
        \centering
        \includegraphics[width=\textwidth]{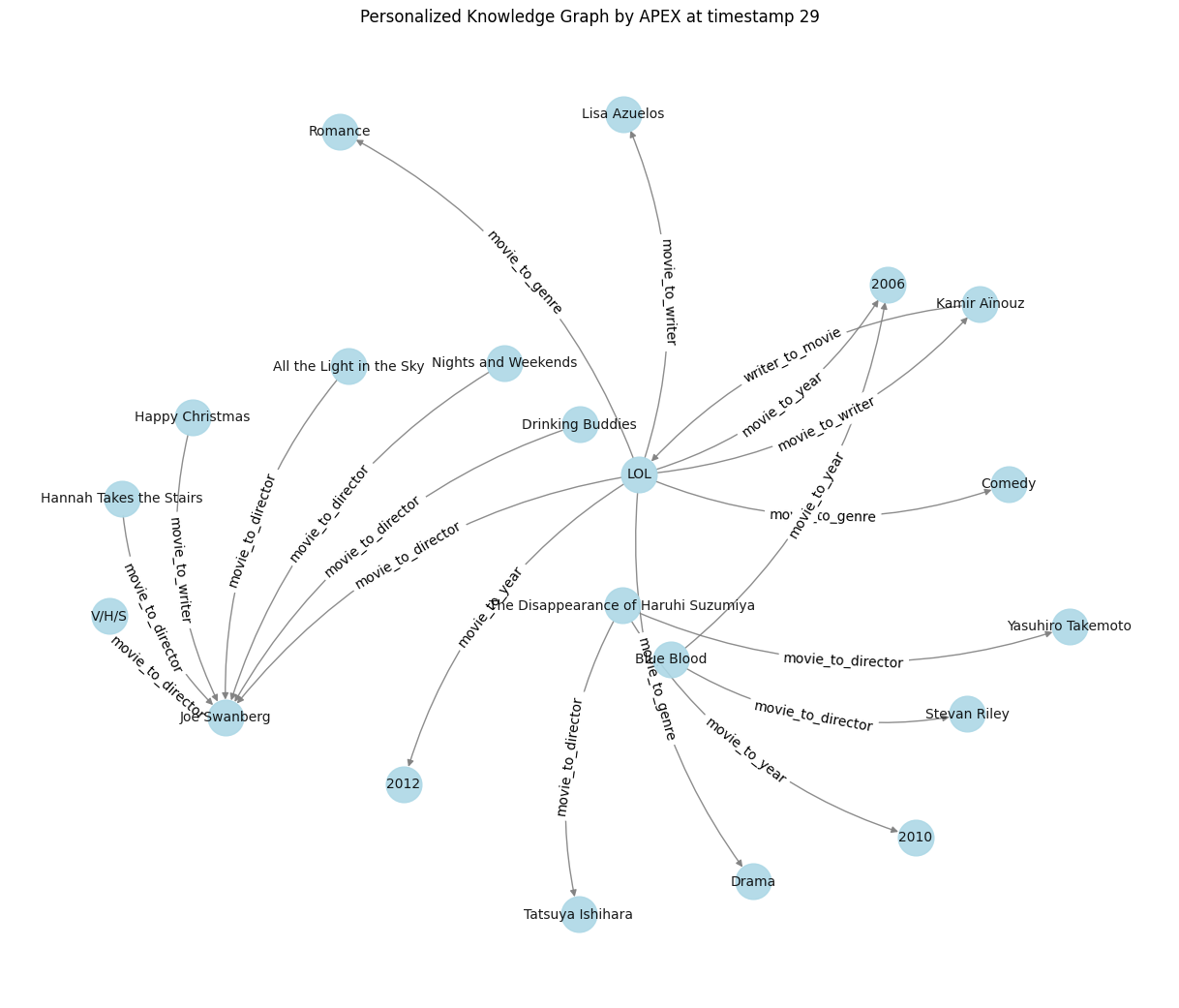}
        \caption{PKG Before the New Topic}
    \end{subfigure}
    \begin{subfigure}[]{0.47\textwidth}
        \centering
        \includegraphics[width=\textwidth]{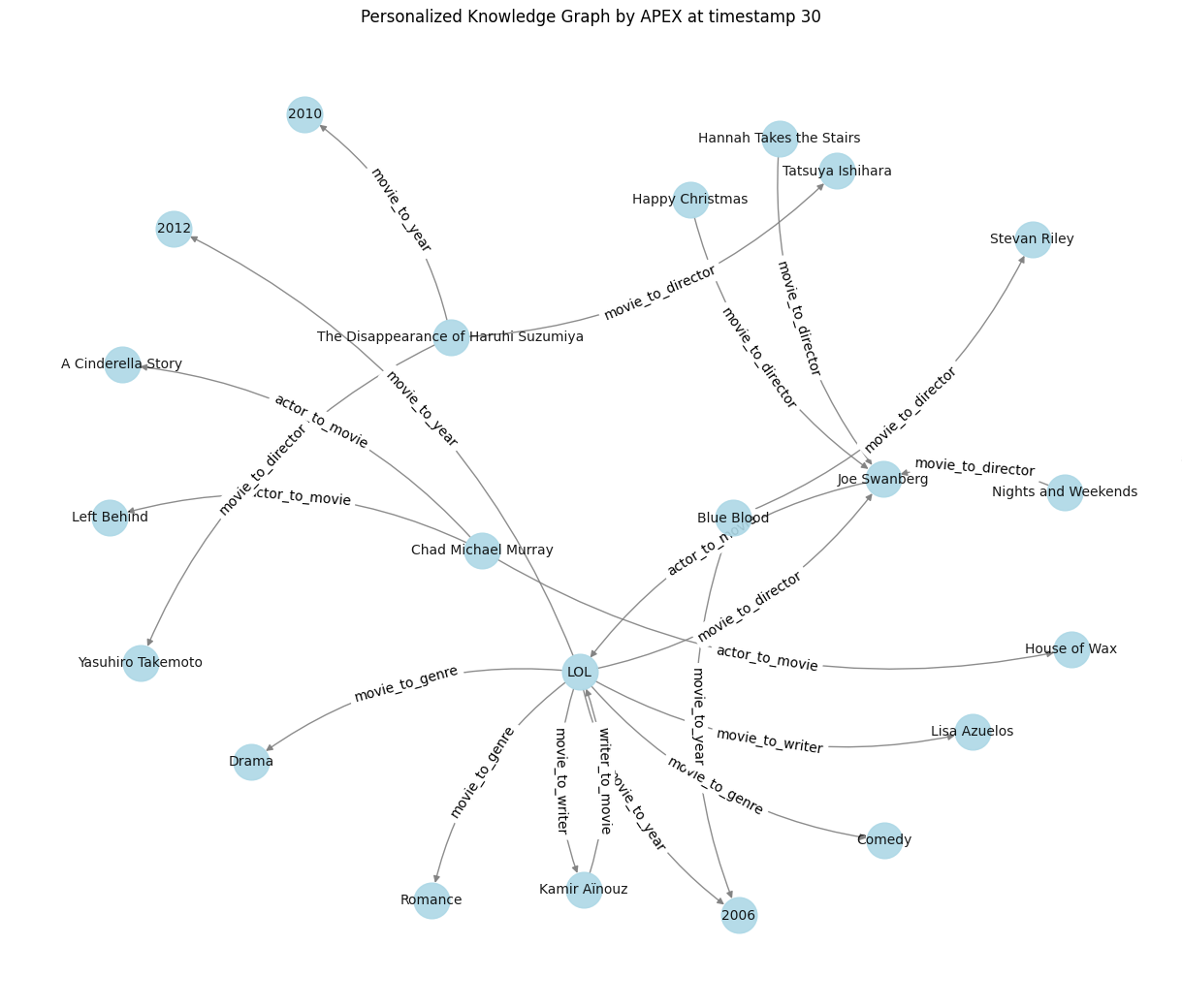}
        \caption{PKG after the First Query on the New Topic}
    \end{subfigure}
    \vskip\baselineskip
    \begin{subfigure}[]{0.47\textwidth}
        \centering
        \includegraphics[width=\textwidth]{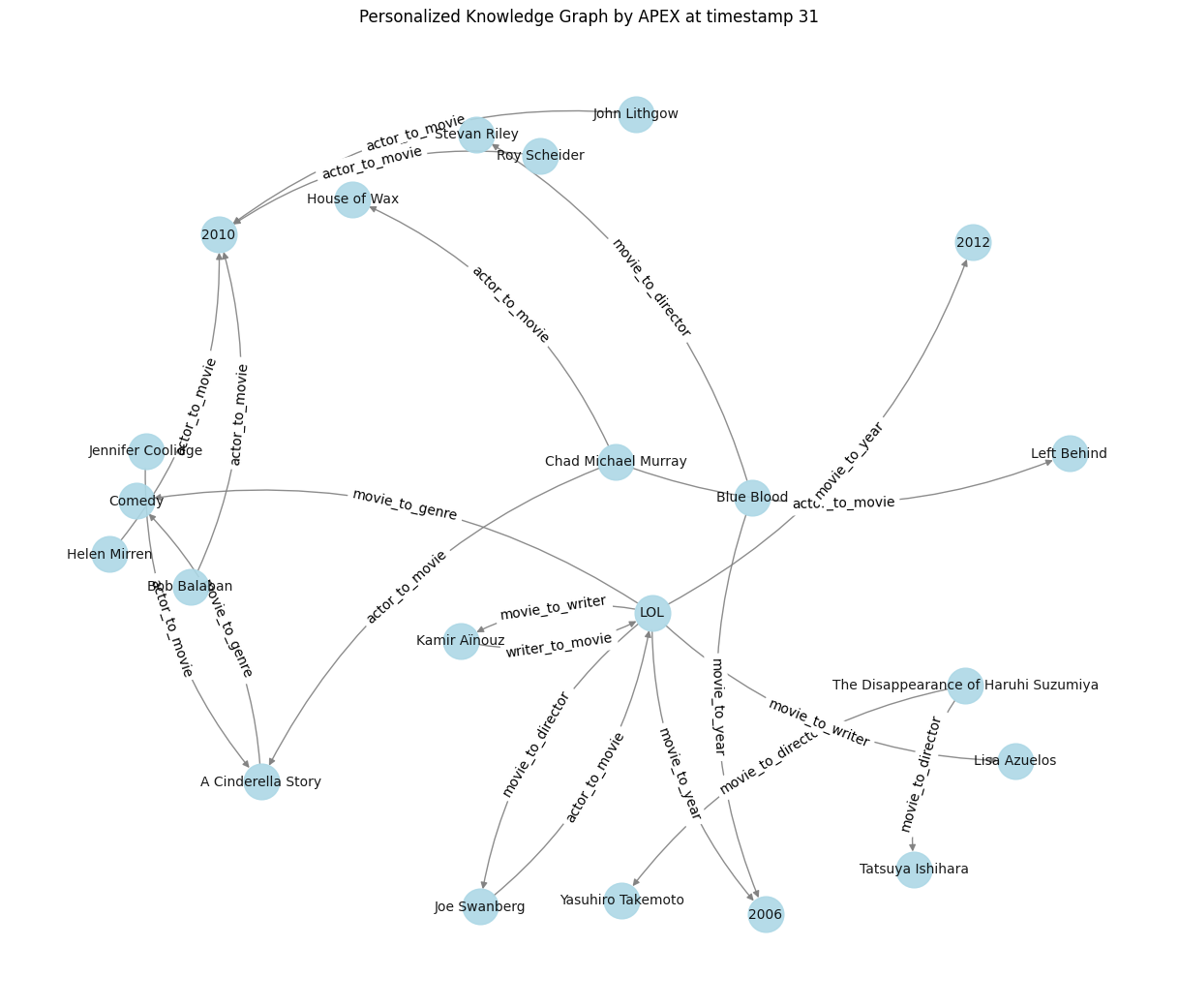}
        \caption{PKG after the Second Query on the New Topic}
    \end{subfigure}
    \begin{subfigure}[]{0.47\textwidth}
        \centering
        \includegraphics[width=\textwidth]{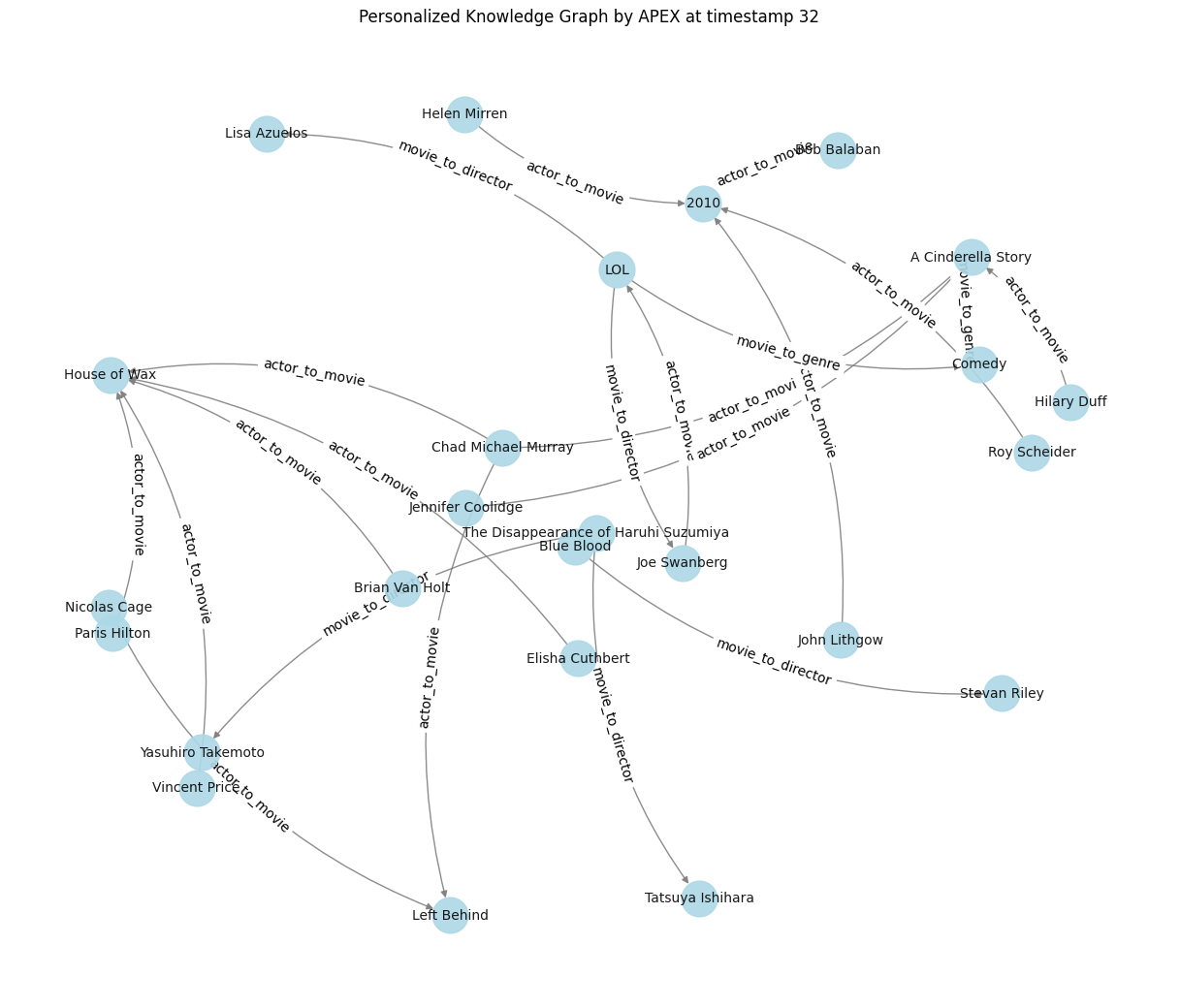}
        \caption{PKG after the Third Query on the New Topic}
    \end{subfigure}
    \caption{Case Study on APEX$^2$}
    \label{fig: case_study_APEX}
\end{figure*}

\begin{figure*}[t]
    \centering
    \begin{subfigure}[]{0.47\textwidth}
        \centering
        \includegraphics[width=\textwidth]{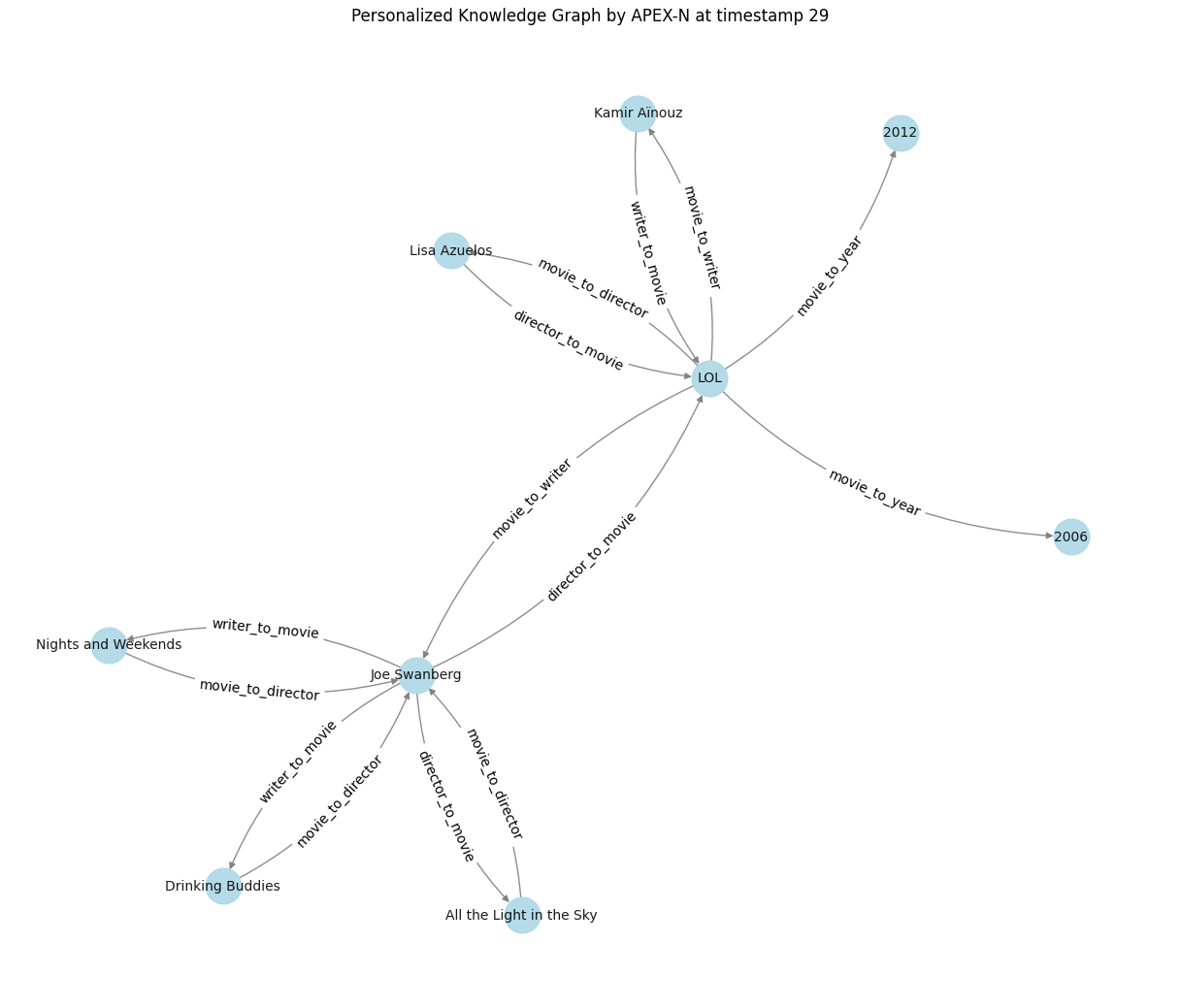}
        \caption{PKG Before the New Topic}
    \end{subfigure}
    \begin{subfigure}[]{0.47\textwidth}
        \centering
        \includegraphics[width=\textwidth]{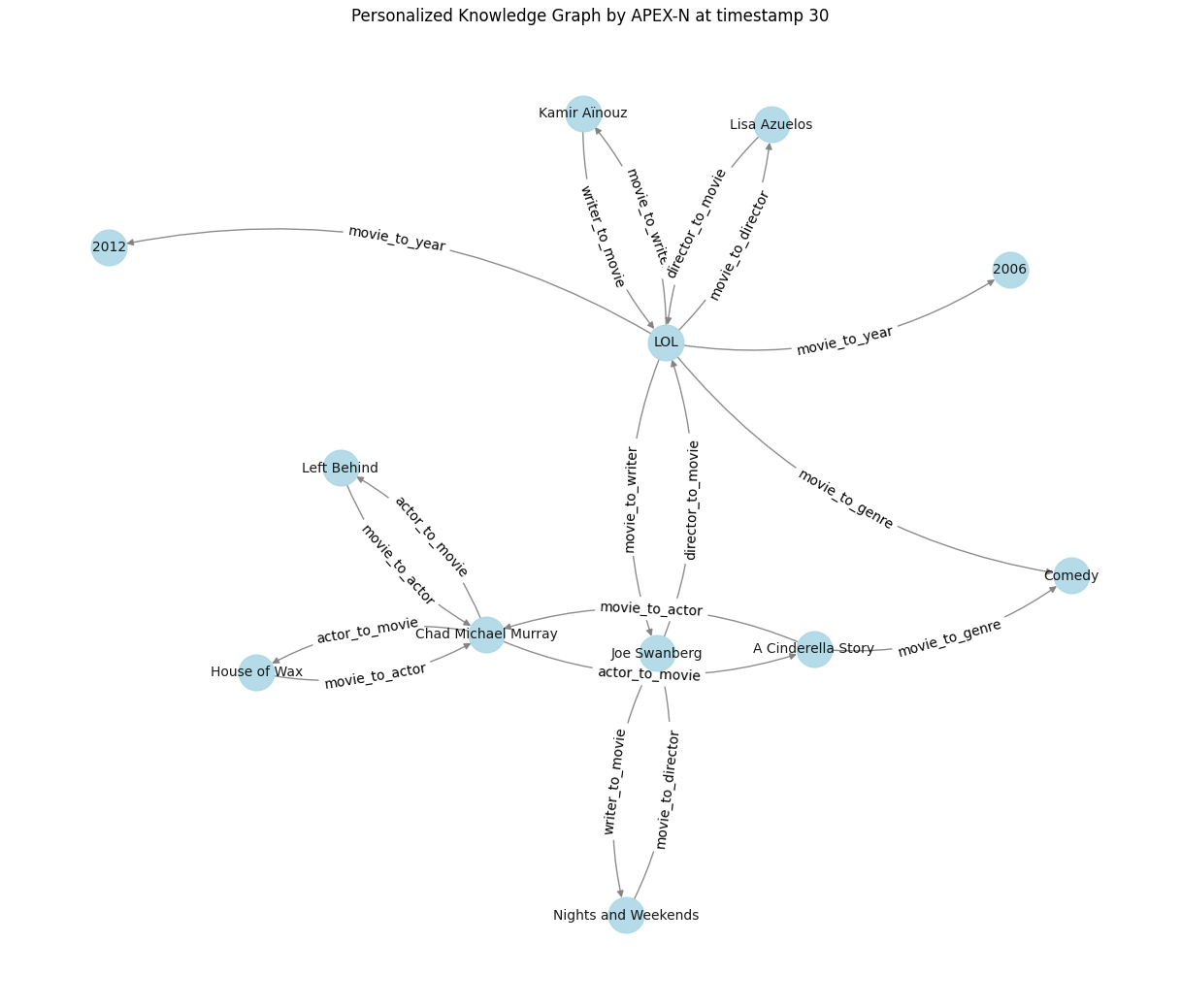}
        \caption{PKG after the First Query on the New Topic}
    \end{subfigure}
    \vskip\baselineskip
    \begin{subfigure}[]{0.47\textwidth}
        \centering
        \includegraphics[width=\textwidth]{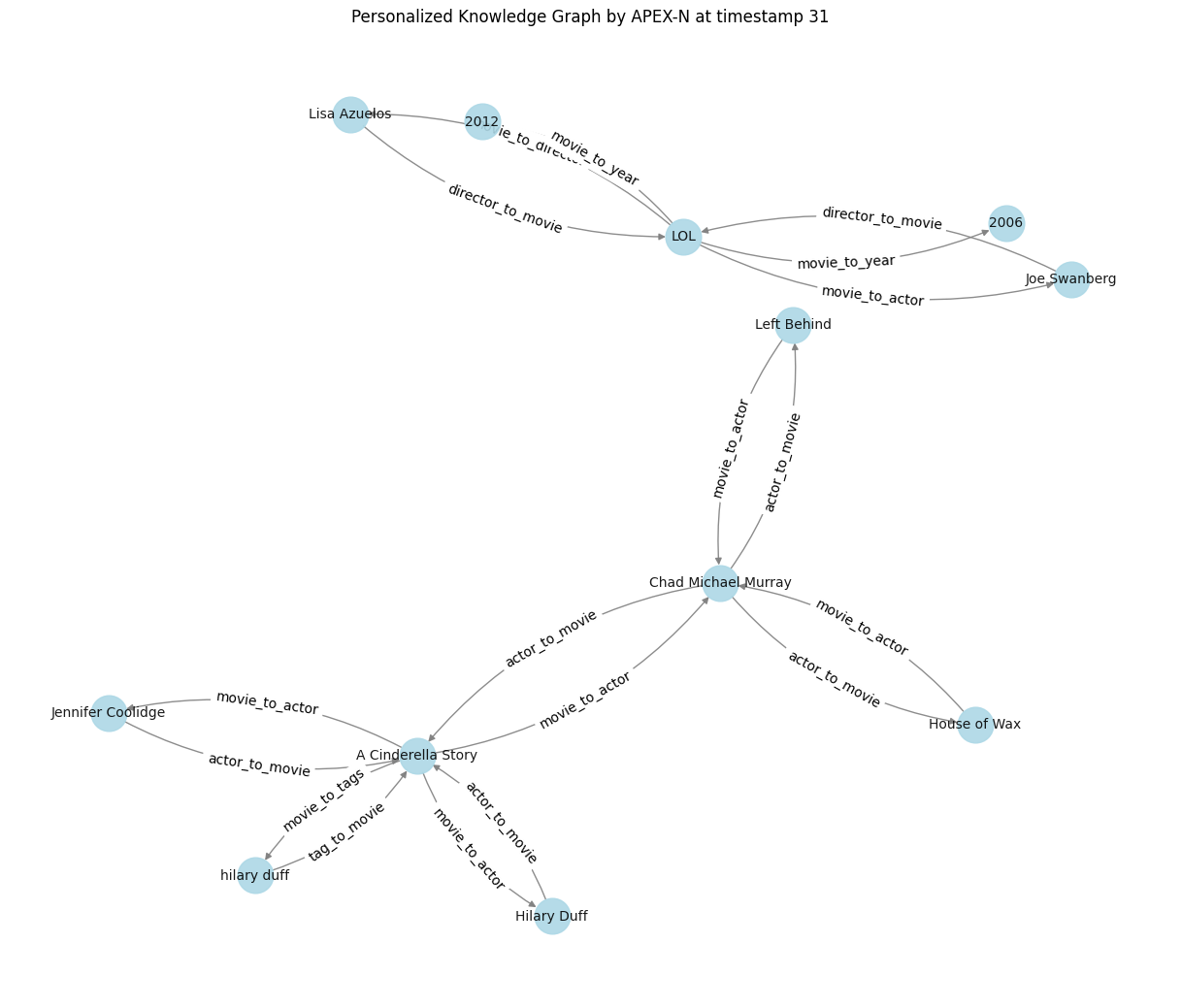}
        \caption{PKG after the Second Query on the New Topic}
    \end{subfigure}
    \begin{subfigure}[]{0.47\textwidth}
        \centering
        \includegraphics[width=\textwidth]{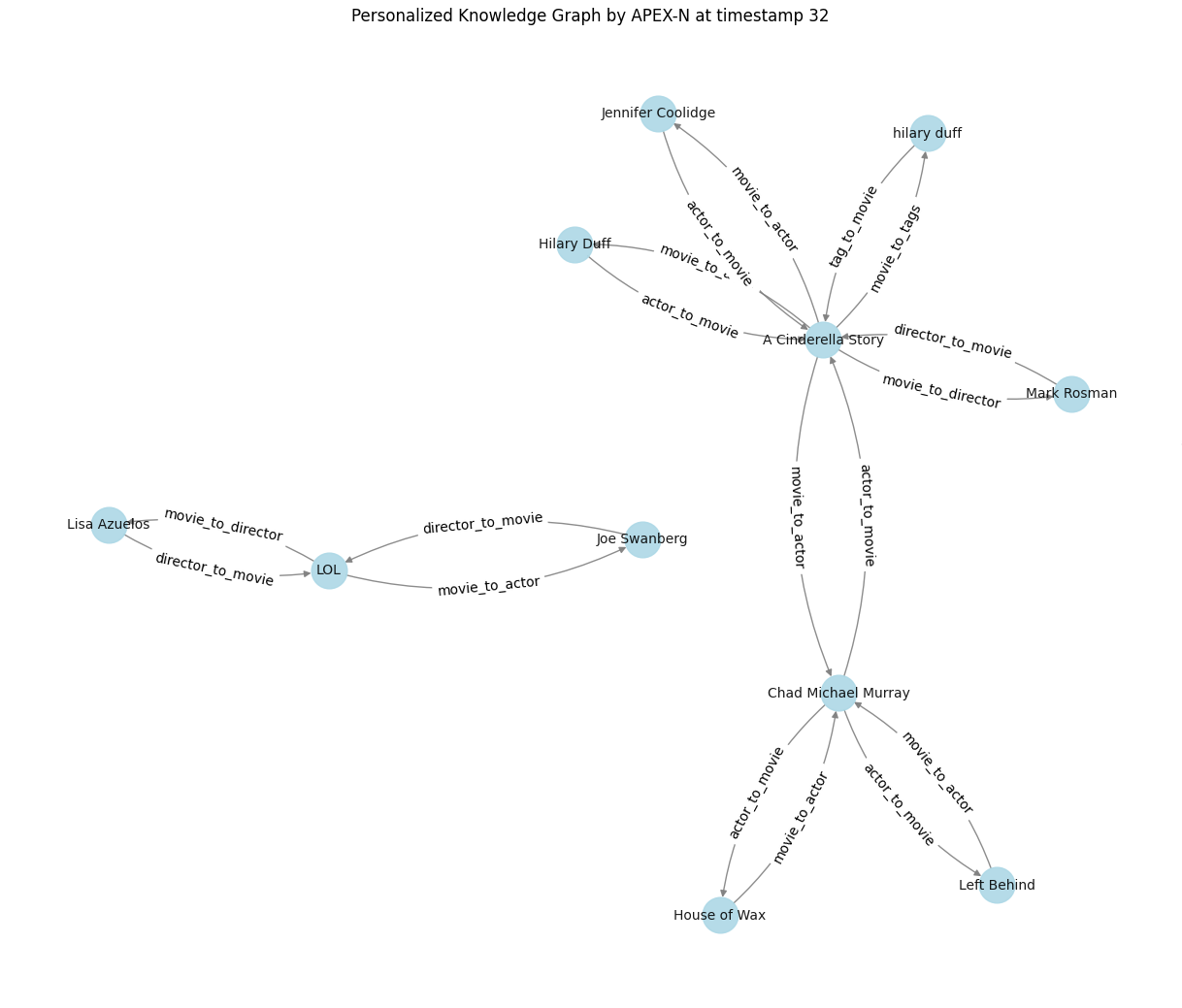}
        \caption{PKG after the Third Query on the New Topic}
    \end{subfigure}
    \caption{Case Study on APEX$^2$-N}
    \label{fig: case_study_APEX_N}
\end{figure*}

\begin{figure*}[t]
    \centering
    \begin{subfigure}[]{0.47\textwidth}
        \centering
        \includegraphics[width=\textwidth]{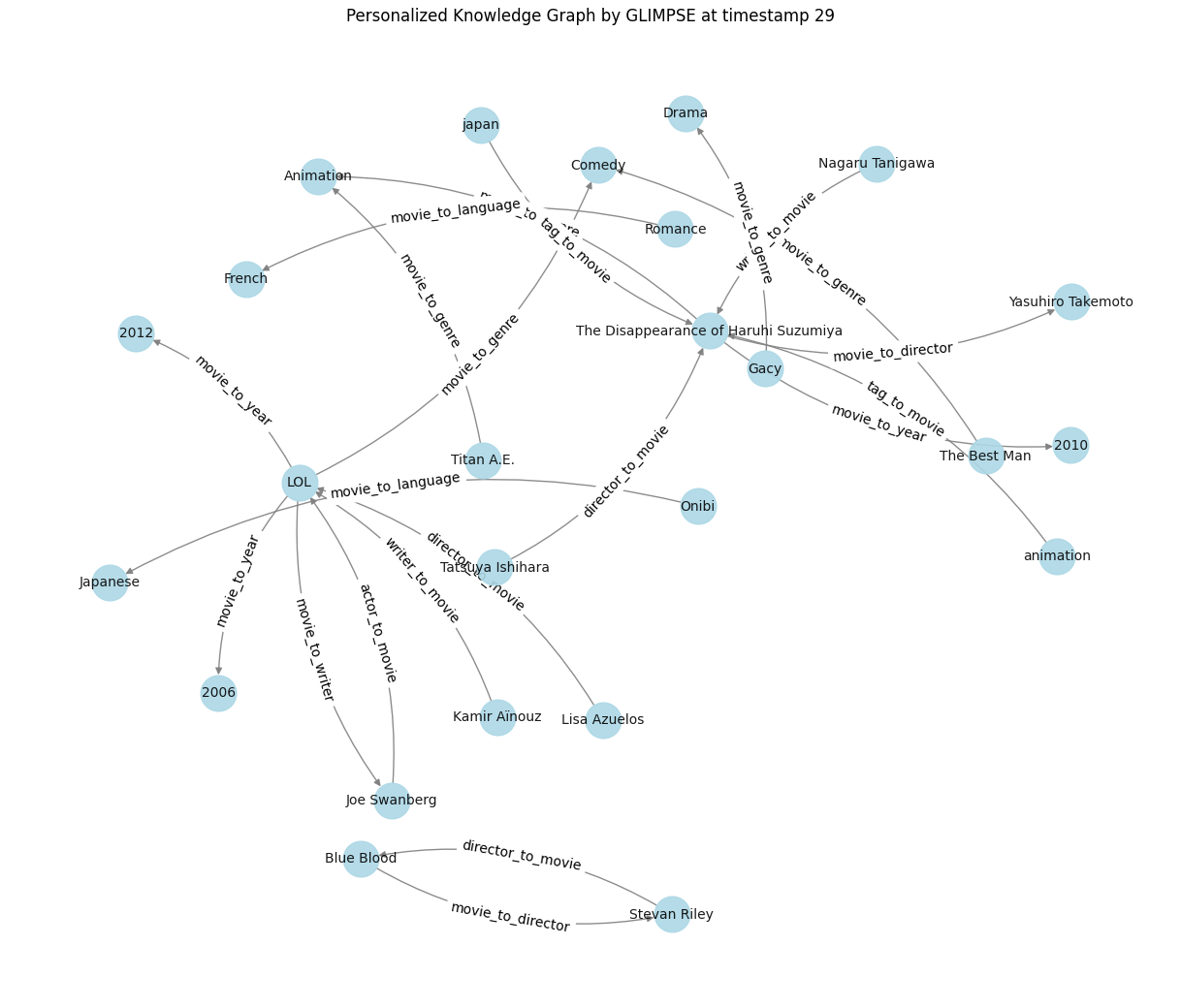}
        \caption{PKG Before the New Topic}
    \end{subfigure}
    \begin{subfigure}[]{0.47\textwidth}
        \centering
        \includegraphics[width=\textwidth]{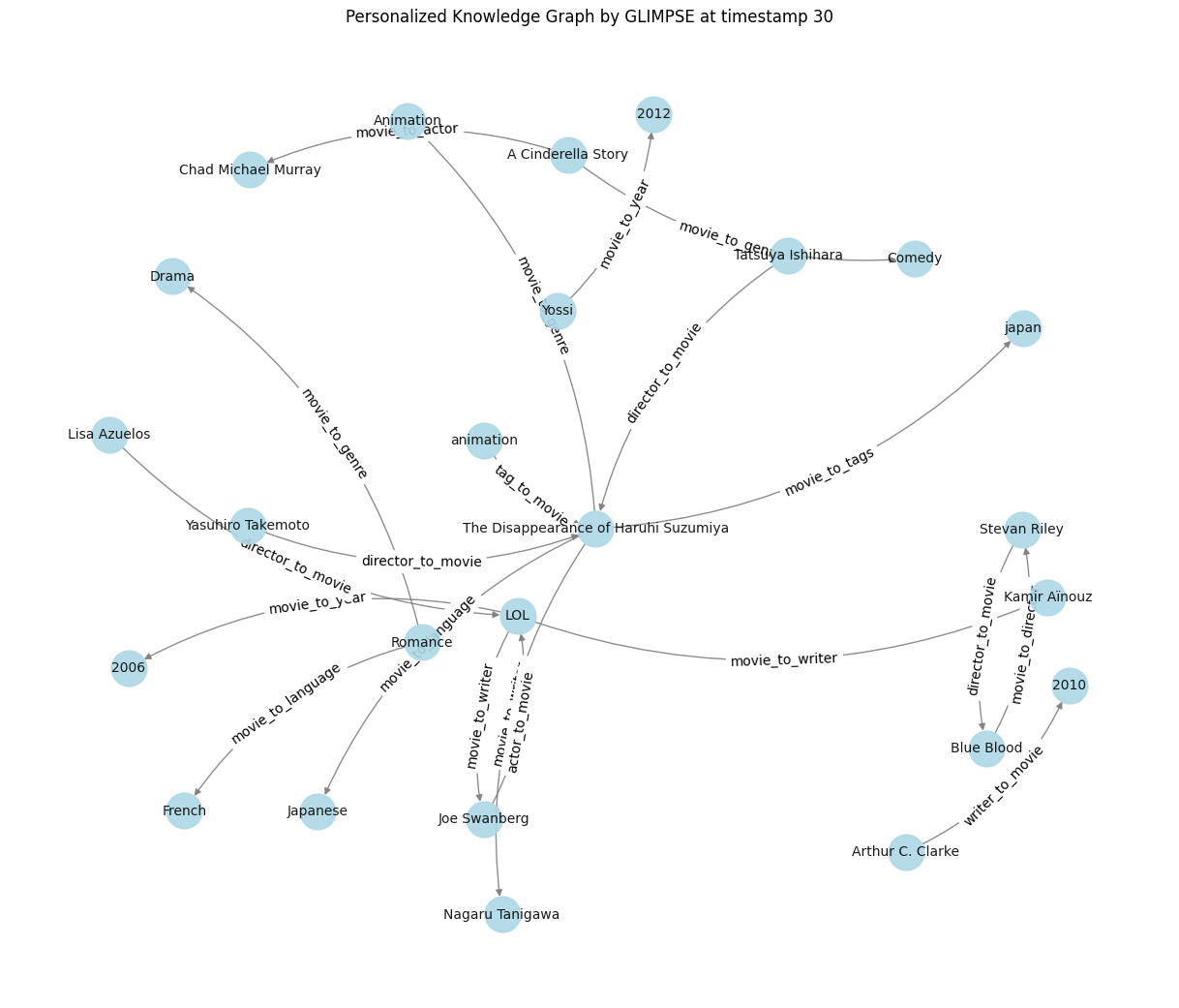}
        \caption{PKG after the First Query on the New Topic}
    \end{subfigure}
    \vskip\baselineskip
    \begin{subfigure}[]{0.47\textwidth}
        \centering
        \includegraphics[width=\textwidth]{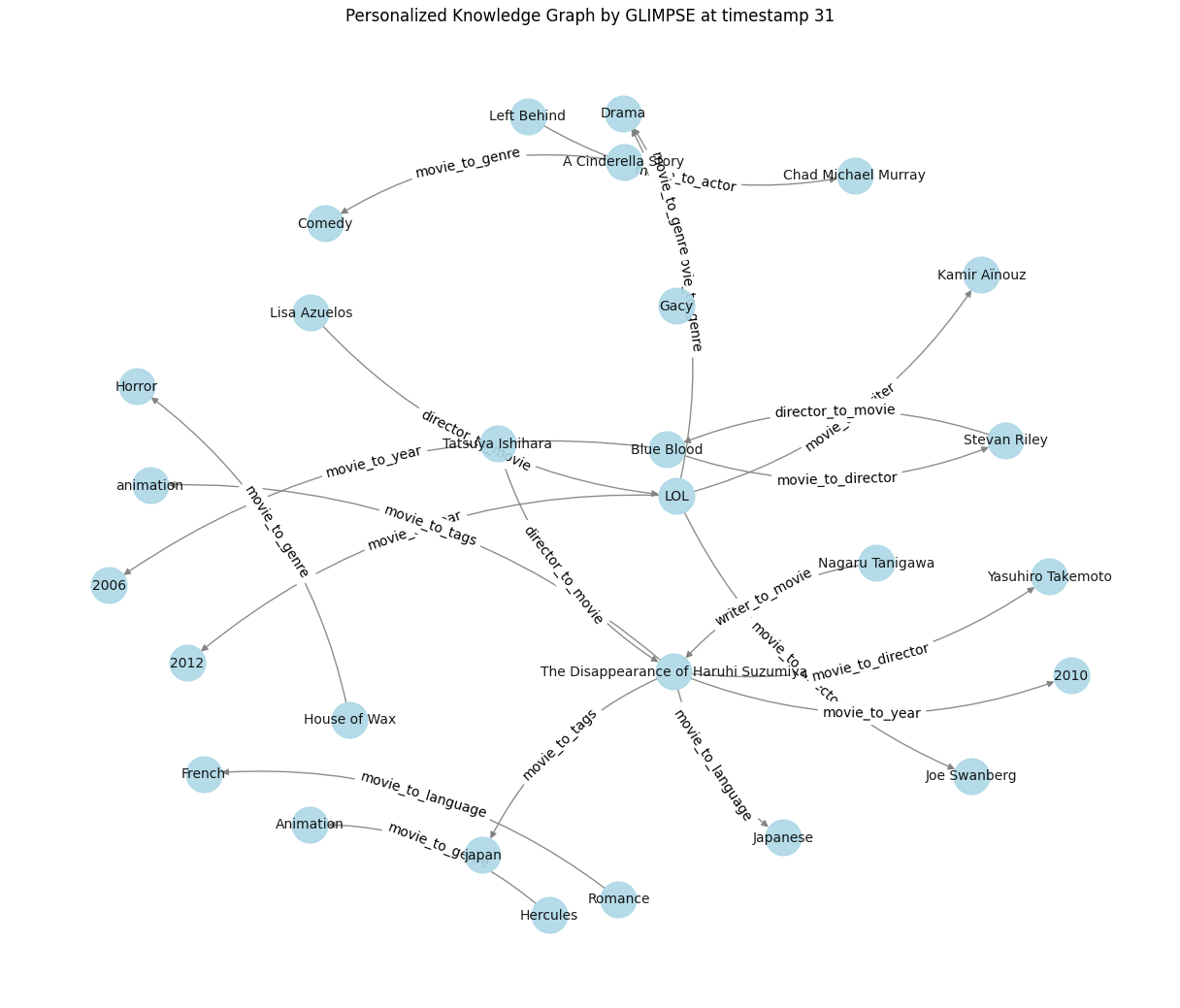}
        \caption{PKG after the Second Query on the New Topic}
    \end{subfigure}
    \begin{subfigure}[]{0.47\textwidth}
        \centering
        \includegraphics[width=\textwidth]{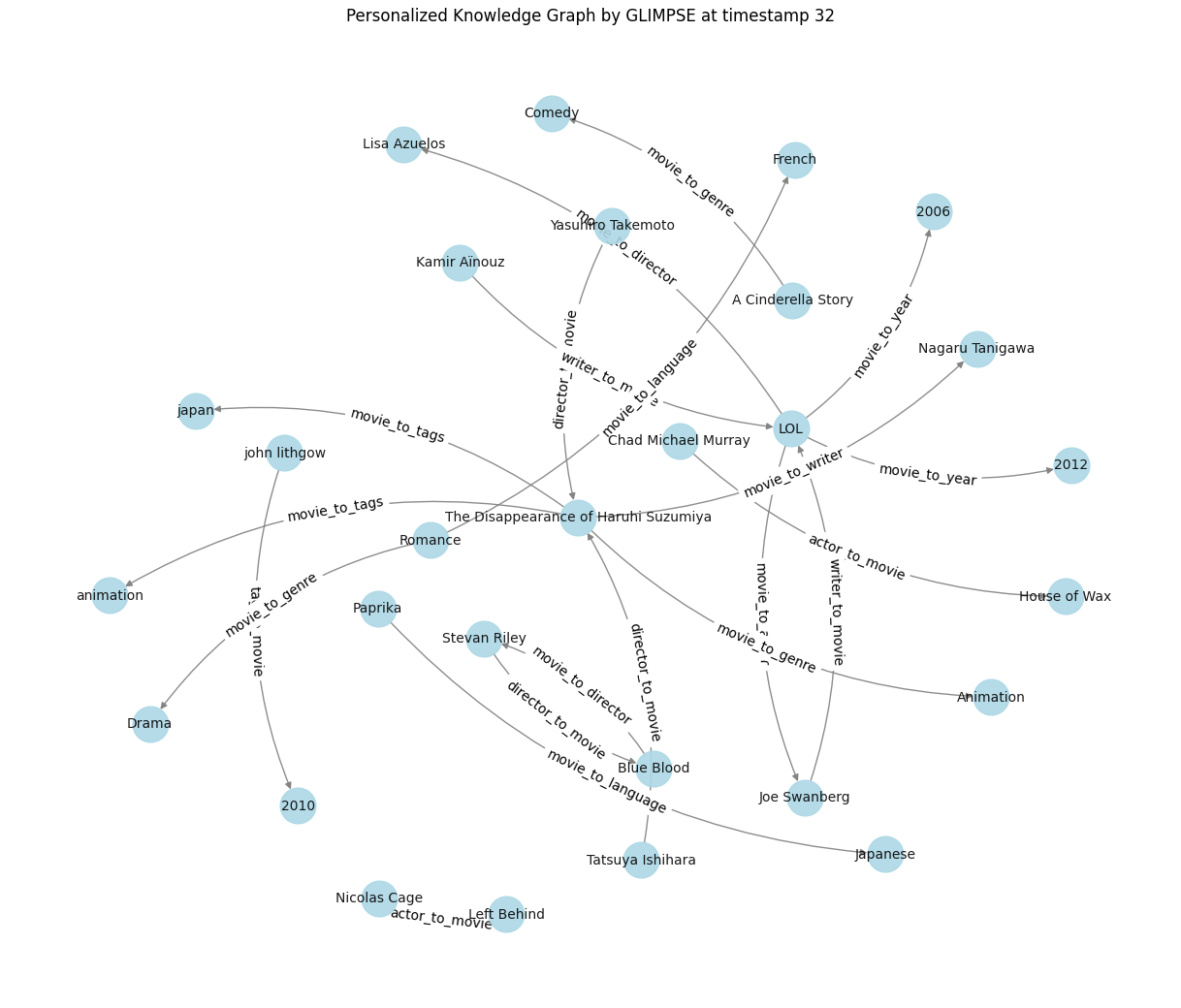}
        \caption{PKG after the Third Query on the New Topic}
    \end{subfigure}
    \caption{Case Study on GLIMPSE}
    \label{fig: case_study_glimpse}
\end{figure*}

\begin{figure*}[t]
    \centering
    \begin{subfigure}[]{0.47\textwidth}
        \centering
        \includegraphics[width=\textwidth]{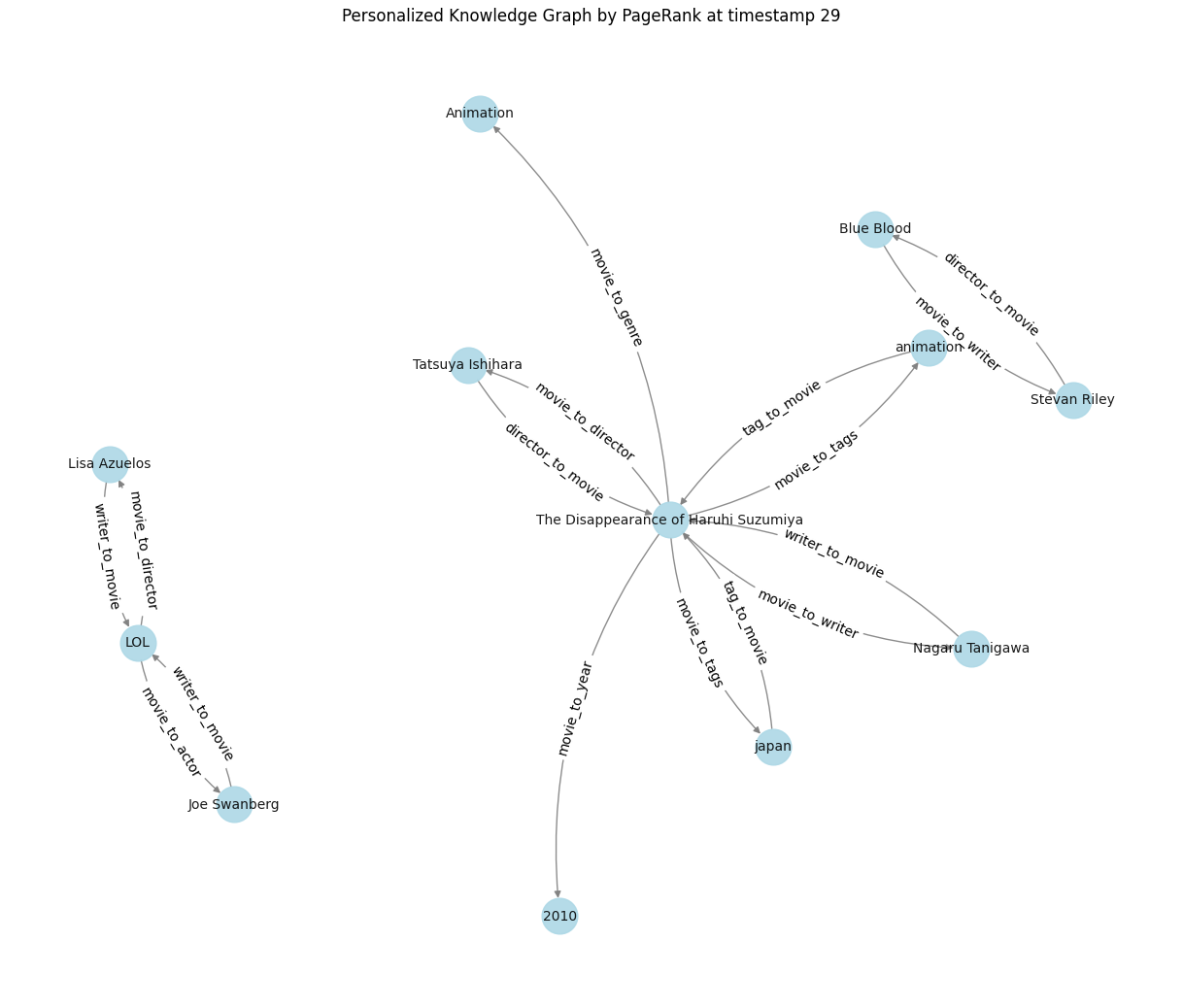}
        \caption{PKG Before the New Topic}
    \end{subfigure}
    \begin{subfigure}[]{0.47\textwidth}
        \centering
        \includegraphics[width=\textwidth]{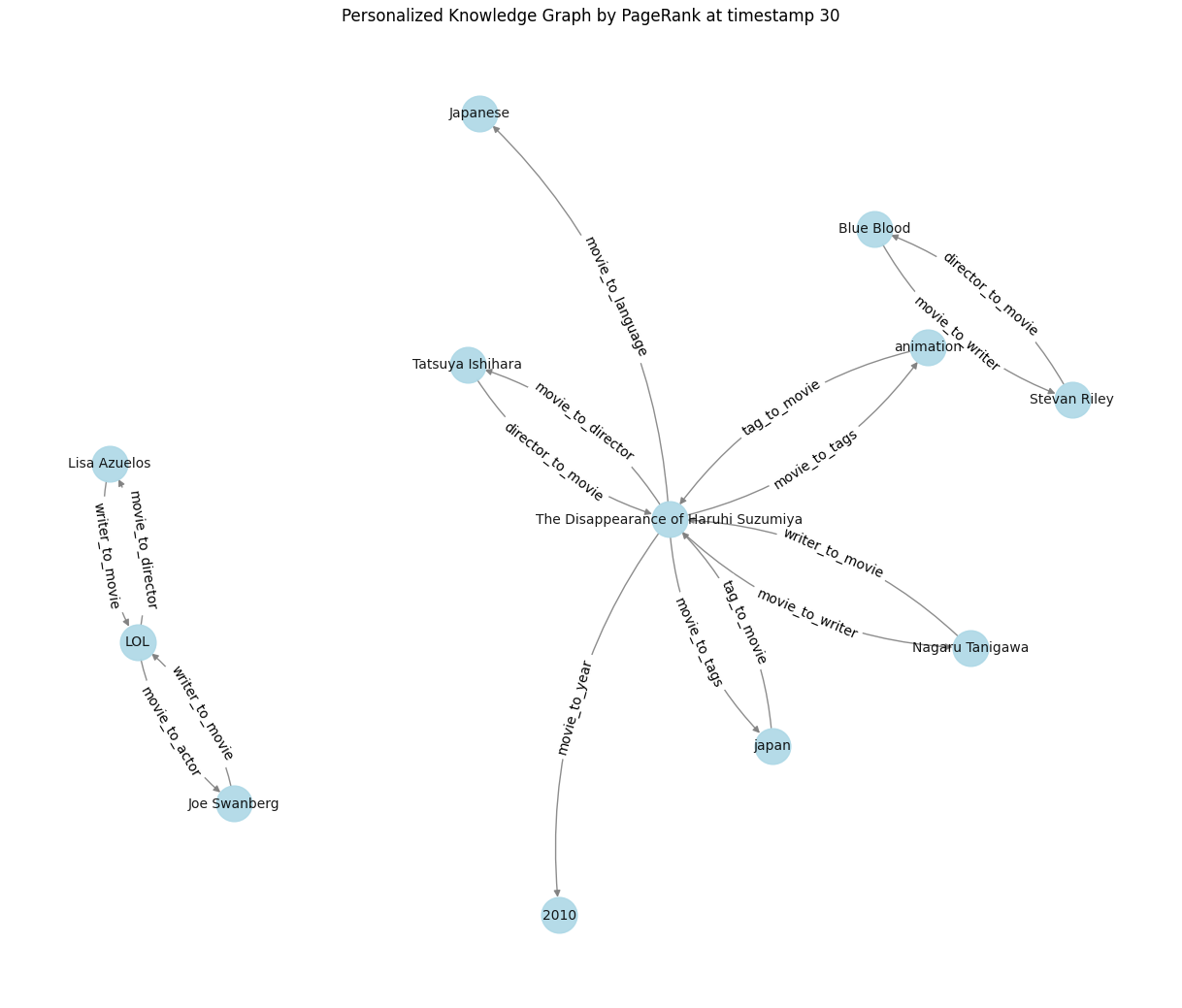}
        \caption{PKG after the First Query on the New Topic}
    \end{subfigure}
    \vskip\baselineskip
    \begin{subfigure}[]{0.47\textwidth}
        \centering
        \includegraphics[width=\textwidth]{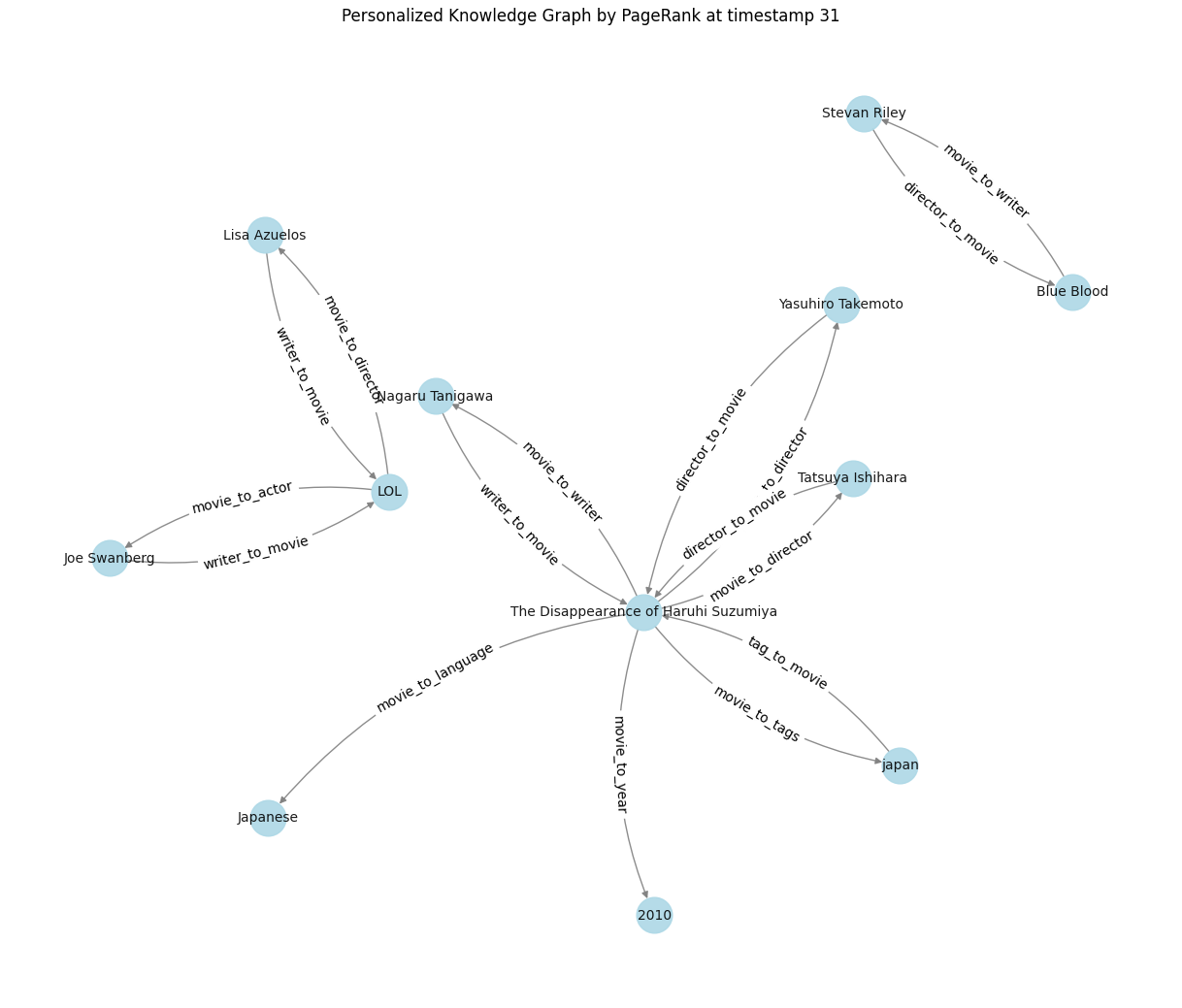}
        \caption{PKG after the Second Query on the New Topic}
    \end{subfigure}
    \begin{subfigure}[]{0.47\textwidth}
        \centering
        \includegraphics[width=\textwidth]{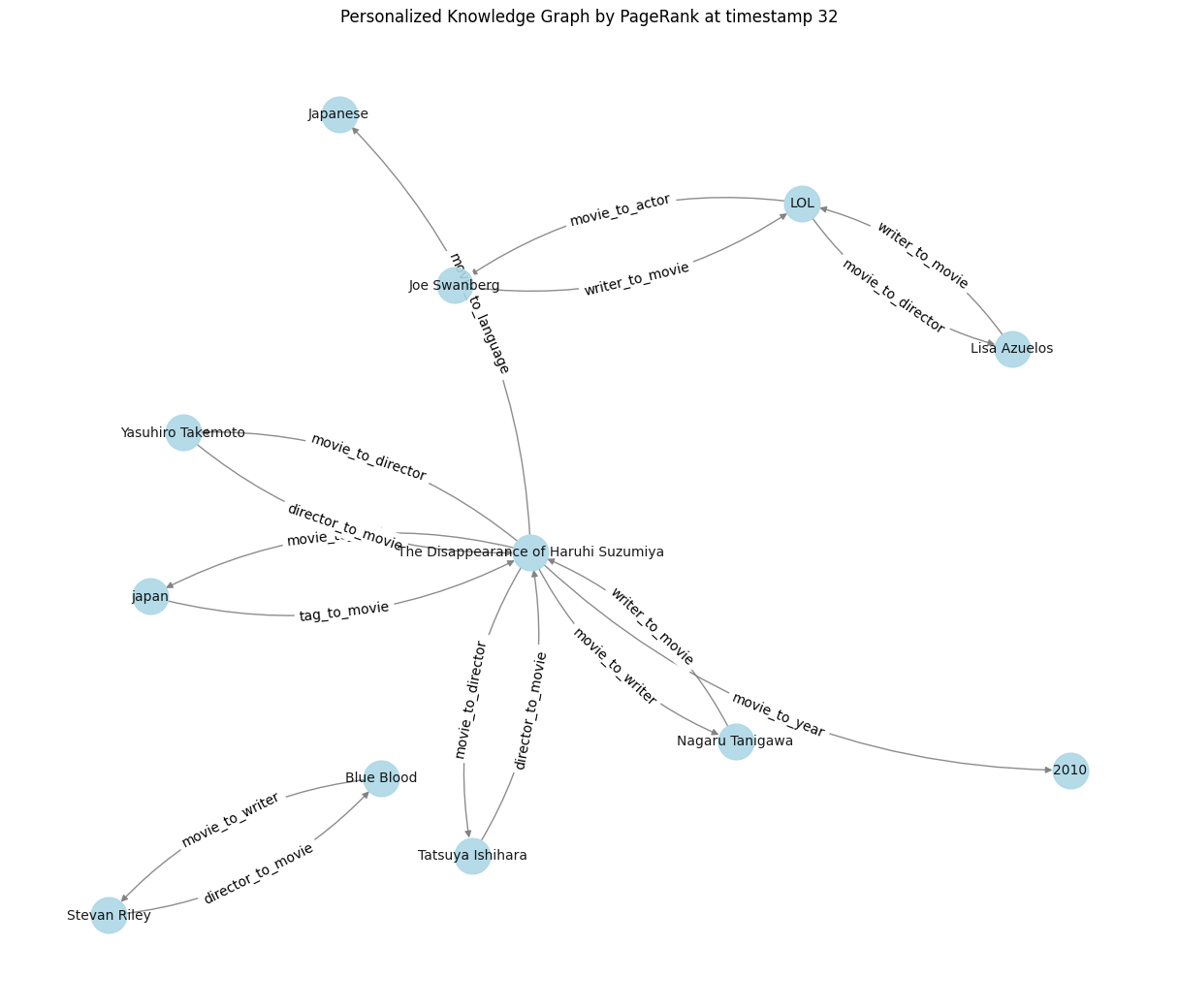}
        \caption{PKG after the Third Query on the New Topic}
    \end{subfigure}
    \caption{Case Study on PageRank}
    \label{fig: case_study_pagerank}
\end{figure*}

\begin{onecolumn}
\end{onecolumn}
\clearpage

\subsection{Proof of Theorem 3.1}
\label{AP: A.1}
\begin{proof}
Here, we give the proof for the non-adaptability of GLIMPSE~\cite{DBLP:conf/icdm/SafaviBFMMK19}.

We define "average connectivity of an area (sub knowledge graph) $\mathcal{V} = (\mathcal{E}_v, \mathcal{R}_v, \mathcal{T}_v)$" as $\frac{\sum_{v \in \mathcal{E}_v}degree(v)}{|\mathcal{E}_v|}$.

We then define "average connectivity of an area (sub knowledge graph) $\mathcal{V} = (\mathcal{E}_v, \mathcal{R}_v, \mathcal{T}_v)$" as $\frac{\sum_{v \in \mathcal{E}_v}degree(v)}{|\mathcal{E}_v|}$.

Assume two areas $\mathcal{U}$ and $\mathcal{V}$ with connectivity $c_u$ and $c_v$.

Let the user initially query $\mathcal{U}$ for $a$ times, i.e., the query set $|\mathcal{Q}_u| = a$.

For each query, $\sum_{e \in \mathcal{E}_u} {\rm Pr}(e, \mathcal{Q}_u)$ is expected to increase $2(1 + \alpha c_u)$. Then, for $a$ many queries, the total increase is $2a(1 + \alpha c_u)$.

Assume the user's interest shifts to $\mathcal{V}$ and queries $\mathcal{V}$ for $b$ times. In the same way, the total increase is $2b(1 + \alpha c_v)$.
s
Assume the total query is $\mathcal{Q}$, where $\mathcal{Q}_u \subseteq \mathcal{Q}$ and $\mathcal{Q}_v \subseteq \mathcal{Q}$.

Based on Eq.~\ref{rk_def}, we have $\frac{\sum_{r_k \in \mathcal{R}_u}Pr(r_k|\mathcal{Q})}{\sum_{r_k \in \mathcal{R}_v}Pr(r_k|\mathcal{Q})} = \frac{a}{b}$. Moreover, we can derive the relation of expectation $\frac{\mathbb{E}_{r,u}}{\mathbb{E}_{r,v}} = \frac{a|\mathcal{R}_v|}{b|\mathcal{R}_u|}$, where $\mathbb{E}_{r,u}$ is the average time being queried for each relation in $\mathcal{R}_u$ and $\mathbb{E}_{r,u} \times |\mathcal{R}_u| = \sum_{r_k \in \mathcal{R}_u} Pr(r_k | \mathcal{Q})$. Also, $\mathbb{E}_{e,u}$ is defined as the average time being queried for each entity in $\mathcal{E}_u$. $\mathbb{E}_{r,v}$, $\mathbb{E}_{e,v}$ are defined similarly.

To adapt to the user's interest shift, the summarized PKG should have $\phi(\mathcal{U}, \mathcal{Q}) < \phi(\mathcal{V}, \mathcal{Q})$. According to Eq.~\ref{phiPQ}, it equals to
\begin{equation}
\begin{split}
    \sum_{e\in\mathcal{E}_u} \log {\rm Pr}(e|\mathcal{Q}) + \sum_{x_{ijk} \in \mathcal{T}_u} \log {\rm Pr}(x_{ijk}|\mathcal{Q}) &< \sum_{e\in\mathcal{E}_v} \log {\rm Pr}(e|\mathcal{Q}) + \sum_{x_{ijk} \in \mathcal{T}_v} \log {\rm Pr}(x_{ijk}|\mathcal{Q})\\
    \prod_{e\in\mathcal{E}_u} {\rm Pr}(e|\mathcal{Q}) \prod_{x_{ijk} \in \mathcal{T}_u} {\rm Pr}(x_{ijk}|\mathcal{Q}) &< \prod_{e\in\mathcal{E}_v} {\rm Pr}(e|\mathcal{Q}) \prod_{x_{ijk} \in \mathcal{T}_v}  {\rm Pr}(x_{ijk}|\mathcal{Q})\\
    \prod_{e\in\mathcal{E}_u} {\rm Pr}(e|\mathcal{Q}) \prod_{x_{ijk} \in \mathcal{T}_u} \lambda{\rm Pr}(e_i|\mathcal{Q}){\rm Pr}(r_k|\mathcal{Q}){\rm Pr}(e_j|\mathcal{Q}) &< \prod_{e\in\mathcal{E}_v} {\rm Pr}(e|\mathcal{Q}) \prod_{x_{ijk} \in \mathcal{T}_v}  \lambda{\rm Pr}(e_i|\mathcal{Q}){\rm Pr}(r_k|\mathcal{Q}){\rm Pr}(e_j|\mathcal{Q})
\end{split}
\end{equation}

We approximate this inequality by evaluating the expectations of items on both sides.
\begin{equation}
\begin{split}
        \prod_{e\in\mathcal{E}_u} \mathbb{E}_{e,u} \prod_{x_{ijk} \in \mathcal{T}_u} \lambda\mathbb{E}_{e,u}\mathbb{E}_{r,u}\mathbb{E}_{e,u} &< \prod_{e\in\mathcal{E}_v} \mathbb{E}_{e,v} \prod_{x_{ijk} \in \mathcal{T}_v}  \lambda\mathbb{E}_{e,v}\mathbb{E}_{r,v}\mathbb{E}_{e,v}\\
        \lambda^{|\mathcal{T}_u|}\mathbb{E}_{e,u}^{|\mathcal{E}_u|} \mathbb{E}_{e,u}^{2|\mathcal{T}_u|}\mathbb{E}_{r,u}^{|\mathcal{T}_u|} &< \lambda^{|\mathcal{T}_v|}\mathbb{E}_{e,v}^{|\mathcal{E}_v|} \mathbb{E}_{e,v}^{2|\mathcal{T}_v|}\mathbb{E}_{r,v}^{|\mathcal{T}_v|} \\
        \lambda^{|\mathcal{T}_u|}\mathbb{E}_{e,u}^{|\mathcal{E}_u| + 2|\mathcal{T}_u|}\mathbb{E}_{r,u}^{|\mathcal{T}_u|} &< \lambda^{|\mathcal{T}_v|} \mathbb{E}_{e,v}^{|\mathcal{E}_v| + 2|\mathcal{T}_v|}\mathbb{E}_{r,v}^{|\mathcal{T}_v|}\\
        \lambda^{|\mathcal{T}_u|}(\frac{2a(1 + \alpha c_u)}{|\mathcal{E}_u|})^{|\mathcal{E}_u| + 2|\mathcal{T}_u|}\mathbb{E}_{r,u}^{|\mathcal{T}_u|} &< \lambda^{|\mathcal{T}_v|} (\frac{2b(1 + \alpha c_v)}{|\mathcal{E}_v|})^{|\mathcal{E}_v| + 2|\mathcal{T}_v|}\mathbb{E}_{r,v}^{|\mathcal{T}_v|}\\
        \lambda'^{|\mathcal{T}_u|}(\frac{2a(1 + \alpha c_u)}{|\mathcal{E}_u|})^{|\mathcal{E}_u| + 2|\mathcal{T}_u|}(a|\mathcal{R}_v|)^{|\mathcal{T}_u|} &< \lambda'^{|\mathcal{T}_v|} (\frac{2b(1 + \alpha c_v)}{|\mathcal{E}_v|})^{|\mathcal{E}_v| + 2|\mathcal{T}_v|}(b|\mathcal{R}_u|)^{|\mathcal{T}_v|}
\end{split}
\end{equation}

If $\mathcal{U}$ and $\mathcal{V}$ have similar size, i.e. $|\mathcal{E}_u| \approx |\mathcal{E}_v|$ and $|\mathcal{R}_u| \approx |\mathcal{R}_v|$, then we can further approximate by
\begin{equation}
\begin{split}
        (a(1 + \alpha c_u))^{|\mathcal{E}| + 2|\mathcal{T}|}a^{|\mathcal{T}|} &< (b(1 + \alpha c_v))^{|\mathcal{E}| + 2|\mathcal{T}|}b^{|\mathcal{T}|}\\
        a^{|\mathcal{E}| + 3|\mathcal{T}|}(1 + \alpha c_u)^{|\mathcal{E}| + 2|\mathcal{T}|} &< b^{|\mathcal{E}| + 3|\mathcal{T}|}(1 + \alpha c_v)^{|\mathcal{E}| + 2|\mathcal{T}|}\\
        a(\frac{(1 + \alpha c_u)}{(1 + \alpha c_v)})^{\frac{|\mathcal{E}| + 2|\mathcal{T}|}{|\mathcal{E}| + 3|\mathcal{T}|}} &< b
\end{split}
\end{equation}

It means that $b$ needs to be roughly the same scale with $a$ to finish the interest shift and completes the proof.
\end{proof}

\subsection{Proof of Theorem 3.2}
\label{AP: A.2}
\begin{proof}
Here, we give the proof for the non-adaptability of PEGASUS~\cite{DBLP:conf/icde/KangLS22}.

Similar to proof on in~\ref{AP: A.2} for GLIMPSE~\cite{DBLP:conf/icdm/SafaviBFMMK19}, we also assume two areas $\mathcal{U}$ and $\mathcal{V}$ with connectivity $c_u$ and $c_v$. Let the user initially query $\mathcal{U}$ for $a$ times. The query set $|\mathcal{Q}_u| = a$. The user's interest then shifts to $\mathcal{V}$ and queries $\mathcal{V}$ for $b$ times.

For PEGASUS here, we show that the historical search on $\mathcal{U}$ will permanently give high weights to some edges in $\mathcal{U}$, and hence these summarized items in $\mathcal{U}$ (with high weights) are hard to get replaced by items of later interests.

Consider two search queries in $\mathcal{U}$ search $u_1$ and $u_2$ (i.e., target nodes), then the edge $u_1u_2$ gets weight $\frac{1}{Z}$ since $D(u_1, \mathcal{T}) = D(u_2, \mathcal{T}) = 0$. The edge between $u_2$ and any $u_1$'s 1-hop neighbor $u'_1$ gets the weight at least $\frac{\alpha^{-1}}{Z}$ because $D(u'_1,\mathcal{T}) <= 1$ and $D(u_2,\mathcal{T}) = 0$. Similarly, we can infer that the edge between any $u_1$'s $x$-hop neighbor and $u_2$'s any $y$-hop neighbor gets weight $\frac{\alpha^{-(x+y)}}{Z}$.

Thus, no matter how the user searches in $\mathcal{V}$, these weights remain unchanged. The newly assigned weights in $\mathcal{V}$ will not exceed $\frac{\alpha^{-1}}{Z}$. But from the previous search, many edges in $\mathcal{U}$ can have weight $\frac{\alpha^{-1}}{Z}$, which means that those items (in $\mathcal{U}$ with high weights) are hard to be replaced and will still appear in the summarized graph even though the user's interest has shifted to $\mathcal{V}$.
\end{proof}

\subsection{Proof of Theorem 4.2}
\label{AP: A.3}
\begin{proof}
First, we assume the query set and its volume $|\mathcal{Q}_u| = a$, and the user's interest shifts to $\mathcal{V}$ and queries $\mathcal{V}$ for $b$ times. Then, we denote the total query is $Q$, where $Q_u \subseteq Q$ and $Q_v \subseteq Q$. Define $\mathbb{E}_{e,u}$ and $\mathbb{E}_{e,v}$ as the average time being queried for each entity in $\mathcal{E}_u$ and $\mathcal{E}_v$, respectively; and $\mathbb{E}_{r,u}$ and $\mathbb{E}_{r,v}$ as the average time being queried for each relation in $\mathcal{R}_u$ and $\mathcal{R}_v$, respectively.
\begin{equation}
\begin{split}
    \sum_{x_{ijk} \in \mathcal{T}_u} \log {\rm Pr}(x_{ijk}|\mathcal{Q}) &< \sum_{x_{ijk} \in \mathcal{T}_v} \log {\rm Pr}(x_{ijk}|\mathcal{Q}) \\
    \prod_{x_{ijk} \in \mathcal{T}_u} {\rm Pr}(e_i|\mathcal{Q}){\rm Pr}(r_k|\mathcal{Q}){\rm Pr}(e_j|\mathcal{Q}) &< \prod_{x_{ijk} \in \mathcal{T}_v}  {\rm Pr}(e_i|\mathcal{Q}){\rm Pr}(r_k|\mathcal{Q}){\rm Pr}(e_j|\mathcal{Q})\\
    \mathbb{E}_{e,u}^{|\mathcal{E}_u| + 2|\mathcal{T}_u|}\mathbb{E}_{r,u}^{|\mathcal{T}_u|} &< \mathbb{E}_{e,v}^{|\mathcal{E}_v| + 2|\mathcal{T}_v|}\mathbb{E}_{r,v}^{|\mathcal{T}_v|}\\
\end{split}
\label{complete_shift}
\end{equation}

Considering the decay, the expectations are computed as
\begin{equation}
\begin{split}
    |\mathcal{E}_u|\mathbb{E}_{e,u} &= \gamma^b\sum_{t = 0}^{a-1}\gamma^t2(\sum_{i = 0}^{d} (\alpha c_u)^i) = \frac{\gamma^b-\gamma^{a+b}}{1-\gamma} \cdot 2 \cdot \frac{1-(\alpha c_u)^{d+1}}{1-\alpha c_u}\\
    |\mathcal{E}_v|\mathbb{E}_{e,v} &= \sum_{t = 0}^{b-1}\gamma^t2(\sum_{i = 0}^{d} (\alpha c_v)^i) = \frac{1-\gamma^{b}}{1-\gamma} \cdot 2 \cdot \frac{1-(\alpha c_v)^{d+1}}{1-\alpha c_v}\\
    |\mathcal{R}_u|\mathbb{E}_{r,u} &= \gamma^b\sum_{t = 0}^{a-1}\gamma^t = \frac{\gamma^b-\gamma^{a+b}}{1-\gamma}, \
    |\mathcal{R}_v|\mathbb{E}_{r,v} = \sum_{t = 0}^{b-1}\gamma^t = \frac{1-\gamma^{b}}{1-\gamma}
\end{split}
\end{equation}

Plug these above into Eq. \ref{complete_shift}. Then, assuming $\mathcal{U}$ and $\mathcal{V}$ have similar size, i.e. $|\mathcal{E}_u| \approx |\mathcal{E}_v|$ and $|\mathcal{R}_u| \approx |\mathcal{R}_v|$ gives
\begin{equation}
    (\frac{\gamma^b-\gamma^{a+b}}{1-\gamma} \cdot \frac{1-(\alpha c_u)^{d+1}}{1-\alpha c_u} \cdot \frac{2}{|\mathcal{E}_u|})^{|\mathcal{E}_u| + 2|\mathcal{T}_u|} (\frac{\gamma^b-\gamma^{a+b}}{|\mathcal{R}_v|(1-\gamma)})^{|\mathcal{T}_u|} < (\frac{1-\gamma^b}{1-\gamma} \cdot \frac{1-(\alpha c_v)^{d+1}}{1-\alpha c_v} \cdot \frac{2}{|\mathcal{E}_v|})^{|\mathcal{E}_v| + 2|\mathcal{T}_v|} (\frac{1-\gamma^b}{|\mathcal{R}_u|(1-\gamma)})^{|\mathcal{T}_v|}
\end{equation}

If $\mathcal{U}$ and $\mathcal{V}$ have similar size, i.e. $|\mathcal{E}_u| \approx |\mathcal{E}_v|$ and $|\mathcal{R}_u| \approx |\mathcal{R}_v|$, then we can further approximate by
\begin{equation}
\begin{split}
    ((\gamma^b-\gamma^{a+b}) \cdot \frac{1-(\alpha c_u)^{d+1}}{1-\alpha c_u} )^{|\mathcal{E}_u| + 2|\mathcal{T}_u|} (\gamma^b-\gamma^{a+b})^{|\mathcal{T}_u|} &< ((1-\gamma^b) \cdot \frac{1-(\alpha c_v)^{d+1}}{1-\alpha c_v})^{|\mathcal{E}_v| + 2|\mathcal{T}_v|} (1-\gamma^b)^{|\mathcal{T}_v|}\\
    (\gamma^b-\gamma^{a+b}) \cdot (\frac{1-(\alpha c_u)^{d+1}}{1-\alpha c_u})^{\frac{|\mathcal{E}_u| + 2|\mathcal{T}_u|}{|\mathcal{E}_u| + 3|\mathcal{T}_u|}} &< (1-\gamma^b) \cdot (\frac{1-(\alpha c_v)^{d+1}}{1-\alpha c_v})^{\frac{|\mathcal{E}_v| + 2|\mathcal{T}_v|}{|\mathcal{E}_v| + 2|\mathcal{T}_v|}}\\
    b &> \log_\gamma \frac{1}{\frac{A}{B}(1-\gamma^a)+1}
\end{split}
\end{equation}
where $A = (\frac{1-(\alpha c_u)^{d+1}}{1-\alpha c_u})^{\frac{|\mathcal{E}_u| + 2|\mathcal{T}_u|}{|\mathcal{E}_u| + 3|\mathcal{T}_u|}}$ and $B = (\frac{1-(\alpha c_v)^{d+1}}{1-\alpha c_v})^{\frac{|\mathcal{E}_v| + 2|\mathcal{T}_v|}{|\mathcal{E}_v| + 2|\mathcal{T}_v|}}$. And $\log_\gamma \frac{1}{\frac{A}{B}(1-\gamma^a)+1} < \log_\gamma \frac{1}{\frac{A}{B}+1}$ further gives us a bound $b < \log_\gamma \frac{1}{\frac{A}{B}+1}$.
\end{proof}

\subsection{Proof of Theorem 4.4}
\label{AP: A.4}
\begin{proof}
First, we assume query set $|\mathcal{Q}_u| = a$, and the user's interest shifts to $\mathcal{V}$ and queries $\mathcal{V}$ for $b$ times. Also, we denote the total query is $Q$, where $Q_u \subseteq Q$ and $Q_v \subseteq Q$.
\begin{equation}
\begin{split}
    \sum_{e\in\mathcal{E}_u} \log {\rm Pr}(e|\mathcal{Q})&< \sum_{e\in\mathcal{E}_v} \log {\rm Pr}(e|\mathcal{Q})\\
    \prod_{e\in\mathcal{E}_u} {\rm Pr}(e|\mathcal{Q}) &< \prod_{e\in\mathcal{E}_v} {\rm Pr}(e|\mathcal{Q})\\
    \mathbb{E}_{e,u}^{|\mathcal{E}_u|} &< \mathbb{E}_{e,v}^{|\mathcal{E}_v|}\\
    (\frac{\gamma^b-\gamma^{a+b}}{1-\gamma} \cdot 2 \cdot \frac{1-(\alpha c_u)^{d+1}}{1-\alpha c_u} \cdot \frac{1}{|\mathcal{E}_u|})^{|\mathcal{E}_u|} &< (\frac{1-\gamma^{b}}{1-\gamma} \cdot 2 \cdot \frac{1-(\alpha c_v)^{d+1}}{1-\alpha c_v} \cdot \frac{1}{|\mathcal{E}_v|})^{|\mathcal{E}_v|}\\
\end{split}
\end{equation}

If $\mathcal{U}$ and $\mathcal{V}$ have similar size, i.e. $|\mathcal{E}_u| \approx |\mathcal{E}_v|$, then we can further approximate by
\begin{equation}
\begin{split}
    (\gamma^b-\gamma^{a+b}) \cdot \frac{1-(\alpha c_u)^{d+1}}{1-\alpha c_u} &< (1-\gamma^{b}) \cdot \frac{1-(\alpha c_v)^{d+1}}{1-\alpha c_v} \iff 
    b > \log_\gamma \frac{1}{\frac{A}{B}(1-\gamma^a)+1}
\end{split}
\end{equation}
where $A = \frac{1-(\alpha c_u)^{d+1}}{1-\alpha c_u}$ and $B = \frac{1-(\alpha c_v)^{d+1}}{1-\alpha c_v}$. Knowing that $\log_\gamma \frac{1}{\frac{A}{B}(1-\gamma^a)+1} < \log_\gamma \frac{1}{\frac{A}{B}+1}$ gives us a bound $b < \log_\gamma \frac{1}{\frac{A}{B}+1}$
\end{proof}

\end{document}